%% file: main.tex
\documentclass[11pt]{article}

\usepackage[T1]{fontenc}
\usepackage[utf8]{inputenc}
\usepackage[english]{babel}

\usepackage[left=1in,right=1in,top=1in,bottom=1in]{geometry}
\usepackage{parskip}

\usepackage{amsmath}

\usepackage{amssymb,amsthm,amsfonts,mathtools}
\usepackage{bm}
\allowdisplaybreaks
\usepackage{bbm}

\usepackage{booktabs,array,tabularx}
\usepackage{graphicx}
\usepackage{algorithm}
\usepackage{algpseudocode}
\usepackage{tikz}
\usetikzlibrary{arrows.meta,positioning}
\usetikzlibrary{decorations.pathreplacing}

\usepackage{xcolor}
\usepackage{url}

\usepackage[authoryear,round]{natbib}

\usepackage{authblk}

\newtheorem{theorem}{Theorem}
\newtheorem{lemma}[theorem]{Lemma}
\newtheorem{proposition}[theorem]{Proposition}

\theoremstyle{definition}
\newtheorem{example}{Example}

\theoremstyle{remark}

\usepackage{hyperref}
\hypersetup{colorlinks=true,linkcolor=blue!70!black,citecolor=blue!70!black,urlcolor=blue!70!black}

\newcommand{\throughput}{\textbf{Throughput}}
\newcommand{\Latency}{\textbf{Latency}}
\newcommand{\TTFT}{\textbf{TTFT}}

\input{command.tex}

\newcommand*{\circled}[1]{\lower.7ex\hbox{\tikz\draw (0pt, 0pt)%
    circle (.5em) node {\makebox[1em][c]{\small #1}};}}

\algnewcommand{\algorithmicand}{\textbf{and }}
\algnewcommand{\algorithmicor}{\textbf{or }}
\algnewcommand{\OR}{\algorithmicor}
\algnewcommand{\AND}{\algorithmicand}
\makeatletter
\newenvironment{breakablealgorithm}
  {%
  \begin{center}
     \refstepcounter{algorithm}%
     \hrule height.8pt depth0pt \kern2pt%
     \renewcommand{\caption}[2][\relax]{%
      {\raggedright\textbf{\ALG@name~\thealgorithm} ##2\par}%
      \ifx\relax##1\relax
         \addcontentsline{loa}{algorithm}{\protect\numberline{\thealgorithm}##2}%
      \else
         \addcontentsline{loa}{algorithm}{\protect\numberline{\thealgorithm}##1}%
      \fi
      \kern2pt\hrule\kern2pt
     }
  }{%
     \kern2pt\hrule\relax
  \end{center}
  }
\makeatother

\title{\Large Optimizing LLM Inference: Fluid-Guided Online Scheduling with Memory Constraints}

\author[1]{Ruicheng Ao\thanks{E-mail: \texttt{aorc@mit.edu}, \texttt{luogan@stu.pku.edu.cn}, \texttt{dslevi@mit.edu}, \texttt{xinshang.w@alibaba-inc.com}.}}
\author[2]{Gan Luo}
\author[1]{David Simchi-Levi}
\author[3]{Xinshang Wang}

\affil[1]{Institute for Data, Systems, and Society, Massachusetts Institute of Technology, Cambridge, MA 02139}
\affil[2]{School of Mathematical Sciences, Peking University, Beijing 100871}
\affil[3]{Alibaba Group}

\date{\today}

\begin{document}

\maketitle

\begin{abstract}
\noindent
\input{abstract}
\end{abstract}


\input{introduction}
\input{model}
\input{fluid}
\input{known_type}
\input{unknown_type}
\input{numerical}
\input{conclusion}

\section*{Acknowledgments}
We sincerely thank Dr. Zijie Zhou and Prof. Jing Dong for their valuable suggestions and insightful comments, which significantly helped improve the quality of this work.

\bibliographystyle{plainnat}
\bibliography{main.bib}

\newpage
\appendix
\renewcommand{\theHsection}{appendix.\Alph{section}}
\renewcommand{\theHsubsection}{appendix.\Alph{section}.\arabic{subsection}}
\renewcommand{\theHsubsubsection}{appendix.\Alph{section}.\arabic{subsection}.\arabic{subsubsection}}
\input{extension}
\clearpage
\input{appendix_notation}
\clearpage

\input{Appendix}

\end{document}

%% file: command.tex
\newcommand{\prn}[1]{\left({#1}\right)}

\newcommand{\ex}[2]{\mathbb{E}_{#1}\left[#2\right]}

\newcommand{\real}{\mathbb{R}}

%% file: abstract.tex
Large language models now serve millions of users daily, with providers incurring costs exceeding \$700{,}000 per day. Each request requires token-by-token inference, making GPU scheduling central to latency, capacity, and cost. The difficulty is endogenous memory growth: generated tokens expand the Key-Value (KV) cache, and overflow can evict in-progress requests and waste prior computation. We formulate inference as a multi-stage online scheduling problem with endogenous memory growth, linear iteration times, and GPU-resident KV-cache constraints. We introduce a fluid model that characterizes equilibrium batch composition, memory requirement, and the fluid stability region. Guided by the fluid model, we design WAIT (Waiting for Accumulated Inference Threshold), a threshold-based admission rule for known output lengths, and Nested WAIT, which extends the rule to unknown output lengths by regulating how requests advance across decode-stage segments. Under the stated threshold and memory conditions, WAIT approximates the fluid benchmark asymptotically; for unknown output lengths, Nested WAIT obtains guarantees on throughput, latency, and eviction avoidance under an additional safety buffer. In Vidur simulations configured for Llama-2-7B on an A100 GPU, with supplemental real-GPU validation reported in the appendix, the policies enlarge the empirically observed stable operating range relative to the tested baseline configurations and reduce latency especially in near-overloaded and overloaded regimes.

%% file: introduction.tex
\section{Introduction}

Large Language Models (LLMs) now dominate natural language processing (NLP) \citep{devlin2019bert, brown2020language, kaplan2020scaling, ouyang2022training, wei2022emergent, wei2022chain, touvron2023llama, openai2024gpt4technicalreport}, powering chatbots \citep{anthropic2023claude, bai2023qwen, openai2024gpt4technicalreport, deepseekai2025deepseekr1incentivizingreasoningcapability}, search engines \citep{google2023bard, microsoft2023bing}, and programming assistants \citep{copilot, claude_3_7}. These models generate text through \textit{inference}, which imposes significant computational demands on memory and processing resources \citep{garcia2019estimation}. Systems like ChatGPT incur daily inference costs exceeding \$700,000 and contribute to substantial carbon emissions \citep{patel2023peeling, patterson2021carbon}. Efficient inference scheduling can reduce these operational costs and energy consumption \citep{strubell2019energy, desislavov2021compute} by balancing system throughput and response latency \citep{wu2022sustainable}.

\begin{figure}
    \centering
    \includegraphics[width=0.55\linewidth]{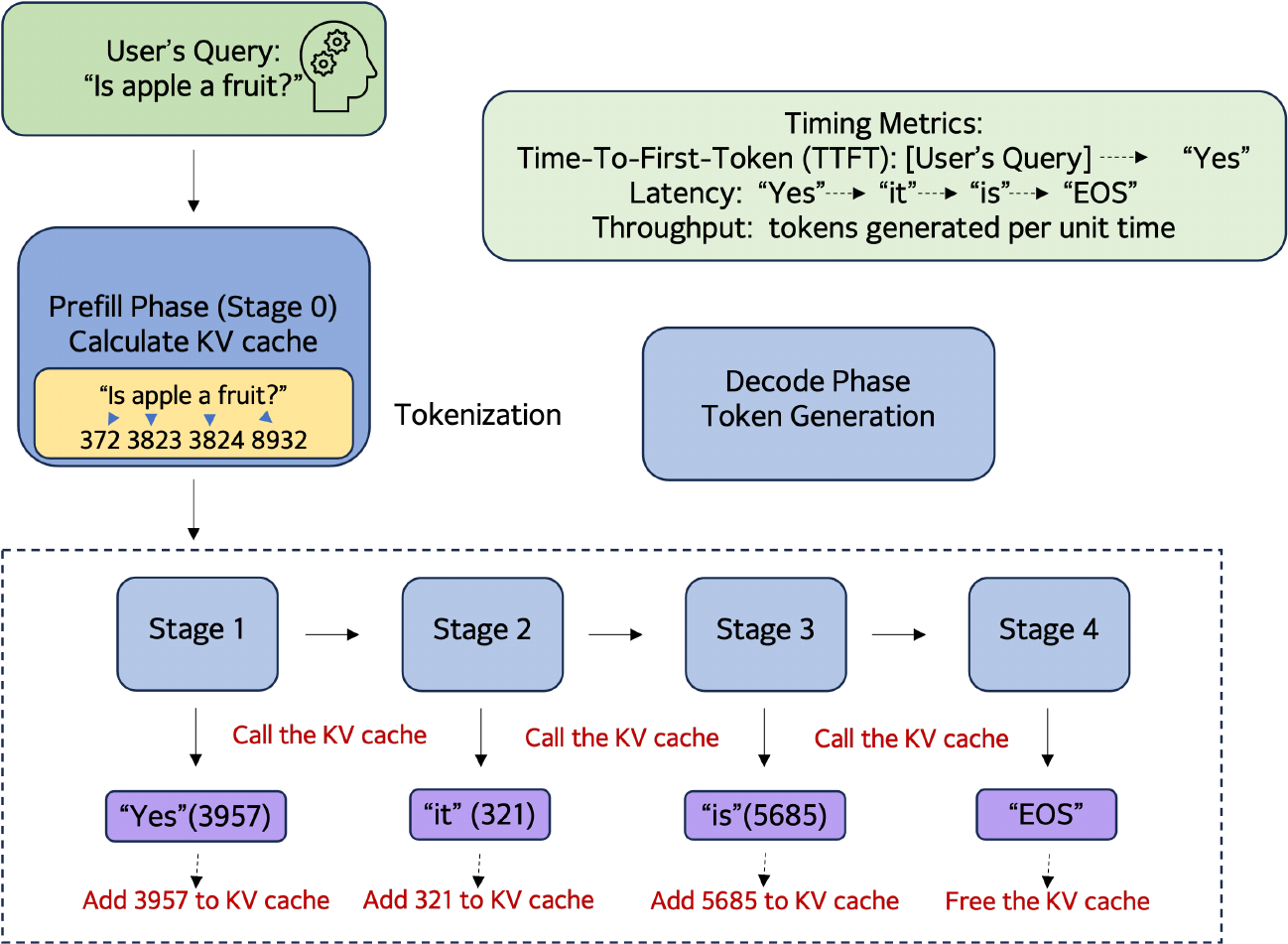}
    \caption{An example of LLM inference.}
    \label{fig:intro_example}
\end{figure}

Figure~\ref{fig:intro_example} illustrates how a query is processed during LLM inference. The user submits a prompt (e.g., ``Is apple a fruit?''), which then passes through two computational phases:
\begin{itemize}
    \item \emph{Prefill phase (stage 0).} Upon receiving the prompt, the model first tokenizes the input into a sequence of discrete units (e.g., ``Is'', ``apple'', ``a'', ``fruit'', ``?''). These tokens are then embedded and simultaneously processed in a single forward pass to compute the \emph{Key-Value (KV) cache}, which stores intermediate representations (i.e., attention keys and values) for each token. These precomputed values enable efficient reuse during subsequent decoding steps. This phase corresponds to Stage 0 in Figure~\ref{fig:intro_example}, where all prompt tokens are embedded and their KV representations are added to the cache.

    \item \emph{Decode phase (stages 1 to $l^\prime$).} After the prefill phase, the model enters the decode phase, where it generates the output one token at a time. At each decode stage, the model queries the existing KV cache to compute the next token, appends the new token to the output sequence, and updates the KV cache with its key-value pair. For instance, the model might first generate ``Yes'' (Stage 1), then ``it'' (Stage 2), followed by ``is'' (Stage 3), and finally the ``End of Sequence'' (EOS) token (Stage 4). This progression is depicted in the lower part of Figure~\ref{fig:intro_example}. Each token generation step involves both reading from and writing to the KV cache, so the KV cache grows linearly with the length of the generated sequence.
\end{itemize}

Figure~\ref{fig:intro_example} also highlights key performance metrics in LLM inference:
\begin{itemize}
    \item \emph{Time-to-first-token} (TTFT) measures the latency from user input to the first generated token.
    \item \emph{Latency} refers to the total time required to complete the generation of all output tokens.
    \item \emph{Throughput} captures the average number of tokens generated per unit time.
\end{itemize}

This inference structure, particularly the growing memory requirements and sequential decoding pattern, introduces fundamental constraints on scheduling and batching. Efficient scheduling must account for both prompt heterogeneity and KV cache dynamics to balance latency and resource use.

Scheduling LLM inference tasks involves grouping prompts into \textit{batches} processed concurrently on a GPU. Batching improves throughput: processing multiple prompts together amortizes the fixed overhead of each iteration and better uses GPU parallelism. However, this benefit comes with a fundamental tension: larger batches consume more memory through their combined KV caches. The KV cache improves efficiency, as it prevents the model from recalculating attention history for each new token; without it, computational cost would scale quadratically with sequence length. But the KV cache also grows dynamically during decode, making memory consumption unpredictable. This creates a core trade-off: \emph{larger batches increase throughput but consume more memory, risking capacity overflow}.

Memory overflow can occur in several practical scenarios: (i) \emph{bursty arrivals}, where sudden traffic spikes temporarily exceed capacity; (ii) \emph{unknown output lengths}, where prompts generate longer responses than anticipated; and (iii) \emph{multi-tenant environments}, where multiple users compete for GPU memory. The unknown output length problem is particularly challenging: since the scheduler cannot predict how much memory each prompt will eventually consume, it cannot reliably plan batch composition or admission decisions. Traditional scheduling methods such as Shortest Job First (SJF) assume fixed job sizes and known processing times, making them unsuitable when memory demands grow unpredictably \citep{tay2022efficient, kang2024gear, hooper2025kvquant}.

When the total KV cache exceeds GPU memory capacity, the system must evict some in-progress prompts. Two approaches exist: (i) \emph{swapping} to CPU or SSD, which incurs I/O overhead \citep{sheng2023flexgen, aminabadi2022deepspeed}; and (ii) \emph{recomputation}, which discards the KV cache and restarts the prompt from prefill \citep{kwon2023efficient}. We focus on recomputation because it is the default in modern production systems such as vLLM \citep{vllm-v1-2025}, and because our goal is to \emph{minimize eviction} through scheduling rather than to optimize the eviction mechanism itself \citep{li2025mirage}. Eviction creates a vicious cycle: it frees memory temporarily, but restarted prompts consume the memory again and can trigger further evictions. Operationally, this appears as failed or aborted generations under heavy load, including user-facing errors such as ``Something went wrong.''

\emph{Memory sufficiency alone does not guarantee load balance.} The key to approaching the offered-load benchmark is \emph{load balance}: arrivals, completions, and memory growth must remain aligned. Supporting such an operating point requires enough memory for the corresponding mix of active prompts, quantified later by the memory requirement \(M^*\). Even when total capacity \(C\) equals or only slightly exceeds this requirement, naive policies can push the stochastic system away from the corresponding composition of active prompts across prefill and decode stages. For instance, first-come-first-served (FCFS) can trigger cascading evictions that lead to a meaningful loss in effective throughput, measured as completed decode-token service net of evictions, near the fluid boundary (Example~\ref{ex:fcfs_cascade}). A slight imbalance causes one prompt type to overflow, triggering evictions; the freed memory then fills with another type, which itself overflows, creating a vicious cycle. Thus, memory feasibility of the fluid composition does not by itself ensure that an online policy will operate near that composition.

Preventing eviction requires \emph{approaching load balance through controlled admission}. The key scheduling decision is the composition of each inference iteration: which output-length classes and prefill/decode stages enter the next batch. This composition determines how much fixed overhead is shared in the current iteration, how much KV cache is carried forward, and whether future iterations approach the overflow boundary. Our algorithms use thresholds to regulate this composition. A prompt advances only when the corresponding threshold has enough room, keeping the in-service population close to the equilibrium composition across prefill and decode stages. In our experiments, operating near this composition \emph{enlarges the empirically observed stable operating range}, because threshold control reduces eviction cascades and moves observed stability closer to the fluid prediction, and \emph{reduces mean end-to-end latency most clearly near capacity and under overload}. In underloaded regimes, our algorithms and the baselines are often close; as load increases, baseline latency grows sharply and completion rates saturate earlier, while our algorithms keep the system closer to the offered load. Section~\ref{sec:experiments} documents these effects.

Recent system-level optimizations for LLM inference \citep{yu2022orca, kwon2023efficient, agrawal2023sarathi, pope2023efficiently, deepseekai2024deepseekv2strongeconomicalefficient, patel2024splitwise, zhong2024distserve} are primarily systems and implementation contributions. They motivate, but do not aim to characterize, the queueing structure created by stochastic arrivals, growing KV caches, and evictions. In practice, output length predictions are imprecise or costly \citep{fu2025efficient}, and algorithms designed for known output lengths can degenerate significantly when this assumption fails (see Proposition~\ref{prop:unknown_lower_bound}).

We develop a fluid approximation that formalizes this operating point as a fluid equilibrium, where prompt arrivals and completions balance, and characterizes the memory requirement \(M^*\) needed to support it. For a fixed capacity \(C\), and under the load and atomic-feasibility assumptions stated in the model, the condition \(M^*\le C\) determines whether the arrival vector lies in the corresponding fluid stability region. This equilibrium guides the threshold design: when the stochastic system operates near the corresponding composition across prefill and decode stages, it converts the offered work into completed service while avoiding eviction. The analysis shows that memory is a resource to exploit rather than merely a constraint, because larger batches process more prompts per iteration and increase GPU utilization.

Achieving this equilibrium in the stochastic system is not automatic, because each admission decision changes both the current batch and the future KV-cache state. We therefore use the structure of the fluid equilibrium to construct threshold policies. With known output lengths, WAIT sets admission thresholds from the reference composition across prefill and decode stages: a prompt advances only when the corresponding prefill/decode-stage threshold is met, so admission restores the load-balanced allocation rather than creating downstream memory pressure. With unknown output lengths, Nested WAIT applies the same rule through nested decode segments: short prompts complete and release memory early, while prompts that remain active advance to later segments without output-length prediction. The threshold structure reduces the original memory-coupled control problem to lower-dimensional queueing dynamics while preserving the batching benefits needed to approach the offered-load benchmark.

\subsection{Summary of Contributions}

Our main contributions are as follows.
\begin{itemize}
    \item \emph{Memory-constrained scheduling model.} We develop a multi-stage online scheduling model with endogenous memory growth, where KV-cache memory grows during LLM inference and exceeding capacity triggers costly prompt evictions (Section~\ref{sec:model}). With fixed external arrivals, a stable policy completes work at the arrival rate, so the design goal is twofold: to operate within the fluid stability region when it is feasible, and to remain near the equilibrium composition across prefill and decode stages supported by the memory requirement \(M^*\). A fluid approximation from queueing theory lets us characterize both (Section~\ref{sec:fluid}).
    \item \emph{Eviction-prevention scheduling algorithms.} We develop threshold-based algorithms that control memory growth and reduce or avoid eviction under the stated memory conditions. The WAIT algorithm (Section~\ref{sec:wait}) handles known output lengths; the Nested WAIT algorithm (Section~\ref{sec:nested_wait}) extends to unknown output lengths through on-the-fly output-length classification. Endogenous memory growth makes the primitive state high-dimensional, because the scheduler must track resident prompts across output-length classes and prefill/decode stages. The threshold construction reduces this state to threshold and boundary queues that capture the memory-coupled dynamics. Under the linear multi-stage model with endogenous memory growth and the stated asymptotic scaling, WAIT approximates the fluid benchmark asymptotically; Nested WAIT provides guarantees on throughput, latency, and eviction avoidance under its safety buffer.
    \item \emph{Experimental validation.} We evaluate our algorithms on synthetic and real-world workloads using Vidur~\citep{agrawal2024vidur}, a widely used LLM-serving simulator, configured for Llama-2-7B on a single A100 GPU (Section~\ref{sec:experiments}). We compare against vLLM \citep{kwon2023efficient} and Sarathi \citep{agrawal2023sarathi}, two widely used open-source greedy FCFS-based scheduling baselines that represent, respectively, memory-reactive eviction and recomputation, and chunked-prefill scheduling. We also report real-GPU validation on A100 hardware, including direct timing measurements and end-to-end GPU experiments, providing implementation evidence as well as support for the simulator's accuracy (Section~\ref{sec:model} and Appendix~\ref{app:gpu_validation}). Consistent with this load-balancing mechanism, WAIT and Nested WAIT expand the empirically observed stable operating range compared with the tested baselines and reduce mean end-to-end latency most visibly near capacity and under overload.
\end{itemize}
Our batching theory also extends naturally to Prefill-Decode Disaggregated (PD) systems such as DistServe \citep{zhong2024distserve} and Splitwise \citep{patel2024splitwise}, in which the decode server handles decode-only workloads. In this setting, the linear iteration-time model in Section~\ref{sec:model} is a close approximation after calibration, because prefill attention effects are separated from decode service. The same threshold construction applies on the decode side, and Appendix~\ref{app:pd_disagg} reports numerical results on representative PD workloads.

\subsection{Other Related Work}

\paragraph*{Online Scheduling Problem and Queueing System.} 
Classical online scheduling problems focus on optimally assigning jobs that arrive sequentially, each with varying completion times. In settings with stochastic arrivals, foundational studies such as \cite{devanur2009adwords,vee2010optimal,cole2014sample,balkanski2016power,lattanzi2020online} have developed algorithms that learn arrival patterns or distributions to refine scheduling strategies over time. Beyond individual job processing, works like \cite{im2013online,lucier2013efficient,liu2015online,li2020online} have explored simultaneous batching and scheduling, grouping similar jobs to enhance efficiency. Within queueing theory, fluid models serve as first-order approximations of stochastic systems, enabling near-optimal control strategies \cite{mandelbaum1998strong,maglaras2000discrete,bauerle2002optimal,liu2011network}, while asymptotic analysis characterizes long-horizon behavior. LLM inference introduces a different control structure. As decode progresses, each prompt increases its KV-cache requirement, so the scheduler must track not only how many jobs wait, but also where admitted prompts sit in the decode pipeline and how their memory requirements will evolve. Capacity pressure can then trigger eviction and recomputation. These mechanisms enlarge both the state space and the action space relative to classical queueing-control formulations with exogenous service requirements, because the system may restart work due to endogenous memory growth. Stability therefore depends on how the scheduling rule controls memory growth before overflow occurs.

\paragraph*{LLM Inference.} 
A growing literature studies scheduling and systems design for Large Language Model (LLM) inference. Unlike classical scheduling, LLM inference combines stochastic arrivals, dynamic memory constraints, and multi-phase prefill/decode processing. System-level works such as Splitwise \citep{patel2024splitwise} and DistServe \citep{zhong2024distserve} split inference into separate prefill and decode pipelines, while scheduling engines such as Orca \citep{yu2022orca} and Sarathi \citep{agrawal2023sarathi,agrawal2024taming} improve throughput through batching and admission control. Other system-level optimizations include multi-head latent attention with low-rank KV compression \citep{deepseekai2024deepseekv2strongeconomicalefficient} and Kendall's-Tau prioritization by predicted output length \citep{fu2025efficient}. These works advance system-level efficiency through implementation and engineering. We complement them by analyzing the queueing-control problem created by endogenous memory growth: as admitted prompts decode, they increase the resident KV-cache load, and overflow can trigger eviction and restart. This mechanism links the fluid stability region and throughput loss to how the scheduler regulates the GPU-resident workload across prefill and decode stages.
From an operations research perspective, several recent works analyze LLM inference scheduling under online-algorithm frameworks, focusing on competitive ratio and regret bounds. \cite{jaillet2025onlineschedulingllminference} develop an online scheduling algorithm with near-optimal regret guarantees for settings with known output lengths. \cite{wang2025llm} study optimization with variable prefill and decode lengths in pipeline-parallel settings. \cite{chen2025adaptively} address robust optimization under prediction uncertainty, considering adversarial scenarios. Complementary theory-and-practice work by \cite{bari2025optimal} studies scheduling algorithms for LLM inference and discusses timing regimes induced by prefill, decode, and model-computation effects. A different theoretical angle is taken by \cite{li2025throughput}, whose stochastic processing model allows general batch processing times, including a piecewise-linear form that captures the compute-bound / memory-bound transition, and shows that work-conserving algorithms achieve the fluid-optimal throughput. On output length uncertainty, \cite{jaillet2025onlineschedulingllminference} and \cite{wang2025llm} assume lengths are known at arrival, \cite{chen2025adaptively} considers adversarial unknown lengths, while our Nested WAIT algorithm handles stochastic unknown lengths through on-the-fly classification at segment boundaries. These works establish competitive-ratio, regret, or throughput-optimality guarantees under their respective timing and information models. Our work focuses on a complementary queueing mechanism: the scheduler must control the composition of the GPU-resident workload across prefill and decode stages when iteration time is memory dependent and overflow causes eviction and restart. The resulting fluid analysis links the fluid stability region to the iteration-time model in Equation~\eqref{eq:time_consump} and explains how threshold-based admission moves realized performance closer to that fluid prediction.

\subsection{Notations}
For integer $n\ge 1,$ we denote $[n]=\{1,2,\dots,n\}$ as the set of integers from $1$ to $n$. For $x\in\real$, denote $\lceil x\rceil$ as the smallest integer not smaller than $x$ and $\lfloor x\rfloor$ as the largest integer not greater than $x$. Denote $x_+=\max\{x, 0\}$. For set $S$, denote $|S|$ as its cardinality. For two functions $f(T)$ and $g(T)$, we use $f(T)=O(g(T))$ if there exists constant $c_1>0$ such that $f(T)\le c_1g(T)$ as $T\to+\infty$ and $f(T)=\Omega(g(T))$ if there exists constant $c_2>0$ such that $f(T)\ge c_2g(T)$ as $T\to +\infty$. 

%% file: model.tex
\section{Model}
\label{sec:model}

We model Large Language Model (LLM) inference as an online scheduling system with stochastic arrivals, staged service, and a shared memory constraint. The key departure from classical scheduling is endogenous memory growth: a job's memory requirement is not fixed at admission, because as the model generates output tokens, the prompt's \textit{Key-Value (KV) cache} grows over time. The notation is defined where it first appears, with a consolidated summary in Table~\ref{tab:notation} of Appendix~\ref{app:notation}.

\subsection{Prompts and Inference Process}

Prompts arrive to a single Graphics Processing Unit (GPU) over time and require two phases of service. The \textit{prefill} phase embeds the input context and initializes the prompt's KV cache. The \textit{decode} phase then generates one output token per iteration, increasing the prompt's KV cache by one unit at each decode step. We focus on a single GPU as the basic scheduling unit; multi-GPU deployments can apply the same admission and batching logic within each GPU partition or pipeline stage.

We group prompts into \( m \) types indexed by \( j \in \{1, \dots, m\} \). Type-\(j\) prompts arrive according to a Poisson process with rate \( \lambda_j \), have input length \(l_j\) after prefill, and require \(l_j'\) decode iterations before completion.
Thus, a prompt of type \( j \) has the following length profile:
\begin{equation*}
    \left\{
    \begin{aligned}
        &\text{a length of } l_j \text{ tokens after prefill;}\\
        &\text{a length of } l_j + l_j' \text{ tokens after decode.}
    \end{aligned}
    \right.
\end{equation*}
Therefore, a type-$j$ prompt undergoes one prefill iteration followed by \(l_j'\) decode iterations. To track service progress, we use \(s=0\) for prompts awaiting prefill and \(s \in \{1, \dots, l_j'\}\) for prompts in decode. A prompt waiting outside the GPU at \(s=0\) has no resident KV cache; processing its prefill allocates \(l_j\) units, and each generated decode token adds one unit. Thus a resident type-$j$ prompt with decode progress \(s\) occupies \(l_j+s\) units of KV cache memory. This memory requirement depends on whether the prompt is awaiting prefill or has advanced into decode, which is the source of the dynamic capacity constraint analyzed below.

Though the total number of different types $m$ can be large (for example, from $1$ to $10000$), the threshold policies below aggregate the memory dynamics through per-stage or per-segment flow controls rather than tracking each prompt's entire future cache path. In the Nested WAIT analysis, the additional safety buffer depends logarithmically on the number of tracked segments. Other quantities, such as the maintained threshold vector and the constants in the delay bounds, can still depend on the number and arrival rates of the tracked classes.

While our main analysis focuses on fixed arrival rates $\lambda_j$, Appendix~\ref{sec:extension} discusses a finite-horizon extension in which thresholds are adjusted to bounded time-varying arrival rates.

\subsection{Batching and Iteration Time}
\begin{figure}[h]
    \centering
    \includegraphics[width=0.55\linewidth]{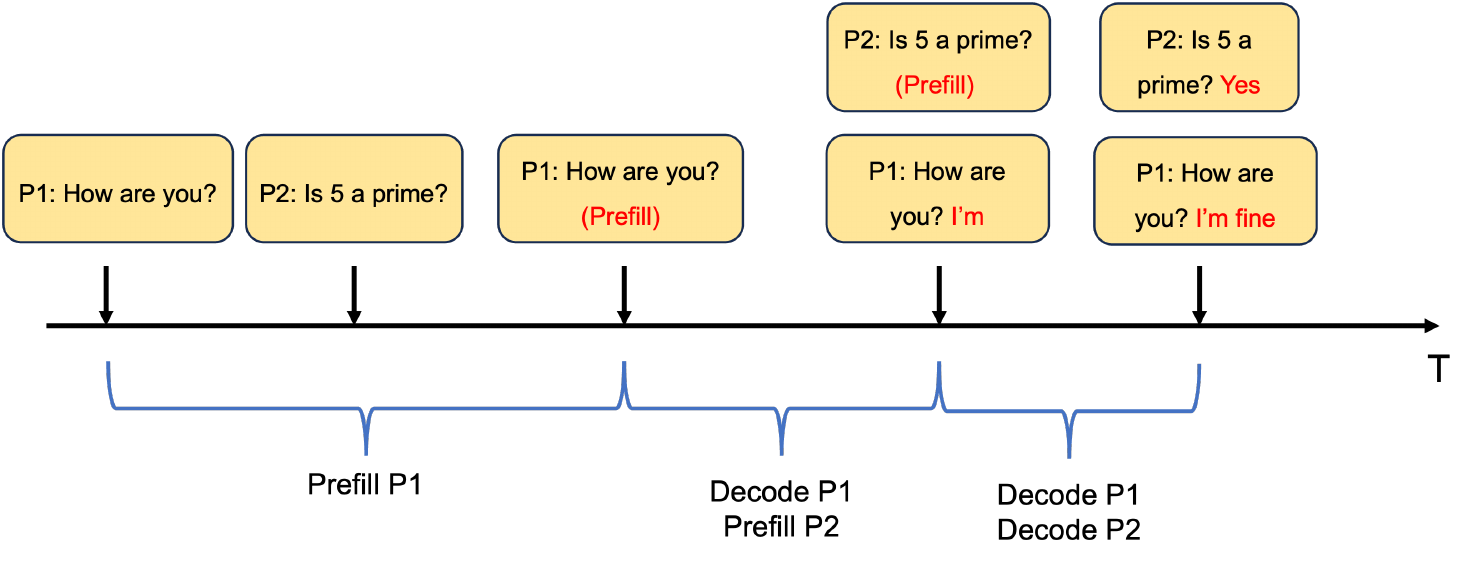}
    \caption{Example of batching and scheduling with two prompts. The three GPU iterations shown are a prefill-only batch for \(P_1\), a mixed batch that decodes \(P_1\) while prefilling \(P_2\), and a decode-only batch for both prompts.}
    \label{fig:batching_example}
\end{figure}

At each iteration, the scheduler forms a \textit{batch} of prompts to process on the GPU. A batch may include newly admitted prompts in prefill and previously admitted prompts in decode, so the scheduling decision determines both how many prompts enter service and how many in-service prompts advance by one token. Because each iteration incurs a fixed launch and synchronization overhead, larger batches generally improve token throughput by spreading this overhead across more prompts. Figure~\ref{fig:batching_example} illustrates this timing with two sequential arrivals. Prompt \(P_1\) arrives first and is processed in prefill, which initializes its KV cache. Prompt \(P_2\) arrives later and also requires prefill. The displayed processing iterations are \(B^1=\{(P_1,s=0)\}\), \(B^2=\{(P_1,s=1),(P_2,s=0)\}\), and \(B^3=\{(P_1,s=2),(P_2,s=1)\}\), so the corresponding sums in Equation~\eqref{eq:time_consump} are \(l_{j(P_1)}\), \((l_{j(P_1)}+1)+l_{j(P_2)}\), and \((l_{j(P_1)}+2)+(l_{j(P_2)}+1)\), respectively. This example highlights the operational trade-off that drives the model: batching improves throughput by sharing fixed overhead across prompts, but it also increases the KV cache load carried by the system.

\begin{figure}
    \centering
    \includegraphics[width=0.55\linewidth]{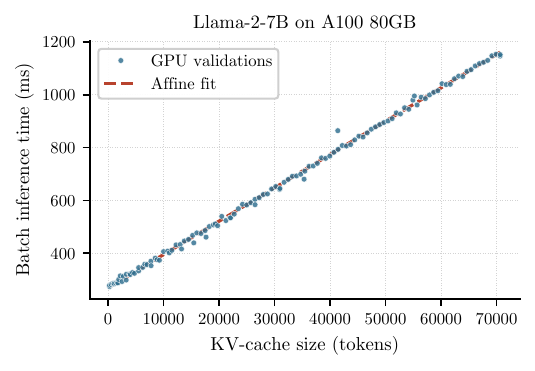}
    \caption{Batch inference time of Llama-2-7B on a single NVIDIA A100~80GB GPU. We fix the prompt length at \(256\) tokens and the output length at \(20\) tokens, and vary the number of requests in the batch from \(1\) to \(256\). Each validation point corresponds to a batch state; requests in the same batch may be at different decode stages, so the horizontal axis reports the resulting total number of tokens stored in the KV cache across active requests. The vertical axis reports measured batch inference time. The linear fit indicates an approximately linear relationship between batch inference time and total cached tokens.}
    \label{fig:inference time}
\end{figure}

We next specify the service time of an iteration. In Transformer inference, prefill evaluates the input sequence and creates the initial KV cache; thereafter, each decode iteration produces one token per active prompt while attending to the prompt's cached history. For a decode-dominated batch, the batch KV-cache size is therefore a natural workload descriptor. Let \(B^t\) denote the batch processed at iteration \(t\), let \(j(i)\) be the type of prompt \(i\), and let \(s_i^t\) be its current prefill/decode stage. A prefill prompt has \(s_i^t=0\) and initializes \(l_{j(i)}\) units of KV cache, while a decode prompt at stage \(s_i^t\) reads and extends a cache of size \(l_{j(i)}+s_i^t\). Guided by profiling studies of LLM serving workloads~\citep{agrawal2023sarathi,kwon2023efficient,zhong2024distserve}, we approximate iteration time by
\begin{equation}
\label{eq:time_consump}
    \tau(B^t) = d_0 + d_1 \sum_{i\in B^t} \bigl(l_{j(i)} + s_i^t\bigr),
\end{equation}
where \( d_0 \) is the fixed per-iteration overhead and \( d_1 \) is the marginal time cost per cached token. The linear form separates this fixed overhead from the marginal cost of serving larger KV caches. It captures the admission-control trade-off central to our model: admitting more prompts spreads \(d_0\) across more generated tokens and can therefore raise throughput, but it also lengthens subsequent iterations and consumes scarce GPU memory through larger KV caches. This regime is especially relevant because prompts typically spend most of their lifetime in decode, so admission decisions primarily determine the population and age distribution of active caches. Figure~\ref{fig:inference time} reports A100~80GB validation results for Llama-2-7B with the prompt setting fixed at input length \(256\) and output length \(20\). We vary the number of requests in the batch from \(1\) to \(256\) and record batch inference time across batch states in which the active requests may be at different decode stages. The horizontal axis is the total number of tokens stored in the KV cache across active requests, corresponding to \(\sum_{i\in B^t}(l_{j(i)}+s_i^t)\) for a serving iteration \(t\). Because requests may occupy different prefill/decode stages in the same iteration, batches with similar numbers of requests can have different total cached-token counts, producing the scattered validation points around the linear fit. Appendix~\ref{app:gpu_validation} reports additional simulator-validation and end-to-end GPU experiments for the same hardware setting.

The same formulation can accommodate other calibrated service-time curves. When prefill dominates, especially when prefill lengths are highly heterogeneous, input attention can add nonlinear length effects; mixed prefill-decode batches and hardware saturation can create distinct compute-bound and memory-bound regimes. These effects can be represented by replacing~\eqref{eq:time_consump} with a monotone function of batch composition, including piecewise-linear specifications such as \(\tau(b') = c + a\max\{0,b'-b_0\}\)~\citep{li2025throughput}. Prefill-decode disaggregated serving gives the complementary case: prefill work is routed to separate workers, and the decode server repeatedly executes one-token decode steps, making aggregate KV-cache usage an especially accurate state variable for iteration time~\citep{patel2024splitwise,zhong2024distserve}. Thus the linear model is both a close approximation in the KV-cache-driven regimes we study and a transparent first-order primitive for the fluid analysis. It keeps the central operational trade-off explicit: larger batches improve utilization, but they also increase iteration time and memory pressure. Even when a richer calibrated service-time model is needed, the same perspective remains useful: the scheduler must control the composition of resident prompts across prefill and decode stages because this composition determines both memory pressure and future service times.

\subsection{Memory Constraint and Challenges}

The service-time model above describes how a batch uses the GPU in one iteration. The second constraint is intertemporal: a prompt that remains in service carries its KV cache across iterations, and this cache grows as decode progresses. Hence the scheduling state must track not only the current batch \(B^t\), but all prompts whose KV caches are currently resident on the GPU.

\textbf{Memory Constraint}: At all times \(t\), the total KV cache stored on the GPU must not exceed capacity \(C\):
\begin{equation}
\label{eq:memory_constraint}
\sum_{i \in \mathcal{G}^t} (l_{j(i)} + s^t_i) \le C,
\end{equation}
where \(\mathcal{G}^t\) is the set of prompts whose KV caches are resident on the GPU at time \(t\). This set includes the current batch \(B^t \subseteq \mathcal{G}^t\) as well as prompts that are waiting between iterations while retaining their KV caches on the GPU. Here \(j(i)\) is the type of prompt \(i\), and \(s_i^t \in \{0,1,\dots,l_{j(i)}'\}\) is its current decode progress. Prompts that have not yet been prefilled are not in \(\mathcal{G}^t\); once their KV cache is allocated, their memory contribution grows from \(l_{j(i)}\) with each generated token. If the constraint~\eqref{eq:memory_constraint} cannot be satisfied, the server must \textit{evict} resident prompts by discarding their KV caches. We model eviction as last-in-first-out: the most recently admitted prompts are evicted first until enough memory is freed. An evicted prompt loses its accumulated computation and restarts from prefill, re-entering the scheduling queue as a new job. This restart rule can create an eviction cascade: each eviction relieves memory pressure temporarily, but the restarted prompts return to the queue, raise future admission pressure, and can force additional evictions. The following example illustrates this mechanism.
We assume throughout that each prompt is individually feasible, i.e., its largest single-prompt KV-cache footprint fits in memory. This atomic feasibility condition is separate from the fluid equilibrium requirement \(M^*\le C\), which concerns the memory needed to sustain a balanced population of many prompts.

\begin{example}[Dynamic Memory Growth and Eviction]
\label{ex:memory_growth}
Consider the running toy prompt ``Hello,'' which receives the one-token response ``Hi.'' This prompt type has prefill length \(l=1\) and decode length \(l'=1\). A prompt waiting for prefill uses no KV cache. Once its prefill is processed, the prompt becomes resident and uses one unit of KV cache; after it generates its single decode token, it uses \(l+1=2\) units just before completion. Suppose the GPU has capacity \(C=12\). One full-memory iteration can process four prefills and four decodes: the four newly prefilled prompts require \(4\times1=4\) units, and the four prompts already in decode require \(4\times2=8\) units, for a total of \(12\) units and four completions.

Now suppose three prompts are already decoding, using \(3\times2=6\) units, and the scheduler admits six new prompts for prefill, using another \(6\times1=6\) units during the iteration. After the iteration, the three old prompts complete and release their caches, while the six newly admitted prompts move into decode and now require \(6\times2=12\) units, filling memory. If four more prompts arrive next, admitting their prefills requires four additional units. The only way to create this space is to evict two resident decode prompts, each freeing two units. These evicted prompts lose their KV caches and return to the queue as new jobs, where they compete with subsequent arrivals and can force further evictions.

This example highlights the \textbf{eviction cascade} and the admission trade-off behind it. A larger batch can raise throughput because the fixed per-iteration overhead is shared across more generated tokens, but every admitted prompt also creates a KV cache that stays on the GPU and grows one token at a time until completion. If the scheduler admits too many prompts now, later decode iterations may leave too little memory for new arrivals, forcing eviction and restart. When output lengths are unknown, this decision must be made before the scheduler knows how long each prompt will remain in memory. Section~\ref{sec:fluid} formalizes the balanced admission level at which arrivals, completions, and memory growth can be kept in balance.
\end{example}

\subsection{Objective and Policy Space}

We evaluate a scheduling policy over a finite horizon \([0,T]\). The primary efficiency metric, denoted \(\throughput^{(T,\pi)}\), is \textit{effective throughput}: the average number of decode tokens from prompts that complete under policy \(\pi\). Tokens generated before a prompt is evicted do not count, because eviction discards the prompt's KV cache and returns it to prefill. This convention makes memory overflow part of the performance criterion rather than a separate implementation detail. For a fixed arrival process, effective throughput is capped by the offered decode-token load. A policy falls below this cap when it leaves capacity idle or admits work that later restarts instead of completing. Admission control must therefore balance utilization against memory feasibility: admitting enough prompts to exploit the GPU, but not so many that endogenous KV-cache growth forces eviction. Maintaining this balance over a wider range of arrival rates expands the realized operating range, and under overload it keeps completed work close to the server's usable capacity. We also evaluate \(\Latency^{(T,\pi)}\), the average elapsed time from prompt arrival to completion, and \(\TTFT^{(T,\pi)}\), the average elapsed time from arrival to the first generated output token. For finite-horizon delay metrics, every request that arrives before time \(T\) is included in the average. If a request has not completed by \(T\), its latency contribution is its elapsed time in the system up to \(T\); if it has not received its first token by \(T\), its TTFT contribution is likewise its elapsed time up to \(T\). Waiting in the external prefill queue is part of the time in system. In the theoretical expectations below, expected finite-horizon delay is the expected total censored delay over all arrivals before \(T\), normalized by the expected number of such arrivals. These delay metrics capture service quality beyond aggregate output. In particular, TTFT rules out policies that keep completion counts high by repeatedly prioritizing short prompts while delaying the first service of longer prompts. Later sections introduce the fluid model and use it to establish throughput, latency, and TTFT guarantees for the proposed policies.

Let \(\Pi\) denote the class of admissible online policies. At each decision epoch \(t\), a policy \(\pi \in \Pi\) observes the prompts that have arrived, their input lengths, their current prefill/decode stages \(s_i^t\), and whether their KV caches are resident on the GPU. The policy is non-anticipating: it cannot use future arrivals or a prompt's output length before that length is revealed by completion.

Given this information, the policy chooses which waiting prompts to admit, which resident prompts to batch in the current iteration, and whether to preempt or evict prompts. Preemption here means scheduler-side pausing: the prompt remains GPU-resident and keeps its KV cache, so it can resume without recomputation. We do not model CPU or SSD swap-out with a separate swap-in latency. Eviction removes the KV cache to free memory; unless a policy restricts the eligible repair set more narrowly, the LIFO rule evicts the most recently admitted resident prompts first. Each evicted prompt returns to stage \(s=0\) and must restart from prefill. All decisions must satisfy the memory constraint~\eqref{eq:memory_constraint} after the resulting state transition.

%% file: fluid.tex
\section{Fluid Model and Equilibrium}\label{sec:fluid}
Section~\ref{sec:model} identifies the central admission-control trade-off created by endogenous memory growth: the scheduler should admit enough prompts to exploit batching, but not so many that their KV caches later overflow memory. This section derives the balanced operating point for that trade-off. The key state variable is the composition of GPU-resident prompts across prefill and decode stages. As prompts move deeper into decode, each prompt carries a larger KV cache; if too many prompts accumulate at late stages, the system can run out of memory and restart work through eviction.

We develop a fluid model that replaces stochastic sample paths with average flows and characterizes a fluid equilibrium in which arrivals, prefill/decode transitions, and completions occur at matching rates. For a given arrival vector in the load regime where this equilibrium is finite, the equilibrium gives two reference quantities: the memory requirement \(M^*\) needed to support the equilibrium and the effective throughput \(\throughput^*\) achieved when the offered load can be completed without eviction. Equivalently, under the atomic feasibility condition from Section~\ref{sec:model} and the denominator-positivity conditions below, the fluid stability region for a fixed memory capacity \(C\) consists of arrival vectors whose equilibrium memory requirement satisfies \(M^*\le C\). We derive these quantities first for a single prompt type, which isolates the accumulation of KV cache over prefill and decode, and then for multiple prompt types sharing one memory constraint. The same equilibrium also supplies the design target for WAIT in Section~\ref{sec:wait}: keep the batch composition close to the implied composition across prefill and decode stages, thereby avoiding evictions while approaching the largest throughput attainable under the offered load and memory constraint.

\subsection{Single-Type Fluid Model}
Consider a single prompt type \(j\). A resident prompt uses \(l_j\) units of KV cache immediately after prefill and \(l_j+s\) units at decode stage \(s=1,\dots,l_j'\); after decode stage \(l_j'\), the prompt completes and releases its cache. Let \(n_j^*\) denote the equilibrium number of type-\(j\) prompts at each prefill/decode stage. The balanced state therefore has \(n_j^*\) prompts in the prefill stage and in each decode stage.

Before deriving the equilibrium, we identify the load condition under which batching can be effective. Adding prompts to a batch is useful only if the additional completions outweigh the longer iteration time. Across its prefill and decode stages, a type-\(j\) prompt contributes memory-dependent work proportional to \((l_j' + 1)(l_j + l_j'/2)\). If arrivals are high enough that this memory-dependent work alone saturates the server, batching cannot restore stability.

\begin{proposition}\label{prop:large_rate}
When the condition $\lambda_j d_1 (l_j' + 1) \left( l_j + \frac{l_j'}{2} \right) \geq 1$ is satisfied, the system is unstable in the sense that the expected average latency under any scheduling policy with arrival rate \( \lambda_j \) grows linearly with time. Formally,
$$
\mathbb{E}[\textbf{Latency}^{(T,\pi)}] = \Omega(T) \quad \text{as} \quad T \to \infty.
$$
\end{proposition}

The proof is a service-capacity cut: each completed prompt must consume \(d_1\sum_{s=0}^{l_j'}(l_j+s)\) units of memory-dependent processing time over its lifetime, and the fixed positive overhead \(d_0\) leaves no slack at the boundary under finite memory capacity (Appendix~\ref{appendix:proof_large_rate}). We therefore focus on the nontrivial regime $\lambda_j d_1 (l_j' + 1) \left( l_j + \frac{l_j'}{2} \right) <1$, where batching can support a fluid equilibrium.

At equilibrium, the \(l_j'+1\) stages each contain \(n_j^*\) prompts, so the memory usage is
\begin{equation}
    M_j^* = n_j^* \sum_{s=0}^{l_j^{\prime}}(l_j+s) = n_j^*(l_j'+1)\prn{l_j+\frac{l_j'}{2}},
\end{equation}
where \(l_j+l_j'/2\) is the average KV-cache usage of a type-\(j\) prompt across its prefill and decode stages.

Flow balance requires arrivals during one iteration to match completions. Since one iteration takes time \(d_0+d_1M_j^*\), \(\lambda_j(d_0+d_1M_j^*)\) prompts arrive and \(n_j^*\) prompts complete, giving
\begin{equation}
\label{eq:single_type_equilibrium}
\underset{\text{Time per iteration}}{\underbrace{(d_0+d_1M_j^{*} )}}\lambda_j =\left(d_0+d_1 n_j^*(l_j'+1)\left( l_j +\frac{l_j^{\prime}}{2} \right) \right)\lambda_j= n_j^*.
\end{equation}

Solving gives \(n_j^* = d_0\lambda_j/[1-d_1\lambda_j(l_j^{\prime}+1)(l_j+l_j^{\prime}/2)]\). Substituting this value into the memory expression yields
\begin{equation}
    \label{eq:memory_single}
M_j^* = n_j^*(l_j'+1)\left( l_j+\frac{l_j^{\prime}}{2} \right)
=  \frac{d_0\lambda_j(l_j^{\prime}+1)\left( l_j+\frac{l_j^{\prime}}{2} \right) }{1-d_1\lambda_j(l_j^{\prime}+1)(l_j+\frac{l_j^{\prime}}{2})}.
\end{equation}

The denominator is the slack in the batching feasibility condition; as it approaches zero, the inventory required for balance diverges. Equation~\eqref{eq:memory_single} also shows why long responses are especially demanding. For fixed input length and arrival rate, away from the critical denominator, the memory needed to support the fluid equilibrium grows on the order of \((l_j')^2\), because the number of resident prefill/decode stages and the average KV-cache size across those stages both increase with decode length. As the denominator approaches zero, this requirement diverges. Thus the memory condition defining the fluid stability region is highly sensitive to long-decode traffic, making control of the composition across prefill and decode stages central to the scheduling problem.
The balanced throughput is completed decode tokens per iteration divided by iteration time:
\begin{equation}\label{eq:throughput_fluid}
\throughput_j^* = \frac{n_j^* \cdot l_j'}{d_0+d_1n_j^*(l_j'+1)\prn{l_j+\frac{l_j'}{2}}} = \lambda_j l_j',
\end{equation}
where the equality follows from the flow-balance condition \eqref{eq:single_type_equilibrium}.

\subsection{Multiple-Type Fluid Model}
We now extend the balance calculation to \(m\) prompt types. Type \(j\) has input length \(l_j\), decode length \(l_j'\), and arrival rate \(\lambda_j\). Because all types draw from the same memory pool, the equilibrium is determined by the joint allocation of active prompts across prefill and decode stages and the aggregate KV-cache usage it induces. Let
\[
    A:=\sum_{j=1}^m \lambda_j (l_j^{\prime}+1)\left(l_j + \frac{l_j^{\prime}}{2} \right).
\]
The finite fluid equilibrium below is defined in the aggregate load regime \(d_1A<1\). If \(d_1A\ge1\), the memory-dependent work created by the offered load saturates the server as in Proposition~\ref{prop:large_rate}, and no finite offered-load fluid equilibrium is available.

The balance equations are written at the scheduler's iteration scale: an iteration selects a batch, advances active prompts by one prefill/decode stage, and updates their KV-cache usage. Let \(n_j^*\) denote the equilibrium number of type-\(j\) prompts at each prefill/decode stage. The resulting memory usage is
\begin{equation}
\label{eq:memory_multi1}
    M^* = \sum_{j=1}^m n_j^*(l_j'+1)\left(l_j + \frac{l_j^\prime}{2}\right)
\end{equation}
Flow balance requires \((d_0+d_1M^*)\lambda_j = n_j^*\) for each type \(j\), because \(n_j^*\) type-\(j\) prompts complete in one iteration. Substituting these identities into \eqref{eq:memory_multi1} gives
\begin{equation}
\label{eq:memory_multi2}
M^* = \frac{d_0 A}{1-d_1 A}
\end{equation}
The aggregate term in \eqref{eq:memory_multi2} inherits the same quadratic dependence on decode length. A small amount of long-decode traffic can therefore account for a disproportionate share of the memory requirement needed for stability, which is precisely the pressure that the threshold policies regulate.
The corresponding iteration time is
\begin{equation}    \label{eq:time_fluid}
    \begin{aligned}
            &\Delta T^* = d_0 + d_1 M^*  \\
                  &= \frac{d_0}{1 - d_1 A},
    \end{aligned}
\end{equation}
and the fluid throughput benchmark is
\begin{equation}
    \label{throughput_fluid_multiple}
    \begin{aligned}
        &\throughput^*
        = \sum_{j=1}^m \frac{n_j^* \cdot l_j^{\prime}}{d_0 + d_1 M^*}
        = \sum_{j=1}^m \lambda_j l_j^{\prime}
    \end{aligned}
\end{equation}
When the system reaches the fluid equilibrium, the equality \(\throughput^*=\sum_j\lambda_jl'_j\) is the throughput delivered by its balanced batch composition. The fluid dynamics therefore provide both an upper bound and a benchmark: they specify the effective throughput associated with the offered load, the resident composition that realizes it, and the memory \(M^*\) needed to sustain that composition. In the denominator-positive regime \(d_1A<1\), the bound is the relevant stabilizable fluid benchmark when the corresponding equilibrium memory \(M^*\) fits in physical memory. If \(d_1A\ge1\) or \(M^*>C\), the arrival vector is outside this fluid stability region, and the system is overloaded relative to the full offered-load benchmark.

\begin{proposition}\label{prop::online policy expected throughput upper bound, multiple type}
For any online policy \(\pi \in \Pi\), the expected throughput satisfies
\[
\mathbb{E}[\throughput^{(T,\pi)}] \leq \throughput^*,
\]
where \(\throughput^*\) is given in \eqref{throughput_fluid_multiple}.
\end{proposition}
The scheduling question is how closely an online policy can approach this benchmark while respecting the memory dynamics when \(M^*\le C\).
This focus on the fluid stability region, and on behavior near its boundary, follows the standard fluid-control viewpoint for queueing systems~\citep{maglaras2000discrete,bauerle2002optimal,liu2011network}. In standard queueing models with exogenous service requirements, the workload created by an admitted job is fixed. However, in this LLM inference system, current admission and batching decisions determine future KV-cache memory usage, because prompts that advance in decode remain resident and carry larger caches. Over-admission shifts future batches toward memory-heavy late-decode prompts and can force eviction before arrivals become completed service. Overly conservative admission leaves the server short of completions. The fluid equilibrium therefore identifies the composition a policy should maintain: roughly \(n_j^*\) type-\(j\) prompts at each prefill and decode stage, with total memory requirement \(M^*\). Section~\ref{sec:wait} uses \(M^*\) and \(n_j^*\) to define thresholds that keep the batch composition near this equilibrium.

%% file: known_type.tex
\section{Scheduling with Known Output Lengths: The WAIT Algorithm}
\label{sec:wait}

This section develops the \textit{Waiting for Accumulated Inference Threshold} (WAIT) algorithm for the case in which output lengths and arrival rates are known when thresholds are set. WAIT uses the fluid equilibrium from Section~\ref{sec:fluid} as a batching target: for each output-length class \(j\), it waits until enough prompts have accumulated before admitting a batch, so that the batch composition remains close to the equilibrium levels \(n_j^*\) across prefill and decode stages. This threshold structure is the key dimensionality reduction in the known-output-length case: instead of tracking every resident prompt's future KV-cache growth, the scheduler controls threshold-sized flows across prefill and decode stages. Under the corresponding memory feasibility condition, this rule prevents eviction cascades while allowing the server to approach the fluid throughput benchmark. The guarantees below require an integer threshold vector whose induced memory satisfies \(M^\pi\ge M^*\) and \(M^\pi\le C\). Section~\ref{sec:experiments} separately examines overloaded regimes, where the full offered-load benchmark is no longer feasible but the same load-balancing principle helps the scheduler use available capacity more effectively and reduce eviction-induced waste.

\subsection{Motivation: Why Naive Scheduling Fails}
\label{subsec:wait_motivation}

The condition \(C\ge M^*\) is a capacity condition, not a scheduling rule. It identifies when the fluid equilibrium can fit in memory, but it does not ensure that an online policy will keep work balanced across prefill and decode stages.

Consider a First-Come-First-Serve (FCFS) policy that batches all resident prompts whenever the server is idle, then greedily scans the external waiting queue in arrival order and admits prompts for prefill, using the LIFO restart rule from Section~\ref{sec:model} whenever the resulting resident state would exceed memory. Ties among prompts admitted in the same iteration are broken by a fixed deterministic rule. Even when \(C=M^*\), so the fluid equilibrium fits exactly in memory, FCFS can move too much work into downstream decode stages and create a persistent throughput loss. The following proposition gives a worst-case lower bound for this failure mode.

\begin{proposition}\label{prop:lower_bound_FCFS}
When \(C=M^*\), there exists an instance in which the FCFS policy described above incurs a throughput gap
\[
\throughput^*-\mathbb E[\throughput^{(T,\mathrm{FCFS})}]=\Omega(1)
\]
as \(T\to\infty\).
\end{proposition}

The proof is provided in Appendix~\ref{appendix:proof_fcfs_lower_bound}. The example below is a simpler single-type illustration of the same eviction cascade.

\begin{example}[Eviction Cascade under FCFS]
\label{ex:fcfs_cascade}
Consider a single prompt type: a prompt such as ``Hello'' with a one-token response ``Hi,'' so \(l=1\) and \(l'=1\). Each prompt uses 1 unit of memory after prefill and 2 units after decode. Let \(C=12\), \(\lambda=4\), and normalize service parameters so that \(d_0+12d_1=1\). The fluid equilibrium keeps 4 prompts at the prefill stage and 4 prompts at the decode stage, uses \(4 \times 1 + 4 \times 2 = 12\) units of memory, and completes 4 prompts per iteration.

Now suppose the system starts slightly imbalanced with 6 prompts at stage 0 (awaiting prefill) and 3 prompts at stage 1 (in decode). Under FCFS:
\begin{enumerate}
    \item \textbf{Iteration 1}: FCFS processes all 9 prompts. The 3 decode prompts complete, while the 6 prefill prompts advance to decode and require \(6 \times 2 = 12\) units of memory. The system is full after only 3 completions.
    \item \textbf{Iteration 2}: About 4 new prompts arrive during the previous iteration. Admitting them for prefill requires 4 additional memory units, so the server evicts 2 decode prompts to free \(2 \times 2 = 4\) units. The system returns to 4 prefill prompts and 4 decode prompts, again using \(4+8=12\) units.
    \item \textbf{Iteration 3}: The 4 decode prompts complete. The 2 evicted prompts have lost their KV caches and must restart from prefill, re-entering the queue alongside genuinely new arrivals. The restart mechanism perpetuates the imbalance.
\end{enumerate}
The eviction-restart cycle yields approximately 3--3.5 completions per iteration instead of the 4 completions sustained by the fluid equilibrium. Thus \(C=M^*=12\) is sufficient to support the fluid balance, but FCFS does not maintain that balance and loses effective throughput.
\end{example}

The operational implication is that having enough memory to support the fluid equilibrium does not by itself guarantee that the stochastic system will operate near that equilibrium. The scheduler must also regulate the composition of prompts across prefill and decode stages. FCFS admits prompts greedily, lets this composition drift away from the fluid balance, and thereby creates the eviction-restart cycle.

\subsection{The WAIT Algorithm}
\label{subsec:wait_algorithm}

The FCFS example points to the missing control variable: the composition of the in-service population across prefill and decode stages. WAIT regulates this composition by assigning each type \(j\) a threshold \(n_j\), interpreted as the target number of type-\(j\) prompts to serve from each prefill/decode stage in an iteration. If fewer than \(n_j\) new type-\(j\) prompts are waiting at entry, WAIT does not admit that type. Once the threshold is reached, the server processes up to \(n_j\) prompts from each prefill/decode stage of that type. In this way, new work enters the system only when it can replenish a balanced allocation across prefill and decode stages, rather than pushing the system toward the downstream buildup seen under FCFS.

The threshold rule also gives the analysis its tractable structure. Once the thresholds are fixed, each type evolves as a threshold queue observed at batch-completion epochs, while the shared memory interaction is captured by the feasibility of the threshold vector. The waiting requirement is therefore not merely a safety device against eviction; it also preserves the batching benefit created by the fixed overhead \(d_0\). Batches form at a scale large enough to use the server effectively, while the thresholds keep memory usage within capacity.

\begin{figure}
    \centering
    \includegraphics[width=0.85\linewidth]{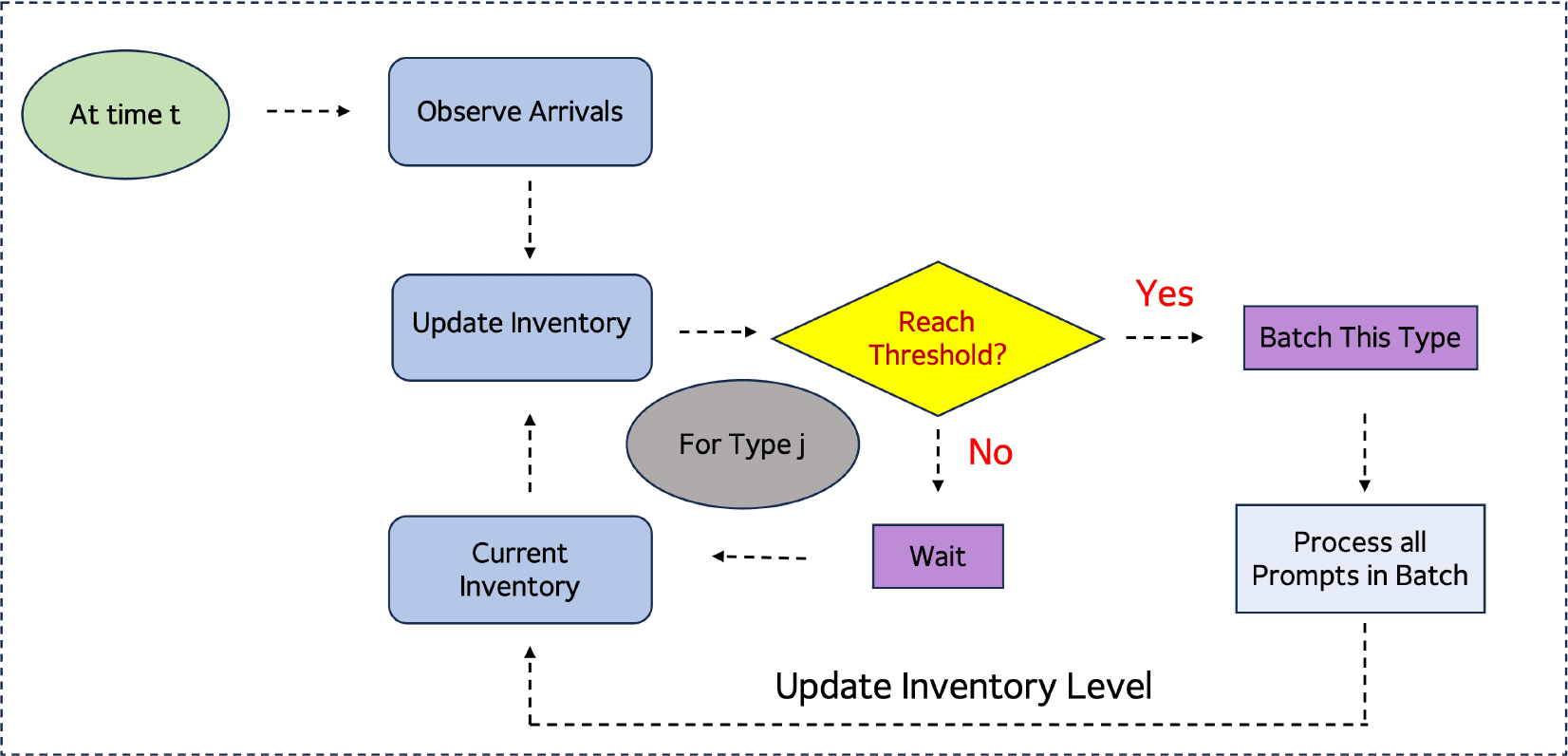}
    \caption{Batch formation under WAIT for a fixed type \(j\). The scheduler updates the type-\(j\) inventory and batches this type only when the prefill-entry queue reaches \(n_j\); otherwise, it waits and carries the inventory forward. When batched, up to \(n_j\) prompts from each prefill/decode stage are advanced.}
    \label{fig:wait}
\end{figure}

\paragraph{Algorithm Description.} For each prompt type \(j \in [m]\), WAIT sets a threshold \(n_j\) derived from the fluid model. At decision epoch \(t\), let \(n_{js}^t\) denote the number of type-\(j\) prompts at stage \(s\). Type \(j\) is eligible for the next batch only when
\[
n_{j0}^t \geq n_j.
\]
When this condition holds, WAIT selects \(\min\{n_j,n_{js}^t\}\) type-\(j\) prompts from each prefill/decode stage \(s\). If no type is eligible, the server waits for more arrivals. Prompts that are not selected remain in their current queues. A prompt waiting outside the GPU before prefill has no resident KV cache, but any prompt that has already been admitted and is waiting between decode iterations keeps its KV cache on the GPU and continues to count against memory. Algorithm~\ref{alg:wait} and Figure~\ref{fig:wait} describe the policy.

\begin{algorithm}[h]
\caption{WAIT: Waiting for Accumulated Inference Threshold}
\label{alg:wait}
{\small\linespread{1.0}\selectfont
\begin{algorithmic}[1]
\Require Memory $C$, arrival rates $\lambda_j$, thresholds $n_j$ satisfying \eqref{eq:wait_thresholds}, $\forall j \in [m]$
\State Initialize prompt inventory $n_{js} \gets 0$ for all $j \in [m], s \in \{0, 1, \dots, l_j'\}$
\State Initialize event queue with arrival events for each type $j$
\State Set current time $t \gets 0$
\State Set server status to idle
\While{True}
    \State Wait for the next event (arrival or batch completion)
    \If{event is an arrival of type $j$}
        \State Update inventory: $n_{j0} \gets n_{j0} + 1$  \Comment{Add new prompt to waiting queue of prefill phase}
    \ElsIf{event is a batch completion}
        \State Retrieve the selected counts $a_{js}$ stored with the completed batch
        \State Simultaneously advance selected prompts one stage: $n_{js}\gets n_{js}-a_{js}$ and $n_{j,s+1}\gets n_{j,s+1}+a_{js}$ for $s<l_j'$
        \State Clear KV caches for the $a_{j,l_j'}$ prompts completing after stage $l_j'$  \Comment{Free memory for completed prompts}
        \State Set server status to idle
    \EndIf
    \State Set \(J(t)\gets\{j\in[m]: n_{j0}\ge n_j\}\)  \Comment{Eligible types}
    \If{server is idle and \(J(t)\neq\emptyset\)}
        \State For every \(j\in J(t)\) and stage \(s\), set \(a_{js}\gets\min\{n_j,n_{js}\}\)
        \State Form batch \(B\) from all selected prompts and store \(a_{js}\) with \(B\)
        \State Process batch $B$; keep already resident prompts outside the batch waiting with their KV caches retained
        \State Set server status to busy until the completion event for \(B\)
    \EndIf
\EndWhile
\end{algorithmic}
}
\end{algorithm}

\subsection{Asymptotic Guarantees}
\label{subsec:asymptotic_analysis}
\label{subsec:main_theorem}
We analyze WAIT in an asymptotic scaling that averages out stochastic fluctuations while preserving the fixed memory constraint. For \(\zeta\ge 1\), arrival rates scale as \(\lambda_j^{(\zeta)}=\zeta\lambda_j\) and processing times scale as \((d_0^{(\zeta)},d_1^{(\zeta)})=(\zeta^{-1}d_0,\zeta^{-1}d_1)\), with memory capacity \(C\) fixed. For WAIT, the required physical memory is \(M_{\mathrm{req}}^{(\zeta,\pi)}=M^\pi\), which is independent of \(\zeta\) because no downstream boundary buffer is needed. For a policy \(\pi\), let \(N_j^{(\zeta,\pi)}\) be the number of completed decode tokens from type-\(j\) prompts over this horizon, and define
\[
\throughput^{(\zeta, \pi)} = \frac{1}{\zeta T} \sum_{j=1}^m N_j^{(\zeta, \pi)}.
\]
For delay metrics, let \(\Latency_{\rm cal}^{(\zeta,\pi)}\) and \(\TTFT_{\rm cal}^{(\zeta,\pi)}\) denote the finite-horizon censored averages defined in Section~\ref{sec:model}, evaluated in calendar time over this scaled system. The theorem reports the service-normalized delays \(\Latency^{(\zeta,\pi)}:=\zeta\Latency_{\rm cal}^{(\zeta,\pi)}\) and \(\TTFT^{(\zeta,\pi)}:=\zeta\TTFT_{\rm cal}^{(\zeta,\pi)}\), so the effective horizon in the bounds is \(\zeta T\). The benchmark is the fluid throughput \(\throughput^*=\sum_{j=1}^m \lambda_j l_j'\) in \eqref{throughput_fluid_multiple}.

Types with \(\lambda_j=0\) are omitted without loss. Assume \(C \ge M^*\), with \(M^*\) given by \eqref{eq:memory_multi2}. Consider a WAIT policy \(\pi=[n_1,\dots,n_m]\), where \(n_j\) is a positive integer threshold applied at every prefill/decode stage of type \(j\). The threshold vector must satisfy a load-balance condition and the memory condition \(M^*\le M^\pi\le C\):
\begin{equation}
\label{eq:wait_thresholds}
\begin{aligned}
\Delta T(n_{1:m}) &:= d_0 + d_1 M^{\pi} \leq \tfrac{n_j}{\lambda_j},\, \forall j\text{ with }\lambda_j>0,\\
 M^{\pi} &= \sum_{j=1}^m n_j (l_j' + 1) \big(l_j + \tfrac{l_j'}{2}\big),\\
 M^* &\leq M_{\mathrm{req}}^{(\zeta,\pi)}:=M^\pi \leq C
\end{aligned}
\end{equation}
The first line is the arrival-completion balance: during an iteration of length \(\Delta T(n_{1:m})\), the expected number of type-\(j\) arrivals is \(\lambda_j\Delta T(n_{1:m})\), and WAIT serves \(n_j\) type-\(j\) prompts from each stage. Thus \(C\ge M^*\) is the fluid memory-feasibility condition, while the WAIT guarantee below additionally assumes a positive integer threshold vector satisfying \eqref{eq:wait_thresholds}.

\begin{theorem}
\label{thm:wait_heavy_traffic}
Starting from the empty system, let \(\pi=[n_1,\dots,n_m]\) be a vector of positive integer thresholds. If \(\pi\) satisfies \eqref{eq:wait_thresholds}, then
\[
\begin{aligned}
&\throughput^*-\ex{}{\throughput^{(\zeta,\pi)}}=O((\zeta T)^{-\frac{1}{2}}),\\
&\ex{}{\Latency^{(\zeta,\pi)}},\,\ex{}{\TTFT^{(\zeta,\pi)}}=O\prn{(\zeta T)^{\frac{1}{2}}}.
\end{aligned}
\]
Moreover, if \(\Delta T(n_{1:m})<n_j/\lambda_j\) for all \(j\) with \(\lambda_j>0\), then
\[
\begin{aligned}
&\throughput^* - \ex{}{\throughput^{(\zeta,\pi)}} = O(1/(\zeta T)), \\
&\ex{}{\Latency^{(\zeta,\pi)}},\, \ex{}{\TTFT^{(\zeta,\pi)}} = O(1).
\end{aligned}
\]
\end{theorem}

The theorem rests on the control structure created by WAIT. The difficult step is not to prove recurrence after a stable queue has been isolated. Instead, it is to design a batching rule that keeps endogenous KV-cache growth under control before memory overflows. Each admitted prompt creates a KV cache that grows throughout decode. If an unstructured batching rule admits too much work, these caches can exceed memory in later iterations, forcing eviction and restart. The restarted prompts then return as future work and can reduce effective throughput even when the fluid memory requirement fits within capacity. WAIT prevents this failure mode by fixing a target allocation across prefill and decode stages through \(n_{1:m}\). Given this threshold vector, every full batch has memory usage \(M^\pi\) and iteration length \(\Delta T(n_{1:m})\), and the main stochastic component for type \(j\) is the time until its entry queue reaches \(n_j\). The proof uses the threshold structure to define an auxiliary embedded full-threshold process observed at review epochs and then couples it to the event-driven WAIT sample path type by type: although realized WAIT batches may contain different subsets of eligible types and same-type threshold batches can accumulate during a busy period, the embedded process's cumulative type-wise removals are matched by actual type-wise service opportunities with a constant review-interval lag. This coupling turns the original memory-coupled controlled process into threshold queues tied together by the deterministic feasibility condition \(M^\pi\le C\). The complete proof is in Appendix~\ref{appexidx::proof of wait_heavy_traffic}.

The square-root delay bound at the boundary is unavoidable. When the fluid memory requirement exactly fills the available memory, stochastic arrival fluctuations cannot be averaged away by a non-predictive online policy while preserving the boundary service capacity. The next proposition gives the matching lower bound for latency and TTFT.
\begin{proposition}\label{prop::simple random walk, tight bound}
There exists a single-type instance with \( C = M^* \) such that any non-predictive online policy \(\pi\) satisfies, as \(\zeta T\to\infty\),
\[
\mathbb{E}[\Latency^{(\zeta,\pi)}],\quad \mathbb{E}[\TTFT^{(\zeta,\pi)}] = \Omega((\zeta T)^{1/2}).
\]
\end{proposition}
The strict inequalities in Theorem~\ref{thm:wait_heavy_traffic} correspond to slack in these threshold queues. If \(\Delta T(n_{1:m})<n_j/\lambda_j\), the type-\(j\) entry queue has negative drift, and the average latency and TTFT of type-\(j\) prompts remain \(O(1)\).

The policy in this section assumes that output lengths and arrival rates are known when the thresholds are set. In practice, these quantities can be estimated from predictors \citep{fu2025efficient}, but an online serving system may not know a request's decode length at admission. The next section relaxes this information requirement by developing a nested version of WAIT for unknown output lengths.

%% file: unknown_type.tex
\section{Scheduling with Unknown Output Lengths: The Nested WAIT Algorithm}
\label{sec:nested_wait}

The WAIT algorithm assumes that output lengths are known when thresholds are set. Nested WAIT relaxes this requirement by using output-length information only as it is revealed by the prompt's own decode trajectory. The \textit{Nested Waiting for Accumulated Inference Threshold} (Nested WAIT) algorithm organizes decoding into nested segments and controls how prompts advance across those segments. In this setting, endogenous memory growth is harder to control because the scheduler does not know, at admission, which resident prompts will continue into later decode stages. Nested WAIT reduces this unknown-output-length state to segment-boundary queues that track the surviving prompts most likely to create downstream memory pressure.

\subsection{Algorithm Overview}

We first return to the running example to show how unknown output lengths complicate policy design and memory control. The same admitted prompts may either finish at an early decode boundary or continue to later decode stages with larger KV caches, so neither an optimistic nor a conservative admission rule is satisfactory.

\begin{example}[Unknown Output Lengths]
\label{ex:unknown_output}
Continuing Example~\ref{ex:fcfs_cascade}, suppose output lengths are now \emph{unknown} at arrival: each ``Hello'' prompt receives either ``Hi'' (1 token) or ``Hi there!'' (2 tokens). The scheduler faces a dilemma. If it optimistically treats all outputs as short, then capacity \(C=12\) appears to allow 6 prompts, because each would use 2 units after one decode token. If some prompts instead generate two tokens and require 3 units each, the same admission decision can overflow memory and trigger eviction. A conservative rule that treats every prompt as long avoids this overflow, but admits only 4 prompts and wastes capacity whenever most prompts are short.
\end{example}

Nested WAIT avoids both extremes by pooling prompts in early decode stages and separating them only at completion boundaries. Segment 1 serves all prompts until the first possible output length \(l_1'\); prompts that complete there leave the system and release memory, while the remaining prompts enter Segment 2. The same logic repeats across later boundaries. The policy therefore does not reserve memory for long outputs before they are observed, but still lets short prompts complete without waiting behind longer ones.

Formally, let output lengths satisfy $l_1' < l_2' < \dots < l_m'$, where type $j$ has decode length $l_j'$, and set $l_0'=0$. For the main algorithm, all prompts have common prefill length \(l\). Nested WAIT defines \(m\) segments, with segment \(k\) covering the decode work between boundaries \(l_{k-1}'\) and \(l_k'\). The segment receives types \(k,k+1,\dots,m\), whose combined arrival rate is \(\lambda_k+\cdots+\lambda_m\). The threshold vector \(\pi=[n_1,\dots,n_m]\) has two roles: \(n_k\) caps the number of prompts processed per prefill/decode stage in segment \(k\), and it is also the number of prompts that must be available at the segment's entry boundary before that segment begins service.

Output lengths are i.i.d. across prompts according to the output-length distribution specified in Section~\ref{sec:model}, and Nested WAIT is nonanticipating: before a prompt's completion or continuation is realized by service, the scheduler does not use information about that unrealized outcome. Formally, conditional on the scheduler's filtration before a boundary revelation, the unrevealed continuation indicators of the selected prompts are independent Bernoulli random variables with the probabilities implied by that distribution.

For \(k\ge2\), Nested WAIT distinguishes two GPU-resident states around segment \(k\). The \emph{resident boundary queue} \(Q_{k,l_{k-1}'}\) contains prompts that have survived segment \(k-1\), retain their KV caches on the GPU, but have not yet been admitted into segment \(k\). These prompts wait outside segment \(k\) at its entry boundary. Once segment \(k\) is selected for service, up to \(n_k\) prompts are removed from \(Q_{k,l_{k-1}'}\) and admitted into the segment's interior stages \(l_{k-1}'+1,\ldots,l_k'\). Prompts in these interior stages have already entered segment \(k\); if they are not selected in the current batch, they still remain GPU-resident and keep their KV caches. Thus a threshold pause for segment \(k\) does not swap out either the boundary queue or the interior prompts. For a queue state \(Q\), the resident KV-cache usage is
\[
M_{\mathrm{res}}(Q)=\sum_{s=1}^{l_1'}(l+s)Q_{1,s}+\sum_{k=2}^m\sum_{s=l_{k-1}'}^{l_k'}(l+s)Q_{k,s},
\]
which excludes the external prefill queue \(Q_{1,0}\). If memory repair is needed in a physical implementation, Nested WAIT evicts prompts only from resident boundary queues, using LIFO order within this restricted repair set, and evicted prompts discard their KV caches and return to \(Q_{1,0}\). Before launch, the repair set is further restricted to unselected resident boundary prompts in the projected state.
Before launching a batch, Algorithm~\ref{alg:nested_wait} checks the conservative projected state \(Q^+\) obtained by applying the same one-stage transition equations as a batch completion, with unrevealed boundary prompts treated as resident; actual boundary completions can only reduce this projected memory.

\begin{figure}[h]
    \centering
    \includegraphics[width=0.85\linewidth]{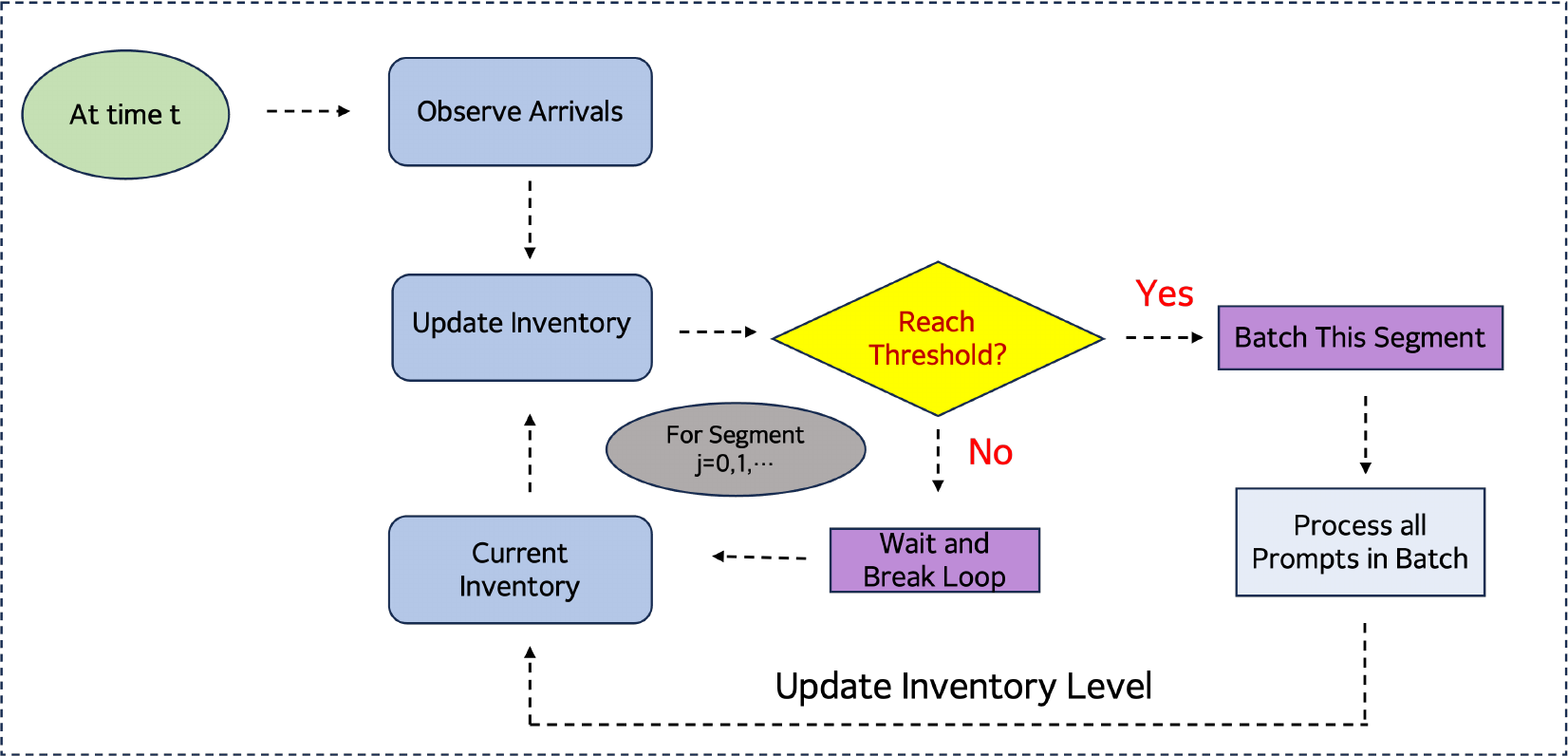}
    \caption{Nested WAIT pipeline. Prompts enter the first segment before their output lengths are known; short prompts complete and exit at early boundaries, while longer prompts advance to later segments where separate thresholds regulate batching.}
    \label{fig:nested_wait}
\end{figure}

Thus, if segment \(k^*\) has fewer than \(n_{k^*}\) prompts in its resident boundary queue, Nested WAIT pauses admission into segment \(k^*\) and all later segments while retaining their KV caches on the GPU. This is a scheduling preemption, not a memory swap: prompts waiting outside a downstream segment in a resident boundary queue, as well as prompts already admitted into a segment's interior stages, continue to count against GPU memory. Only prompts still outside the GPU before prefill, namely the external entry queue \(Q_{1,0}\), have no resident KV cache. Earlier segments \(k<k^*\) can continue batching up to \(n_k\) prompts per prefill/decode stage, so new arrivals can keep moving through the shared early segment. Once enough prompts reach the resident boundary queue of segment \(k^*\), that segment resumes. Algorithm~\ref{alg:nested_wait} gives the main scheduling logic, Appendix~\ref{app:nested_wait_pseudocode} gives the full event-driven implementation, and Figure~\ref{fig:nested_wait} illustrates the resulting pipeline.

Nested WAIT's segmentation rule is based on decode progress. The common-prefill-length convention keeps the memory expressions transparent; Appendix~\ref{sec:extension} discusses heterogeneous prefill lengths and coarser segmentations.

\begin{figure}[h]
    \centering
    \includegraphics[width=0.85\linewidth]{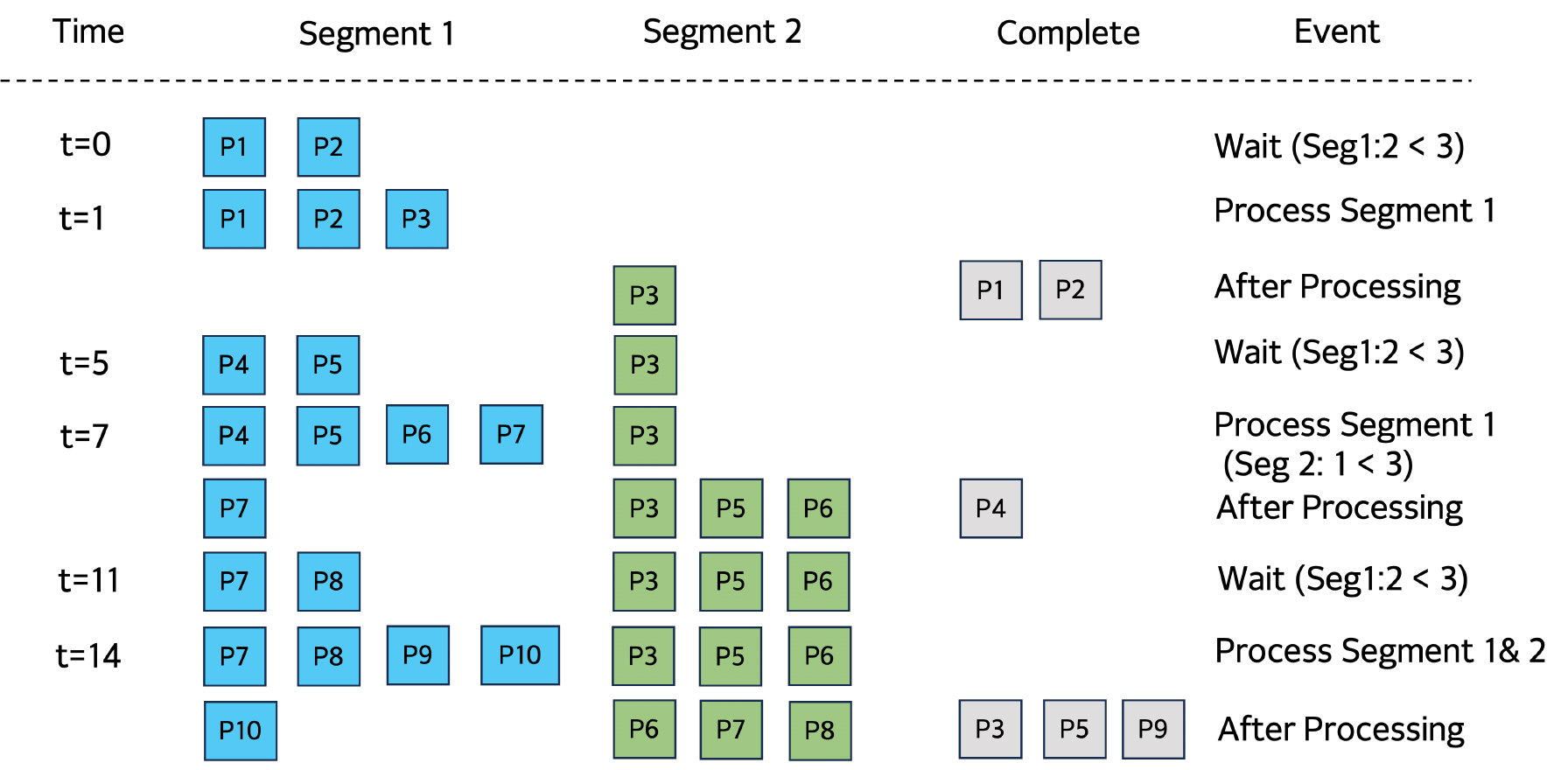}
    \caption{Two-segment Nested WAIT example. Segment 1 batches newly arrived prompts once its threshold is met; Segment 2 starts only after enough prompts have survived the first segment and revealed themselves as longer jobs.}
    \label{fig:nested_wait_example}
\end{figure}

In the two-type instance from Example~\ref{ex:unknown_output}, Nested WAIT uses two segments. Segment 1 processes all newly arrived prompts through prefill/decode stages \(0\) and \(1\) at combined rate \(\lambda_1+\lambda_2\); type-1 prompts then complete, while type-2 prompts wait in Segment 2's resident boundary queue until enough survivors have accumulated to enter Segment 2. Figure~\ref{fig:nested_wait_example} shows this operation over time. Segment 1 waits until enough newly arrived prompts have accumulated, and Segment 2 waits until enough prompts have survived the first segment. When both thresholds are met, the active segments are scheduled together. Thus early segments provide shared batching, while later segments regulate service for prompts that have been identified as longer jobs.

\begin{algorithm}[h]
\caption{Nested WAIT: Nested Waiting for Accumulated Inference Threshold}
\label{alg:nested_wait}
{\small\linespread{1.0}\selectfont
\begin{algorithmic}[1]
\Require Memory capacity \( C \), arrival rates \( \lambda_j \) for \( j \in [m] \), thresholds \( n_k \) for segments \( k \in [m] \)
\Require Output lengths \( l_1' < l_2' < \cdots < l_m' \) for output-length classes \( j \in [m] \)
	    \Statex \Comment{$Q_{1,0}$ is external; for $k\ge2$, $Q_{k,l_{k-1}'}$ waits outside segment $k$ but remains GPU-resident}
	    \Statex \Comment{Prompts in stages $l_{k-1}'+1,\ldots,l_k'$ have already been admitted into segment $k$}
	    \Statex \Comment{Memory repair uses resident boundary queues; before launch, only unselected boundary prompts are eligible}
    \State Initialize all segment-entry, boundary, and interior queues to zero
    \While{the system is active}
        \State Add new arrivals to \(Q_{1,0}\)
        \State On each batch completion, advance selected prompts by one stage
        \State At boundary \(l_{k'}'\), completed prompts release KV cache; if \(k'<m\), survivors join \(Q_{k'+1,l_{k'}'}\)
        \While{\(M_{\mathrm{res}}(Q)>C\) and a resident boundary queue is nonempty}
            \State Evict the most recently admitted prompt from a resident boundary queue and return it to \(Q_{1,0}\)
        \EndWhile
        \State Find the largest prefix \(1,\ldots,k\) whose entry queues satisfy \(Q_{k',l_{k'-1}'}\ge n_{k'}\) for all \(k'\le k\)
        \If{such a prefix exists and the server is idle}
            \State Select up to \(n_{k'}\) prompts from each stage of every segment \(k'\le k\)
            \State Before launch, evict unselected resident boundary prompts in LIFO order until the projected post-iteration state satisfies \(M_{\mathrm{res}}(Q^+)\le C\)
            \State Process the prefix batch; all unselected resident prompts keep their KV caches on GPU
        \EndIf
    \EndWhile
\end{algorithmic}
}
\end{algorithm}

Algorithm~\ref{alg:nested_wait} summarizes the Nested WAIT mechanism. Completion events determine which prompts exit and which continue downstream; thresholds determine when each active segment has enough work to process without sacrificing batching efficiency. The full queue-update and eviction formulas used in implementation are given in Algorithm~\ref{alg:nested_wait_full} of Appendix~\ref{app:nested_wait_pseudocode}.

\subsection{Main Theorem}
\label{subsec:nested_main_theorem}

The analysis uses the same asymptotic scaling and scaled delay metrics as Section~\ref{sec:wait}: arrivals and service speed scale with \(\zeta\), and the effective horizon in the bounds is \(\zeta T\). We suppress \(T\) in the scaled notation; all three metrics are the finite-horizon censored metrics over \([0,T]\) from Section~\ref{sec:model}. As in WAIT, \(M^\pi\) denotes the base threshold memory and is independent of \(\zeta\). The required physical memory \(M_{\mathrm{req}}^{(\zeta,\pi)}\), however, also includes the logarithmic safety buffer in \eqref{eq:nested_wait_memory_total}; this buffer allows Nested WAIT to hedge against the eviction risk created by memory overflow under unknown output lengths. The new difficulty is that downstream segments receive endogenous input. The prompts entering segment \(k\) are exactly those that survived earlier segments, so their arrivals depend on upstream batching, completion events, and the resident KV caches carried across segment boundaries. We study throughput \(\throughput^{(\zeta,\pi)}\), latency \(\Latency^{(\zeta,\pi)}\), and time to first token \(\TTFT^{(\zeta,\pi)}\) under a threshold vector \(\pi=[n_1,\dots,n_m]\).

The need for additional memory is already visible at \(C=M^*\). At this capacity, the fluid equilibrium fits in memory if output lengths are known. A non-predictive policy, however, must admit prompts before knowing whether they will finish at an early boundary or continue downstream. Optimistic admission can overflow memory once prompts reveal longer outputs; conservative admission avoids that risk but leaves capacity unused. The resulting uncertainty can create a persistent throughput loss.

\begin{proposition}
\label{prop:unknown_lower_bound}
When memory capacity is exactly at the known-type fluid memory boundary, \(C=M^*\), there exists a two-output-length instance in which any non-predictive online policy incurs a throughput gap
\[
    \throughput^*-\mathbb E[\throughput^{(T,\pi)}]=\Omega(1)
\]
as \(T\to\infty\).
\end{proposition}

This lower bound shows that uncertainty has a memory cost: \(M^*\) supports the fluid equilibrium with known output lengths, but does not provide room for positive survivor fluctuations at unknown-output boundaries. The proof constructs a two-output-length instance in which the number of prompts continuing past the first boundary has binomial fluctuations (Appendix~\ref{appendix:proof_unknown_lower_bound}). Theorem~\ref{thm:nested_wait_heavy_traffic} shows that Nested WAIT controls these fluctuations with a logarithmic safety buffer. The construction has two parts: choose thresholds that give these boundary queues negative drift, and reserve enough memory for their residual stochastic fluctuations.

The threshold vector must make each segment queue stable while keeping batches large enough to exploit the GPU. Let \(\Delta T_{[1,\dots,m]}(n_1,\dots,n_m)\) denote the iteration time when segment \(k\) contains exactly \(n_k\) prompts at each prefill/decode stage in that segment. The required drift conditions are
\begin{equation}
    \label{eq:nested_wait_thresholds}
    \begin{aligned}
    &\Delta T_{[1,\dots,m]}(n_1, \dots, n_m) <\frac{n_1}{\sum_{j=1}^m \lambda_j}, \\
    &p_k < \frac{n_k}{n_{k-1}} < 1,\quad \forall k \in \{2, \dots, m\},
    \end{aligned}
\end{equation}
where \(p_k = (\sum_{j=k}^m\lambda_j)/(\sum_{j=k-1}^m\lambda_j)<1\) for \(k\ge2\). The first inequality gives negative drift for the entry queue of Segment 1. The second gives negative drift for each downstream boundary queue, since only a \(p_k\) fraction of prompts leaving segment \(k-1\) continue to segment \(k\); the upper bound \(n_k<n_{k-1}\) keeps the boundary queue in the nondegenerate binomial-tail regime used in the martingale-root bound. If \(n_k\ge n_{k-1}\), the downstream boundary queue is deterministically bounded and can be handled separately without this logarithmic tail term. Without loss of generality, the retained downstream segments satisfy \(p_k>0\). For memory accounting, define
\[
\delta_1=l_1'+1,\quad \bar s_1=\frac{l_1'}{2},\qquad
\delta_k=l_k'-l_{k-1}',\quad \bar s_k=\frac{l_{k-1}'+1+l_k'}{2}\quad (k\ge2).
\]
The base memory used by threshold-sized segment batches is
\begin{equation}
    \label{eq:nested_wait_memory}
\begin{aligned}
        M^\pi &= \sum_{k=1}^m n_k (l+\bar s_k)\delta_k .
\end{aligned}
\end{equation}
Here \(\delta_k\) is the number of prefill/decode stages covered by segment \(k\), and \(\bar s_k\) is the average decode progress across those stages. The term \(M^\pi\) is the threshold analogue of the fluid memory requirement \(M^*\): if \(n_k\) is chosen on the scale of the cumulative fluid population that survives to segment \(k\), then \(M^\pi\) is on the same scale as \(M^*\), up to rounding and segment aggregation.

For \(k\ge2\), the resident queue at boundary \(l_{k-1}'\) is measured after the threshold review for segment \(k\). When segment \(k\) is selected, the \(n_k\) boundary prompts removed from this queue are charged to the first interior stage of segment \(k\) in \(M^\pi\), which is a safe upper bound because that stage has one additional decode token. The additional memory buffer in the theorem counts only boundary prompts not selected in the current prefix batch. Appendix~\ref{sec:segment_design}, especially Figures~\ref{fig:fixed_T_200_rate_50_delta_0.1_1e-5}--\ref{fig:fixed_delta_0.001_rate_50_T_20_2000}, gives a numerical decomposition showing this base term relative to the safety buffers. With these definitions, the theorem below gives the performance guarantee.

With unknown output lengths, Nested WAIT chooses thresholds with \(M^\pi\ge M^*\) and adds a safety buffer for the downstream boundary queues created as prompts reveal whether they continue to later segments. The theorem below gives performance guarantees for Nested WAIT on throughput, latency, and avoidance of evictions.
We use a common prefill length \(l\) for a clearer illustration of the memory accounting; heterogeneous prefill lengths use the segment-wise upper-envelope replacement in Appendix~\ref{sec:extension}.

\begin{theorem}
\label{thm:nested_wait_heavy_traffic}
Consider a fixed problem instance: the arrival-rate primitives, output lengths, thresholds \(\pi\), and strict slack margins are held fixed as \(\zeta\) and \(T\) vary. Output lengths are i.i.d.\ across prompts according to the output-length distribution specified in Section~\ref{sec:model}, a prompt's exact output length is revealed only when it completes, and Nested WAIT is nonanticipating with respect to unrevealed output lengths. Fix a constant \(K<\infty\) independent of \(\zeta\) and \(T\). Choose \(\delta\in(0,1)\), possibly depending on \(\zeta\) and \(T\), such that \(\delta\le K(1+\zeta T)^{-1}\). Starting from the empty system, assume \(\Lambda_1:=\sum_{j=1}^m\lambda_j>0\), omit zero-rate downstream segments, and let retained downstream segments satisfy \(0<p_k<n_k/n_{k-1}<1\). Assume thresholds \( \pi = [n_1, \dots, n_m] \) satisfy \eqref{eq:nested_wait_thresholds} with \(M^\pi\ge M^*\), and the physical memory capacity \(C\) satisfies
\begin{equation}
    \label{eq:nested_wait_memory_total}
\begin{aligned}
    C &\ge M_{\mathrm{req}}^{(\zeta,\pi)}
:= M^\pi + \sum_{k=2}^m (l + l_{k-1}') \bigg(n_k + \theta_k^{-1} \ln\Big( \frac{m(1+\zeta T/d_0)}{\delta} \Big)\bigg) \\
&= O\bigg( M^\pi + \sum_{k=2}^m (l + l_{k-1}') \theta_k^{-1}\ln\Big( \frac{m(1+\zeta T/d_0)}{\delta} \Big) \bigg),
\end{aligned}
\end{equation}
where, for each downstream segment \(k\ge2\), $\theta_k>0$ is the unique solution of \(e^{-\theta_k n_k}(1-p_k+p_ke^{\theta_k})^{n_{k-1}}=1\); in particular, $\theta_k \geq 8(n_k - n_{k-1}p_k)/n_{k-1}$. Then Nested WAIT satisfies the performance guarantees
$$
    \begin{aligned}
        &\throughput^* - \mathbb{E}\big[ \throughput^{(\zeta, \pi)} \big] 
   = O\big( (\zeta T)^{-1} \big), \\
&\mathbb{E}\big[ \Latency^{(\zeta, \pi)} \big],\,\mathbb{E}\big[ \TTFT^{(\zeta, \pi)} \big] = O(1).
    \end{aligned}
$$
Moreover, the probability that any memory-overflow-induced eviction occurs over horizon \([0,T]\) is at most \(\delta\).
\end{theorem}

Unknown output lengths make this memory-control problem harder. At admission, the scheduler does not know which prompts will finish early and which will continue into later, more memory-intensive decode stages. Yet endogenous memory growth starts as soon as a prompt's KV cache becomes resident on the GPU. As Example~\ref{ex:unknown_output} illustrates, prompts reveal their output lengths only by either completing at a boundary or continuing beyond it. Completed prompts leave and release memory, while surviving prompts carry larger KV caches downstream. Without segment-level control, these surviving prompts can create late-stage buildup, overflow memory, and trigger evictions. Nested WAIT therefore regulates both admission and advancement through prefill and decode while the output-length mix is still being revealed.

The proof follows the prompts at these segment boundaries, where output-length information is revealed and downstream memory growth must be controlled. Under the i.i.d.\ output-length condition, conditional on a threshold-sized batch leaving segment \(k-1\), the number that continue to segment \(k\) follows binomial thinning with probability \(p_k\). This coupling turns the unknown-output-length process into boundary queues. These queues track how much surviving work is allowed to continue into later segments, which is the part of the memory-growth trace that can cause overflow. In the memory condition \eqref{eq:nested_wait_memory_total}, \(M^\pi\) covers threshold-sized segment batches, while the remaining terms keep retained KV caches at downstream boundaries from causing overflow and eviction. The theorem states the clean case with one boundary per output length. In practice, the number of possible output lengths can be large, and the same construction can be implemented with coarser decode segments that group nearby lengths and maintain thresholds only at the segment boundaries. Appendix~\ref{sec:extension} discusses this general segment design, and Section~\ref{sec:experiments} evaluates it in simulated serving workloads.

%% file: numerical.tex
\section{Numerical Experiments}
\label{sec:experiments}

We evaluate WAIT and Nested WAIT on synthetic Poisson workloads (Section~\ref{sec:exp_synthetic}) and a real workload sampled from lmsys-chat-1m (Section~\ref{sec:exp_real}). We benchmark against vLLM~\citep{kwon2023efficient} and Sarathi~\citep{agrawal2023sarathi}, both greedy FCFS-based serving policies in our experiments; Sarathi differs by splitting long prefills into chunks that can be interleaved with decode work. We use Vidur~\citep{agrawal2024vidur} as the discrete-event simulator, configured for a single NVIDIA A100~80\,GB serving Llama-2-7B. Appendix~\ref{app:additional_experiments} validates Vidur's iteration-time predictions against physical A100 measurements over the profiled range and at an extrapolation point \(B=256\), and also reports end-to-end real-GPU runs of WAIT and Nested WAIT. The workloads below stress decode-side KV-cache growth, the regime targeted by the linear iteration-time approximation in Section~\ref{sec:model}.

With exogenous arrivals at rate $\lambda$, policies in the stable region complete requests at the offered load, so completion-rate differences appear only when a policy approaches or exceeds the arrival rates it can sustain. We use mean end-to-end latency as the primary delay metric in the numerical study. The main figures plot this latency against $\lambda$ and pair it with effective completion rate, which shows whether a policy continues to handle the offered load or instead saturates because of overload or evictions. Each latency point averages $10$ independent simulation replications. Within each replication, latency is averaged over a central measurement window after the initial transient, using the same window across policies. We use $\hat{\lambda}^*$ to mark the observed transition point from the near-overloaded to the overloaded regime: it is the largest reported rate at which latency remains controlled and effective completion rate still tracks the offered load. Above that range, completion rate saturates or declines because the policy suffers overload or evictions, and the paired completion-rate panel records this loss of sustainable service.

\textbf{Hardware.} All experiments use Vidur to simulate a single NVIDIA A100~80\,GB serving Llama-2-7B~\citep{nvidia-a100-specs,huggingface-llama2-docs}. A FP16 KV-cache token for this model occupies \(2\) (key/value) \(\times 32\) layers \(\times 32\) KV heads \(\times 128\) head dimension \(\times 2\) bytes, or \(524{,}288\) bytes (\(0.5\) MiB)~\citep{huggingface-kv-cache-basics}. After model weights and the baseline \(1\%\) memory margin, the baseline-memory runs use a KV-cache cap of approximately \(1.37\times 10^5\) tokens. If resident KV cache exceeds the cap, the server evicts and restarts in-progress requests. The iteration-time model is Equation~\eqref{eq:time_consump}; its A100 validation is reported in Appendix~\ref{app:gpu_validation}.

\textbf{Baseline configurations (vLLM, Sarathi).} Both baselines use their default Vidur configurations and greedily admit arrivals in FCFS order. They also use the same local capacity-reservation rule: before admitting a new request, the scheduler requires a small memory buffer ($1\%$ of KV-cache capacity) to remain unused. This rule protects against admitting a request when the server is already essentially full, but it is not a dynamic limit on the total number of jobs in service. Once admitted requests continue decoding, their KV caches can still grow beyond the available capacity, in which case the scheduler evicts and restarts an in-progress request. The difference between the baselines is in how prefill is batched. vLLM processes a newly admitted request's full prefill in one iteration. Sarathi bounds the prefill work per iteration to chunks of at most $512$ tokens, allowing partial prefills to be interleaved with decode iterations. This chunked-prefill rule reduces head-of-line blocking from long prefills, but it does not control the cumulative decode-side population. Thus neither baseline enforces the load-balancing threshold used by WAIT.

\textbf{WAIT and Nested WAIT settings.} In the experiments below, the main parameter for WAIT and Nested WAIT is a system-wide batch-size cap $\mathrm{tl}$, calibrated on the common finite parameter grid \(\{20,30,\ldots,300\}\). The cap controls how many requests can be served concurrently across prefill and decode stages, but it is not itself one of the per-stage scheduling thresholds. In a single-type WAIT run, once $\mathrm{tl}$ is chosen, the integer per-stage thresholds are computed by distributing this cap across the prefill/decode stages of the workload. For Nested WAIT, we first partition the decode horizon into \(L\) ordered segments. Let \(\Delta l_k'\) be the number of decode stages in segment \(k\), and let \(\lambda_k'\) be the arrival rate of requests whose final decode length lies in that segment. A request reaches segment \(k{+}1\) only if its decode length exceeds the end of segment \(k\), so the continuation probability from segment \(k\) to segment \(k{+}1\) is \(p_k=(\sum_{r=k+1}^L\lambda_r')/(\sum_{r=k}^L\lambda_r')\). Given the margin parameter \(\eta\), we set \(q_k=\min\{p_k+\eta,0.99\}\). Once \(\mathrm{tl}\), \(L\), and \(\eta\) are fixed, the real-valued per-stage thresholds are computed as \(n_k=n_1\prod_{r<k}q_r\), where \(n_1\) solves \(\sum_{k=1}^L \Delta l_k' n_k=\mathrm{tl}\). The segment-level cap \(B_k=\Delta l_k'n_k\) is then rounded to an integer and spread as evenly as possible across the stages in that segment. Thus, \(\mathrm{tl}\) fixes the overall threshold scale, while \(L\), \(\eta\), and the continuation probabilities determine how that scale is allocated across decode segments. The workload-specific subsections describe the corresponding threshold and segment design. The calibration uses only the workload distribution and the offered arrival rate; the online scheduler does not use realized future arrivals or unrevealed final output lengths. Under the baseline settings used for the arrival-rate figures, WAIT and Nested WAIT do not trigger eviction.

\subsection{Synthetic Workloads}
\label{sec:exp_synthetic}

We consider three synthetic Poisson workloads. The single-type workload \texttt{p512d20} has prefill length $512$ and decode length $20$, isolating the effect of WAIT's threshold mechanism (Algorithm~\ref{alg:wait}). The two-type workload $(0.7\,\texttt{p512d20},\,0.3\,\texttt{p512d50})$, meaning that each arrival is \texttt{p512d20} with probability $0.7$ and \texttt{p512d50} with probability $0.3$, tests the segment design of Nested WAIT (Algorithm~\ref{alg:nested_wait}) on prompts that share the prefill profile but differ in decode length. The long-decode workload \texttt{p512d1000} keeps the same prefill length but increases the decode length to $1000$, allowing us to examine the same threshold mechanism when KV cache growth is much more pronounced.

\begin{figure}[htbp]
    \centering
    \includegraphics[width=0.76\textwidth]{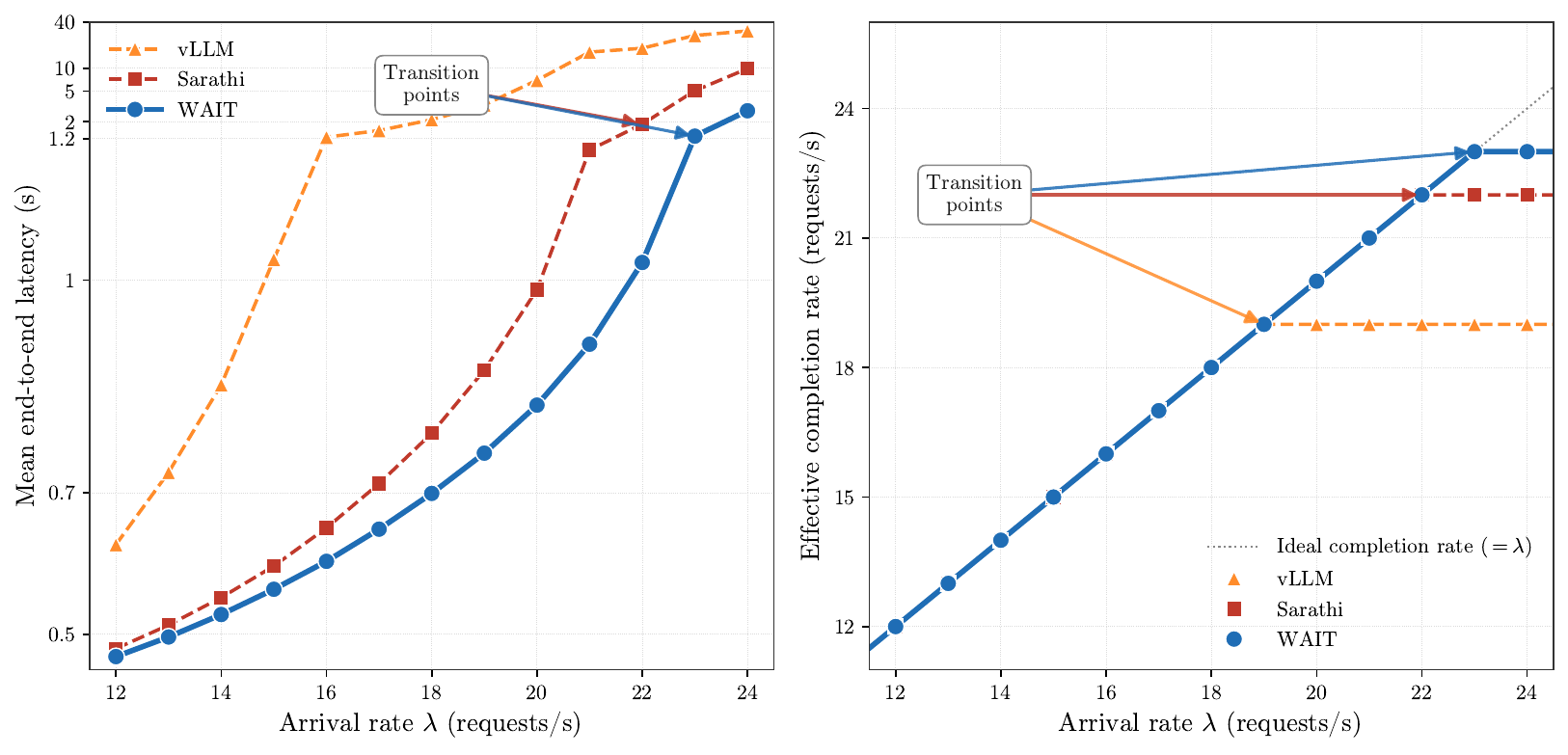}
    \caption{Single-type workload (\texttt{p512d20}): mean end-to-end latency (left) and effective completion rate (right) versus $\lambda$. Arrows mark the transition from near-overloaded to overloaded operation; the dotted line is the ideal completion rate \(\lambda\).}
    \label{fig:single_type_rate}
\end{figure}

\paragraph{Single-type workload (WAIT).}
For this workload, WAIT converts the selected $\mathrm{tl}$ into the corresponding per-stage limits. Figure~\ref{fig:single_type_rate} shows mean end-to-end latency (left, on a stretched-symlog scale that expands the stable region) and effective completion rate (right), i.e., completed requests per second net of evictions, versus arrival rate $\lambda$. Because all requests are homogeneous, this workload highlights the effect of threshold-based admission without the additional segment structure used in Nested WAIT. WAIT keeps latency controlled up to $\lambda \approx 23$, whereas Sarathi and vLLM lose stability near $\lambda \approx 22$ and $\lambda \approx 19$, respectively. At low load the three policies are close, but the gap widens rapidly as $\lambda$ approaches the boundary: WAIT attains the smallest mean latency at every tested rate in $\lambda \in \{12, \dots, 24\}$, with the gap to Sarathi growing from ${\sim}2\%$ at $\lambda = 12$ to over $70\%$ at $\lambda = 23$. Because there is only one request type, the gain does not come from type separation; it comes from threshold-based admission itself, which keeps the batch composition closer to the balanced prefill/decode composition and delays the onset of instability. The right panel shows the same insight in completion-rate terms: WAIT follows the ideal $\lambda$-line up to a higher arrival rate than Sarathi and vLLM.

\begin{figure}[htbp]
    \centering
    \includegraphics[width=0.76\textwidth]{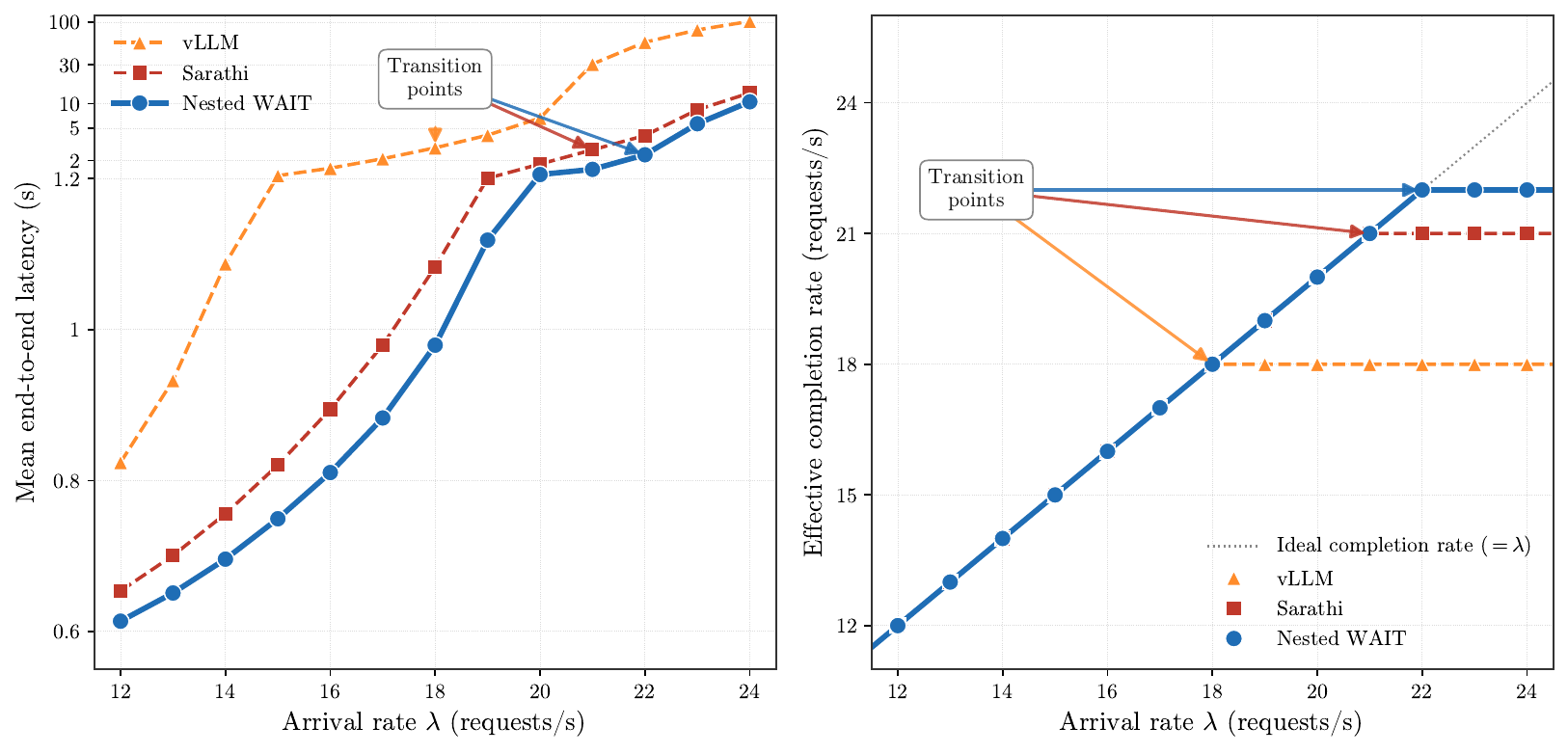}
    \caption{Two-type workload $(0.7\,\texttt{p512d20},\,0.3\,\texttt{p512d50})$: mean end-to-end latency (left) and effective completion rate (right) versus $\lambda$. The transition ordering is vLLM ($\hat{\lambda}^* = 18$), Sarathi ($\hat{\lambda}^* = 21$), and Nested WAIT ($\hat{\lambda}^* = 22$).}
    \label{fig:multi_type_rate}
\end{figure}

\paragraph{Two-type workload (Nested WAIT).}
For the two-type workload, Nested WAIT uses the same system-wide batch-size cap $\mathrm{tl}$ and two decode-stage segments, $1$--$20$ and $21$--$50$. The second segment receives the prompts that survive the first $20$ decode stages, so its continuation ratio is $0.3$. Thus the allocation rule above gives real-valued per-stage thresholds in the ratio $n_2/n_1=0.3$ and solves $20n_1+30n_2=\mathrm{tl}$ before rounding the corresponding segment-level caps to integers; for example, $\mathrm{tl}=30$ gives segment-level caps approximately $21$ and $9$. Figure~\ref{fig:multi_type_rate} shows the same comparison for the two-type workload $(0.7\,\texttt{p512d20},\,0.3\,\texttt{p512d50})$. The stability ordering is again vLLM first ($\hat{\lambda}^* \approx 18$), Sarathi second ($\hat{\lambda}^* \approx 21$), and Nested WAIT last ($\hat{\lambda}^* \approx 22$). Nested WAIT attains the lowest mean latency over these tested rates, with the gap to Sarathi growing from ${\sim}6\%$ at low load to over $40\%$ near the boundary. The mechanism also differs from the single-type case. Here the two request classes have the same prefill length and differ only in decode length, so the main source of congestion is the accumulation of the long-decode minority in later decode stages. Nested WAIT's segment thresholds keep that downstream inventory from expanding too far, which prevents those delays from propagating to the short-decode majority. The resulting gain is therefore not only at entry: by controlling downstream buildup, Nested WAIT preserves low latency for the short jobs and enlarges the stable operating region beyond what a single global cap can achieve.

\begin{figure}[htbp]
    \centering
    \includegraphics[width=0.74\textwidth]{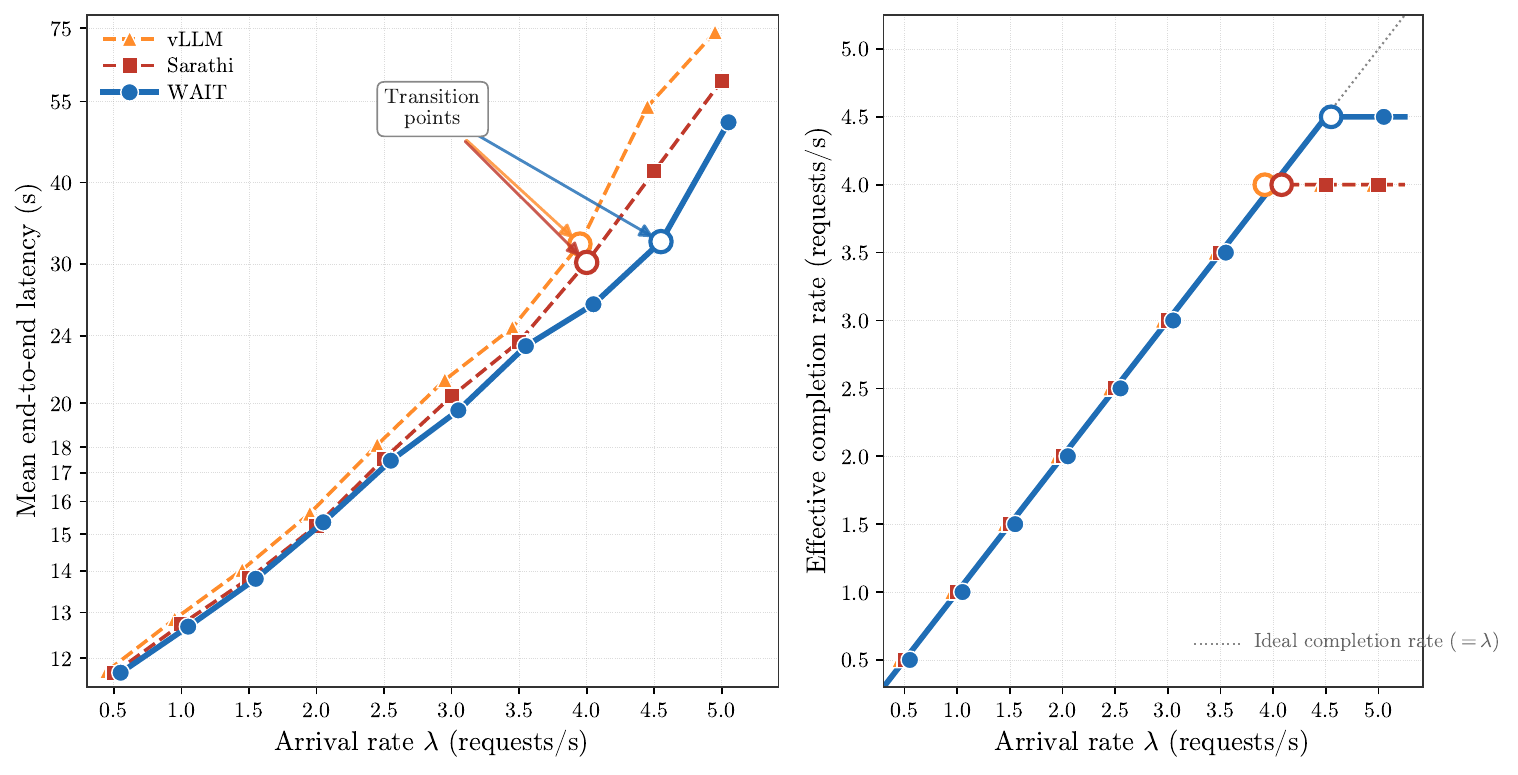}
    \caption{Long-decode workload (\texttt{p512d1000}): mean end-to-end latency (left) and effective completion rate (right) versus $\lambda$. The main separation emerges from $\lambda \approx 3.0$ onward; near the boundary, WAIT attains the lowest latency and highest completion rate.}
    \label{fig:long_decode}
\end{figure}

\paragraph{Long-decode regime.}
We also study the long-decode workload \texttt{p512d1000} with Poisson arrivals at $\lambda = 0.5, 1.0, \ldots, 5.0$. Relative to the single-type workload \texttt{p512d20}, each request now occupies about $30\times$ more KV cache, so all three policies lose stability at much smaller arrival rates. Figure~\ref{fig:long_decode} shows that at low rates $\lambda \leq 2.5$ the three policies are nearly indistinguishable, but near the boundary WAIT again has the lowest latency: at $\lambda = 4.5$ it attains $32.3$\,s versus $41.7$\,s for Sarathi and $53.8$\,s for vLLM. The intuition is that long-decode requests stay in the system for many more iterations, so excess admission persists longer and makes it harder to keep arrivals and completions in balance. WAIT performs better in this regime because it keeps the in-system inventory closer to that balance point. The same pattern is already visible at $\lambda = 4.0$, where the baseline-memory setting gives $26.4$\,s for WAIT versus $30.2$\,s for Sarathi. Table~\ref{tab:eviction} reports the accompanying eviction evidence at the same arrival rate: Sarathi incurs a $2.88\%$ restart rate, whereas WAIT still avoids eviction altogether. Under the server's last-in-first-out eviction rule, each evicted request loses its partial decode progress and begins again from prefill, so the restart gap is another manifestation of the same imbalance.

\begin{table}[h]
\centering
\small
\begin{tabular}{lcc}
\toprule
Metric & Sarathi & WAIT (ours) \\
\midrule
Mean end-to-end latency ($\lambda = 4.0$) & $30.2$\,s & $26.4$\,s \\
Eviction-induced restart rate ($\lambda = 4.0$) & $2.88\%$ & $0\%$ \\
\bottomrule
\end{tabular}
\caption{Eviction diagnostic for \texttt{p512d1000} at $\lambda = 4.0$.}
\label{tab:eviction}
\end{table}

\begin{figure}[htbp]
    \centering
    \includegraphics[width=0.76\textwidth]{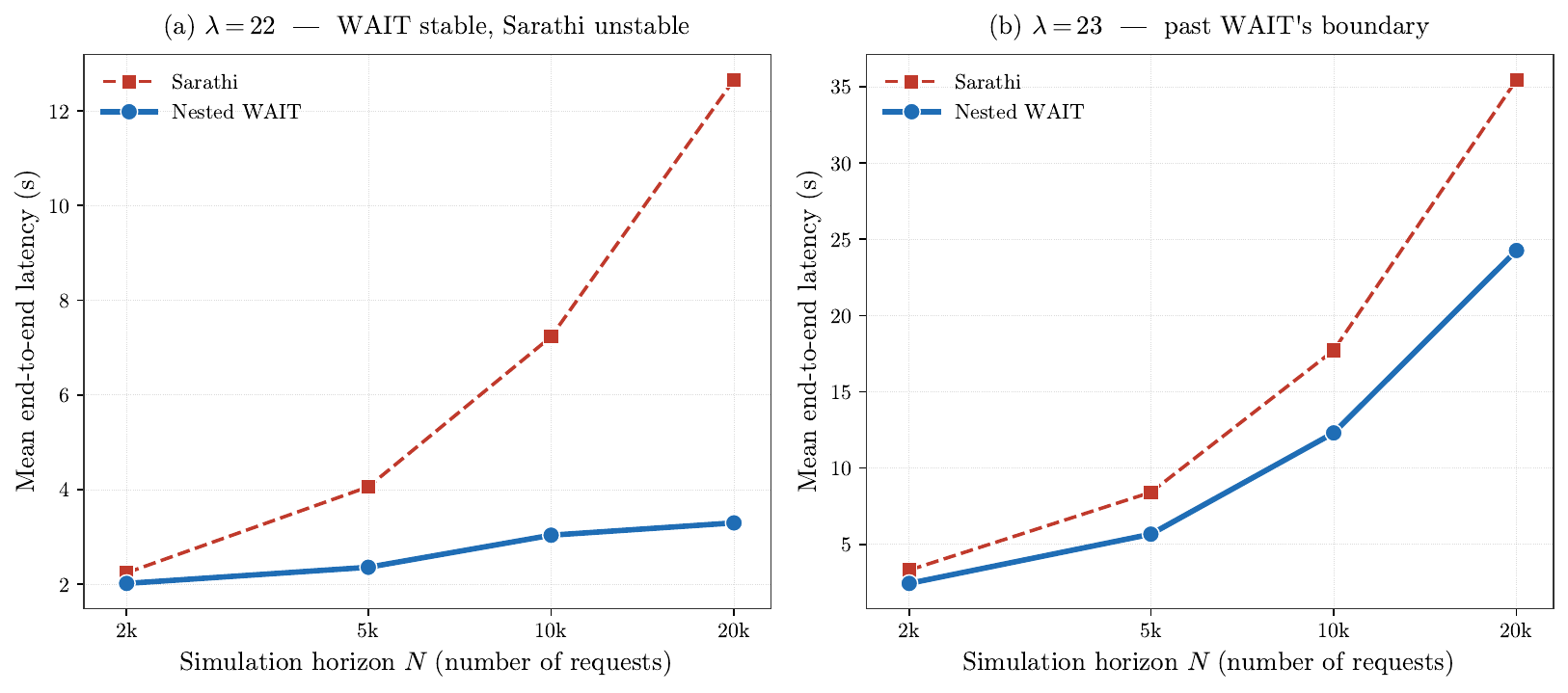}
    \caption{Long-horizon check near the transition on $(0.7\,\texttt{p512d20},\,0.3\,\texttt{p512d50})$. At $\lambda = 22$, Nested WAIT approaches a finite latency plateau; at $\lambda = 23$, both policies show continued latency growth, but Nested WAIT remains lower-latency.}
    \label{fig:fin_horizon}
\end{figure}

\paragraph{Long-horizon check near the transition.}
To check the estimated transition, we vary the simulation horizon at the two arrival rates nearest Nested WAIT's transition. Figure~\ref{fig:fin_horizon} plots mean latency on $(0.7\,\texttt{p512d20},\,0.3\,\texttt{p512d50})$ for $\lambda \in \{22, 23\}$ as $N$ increases from $2{,}000$ to $20{,}000$. At $\lambda = 22$, Nested WAIT approaches a finite latency plateau, whereas Sarathi's mean latency grows rapidly over the tested horizons. At $\lambda = 23$, both policies exhibit continued latency growth, but Nested WAIT remains $30\%$--$50\%$ below Sarathi at every horizon we consider. These comparisons are consistent with the observed transition from near-overloaded to overloaded operation lying between the two rates.

\subsection{Application to the \texttt{lmsys-chat-1m} Dataset}
\label{sec:exp_real}

We next evaluate the policies on \texttt{lmsys-chat-1m}, a real dataset with heterogeneous prompt and response lengths, where decode lengths are unknown at admission. This empirical setting lets us assess the robustness of Nested WAIT's state-dependent admission rule under a heterogeneous workload distribution, and characterize how the induced capacity allocation varies across prefill and decode stages.

\begin{figure}[ht]
    \centering
    \includegraphics[width=0.66\textwidth]{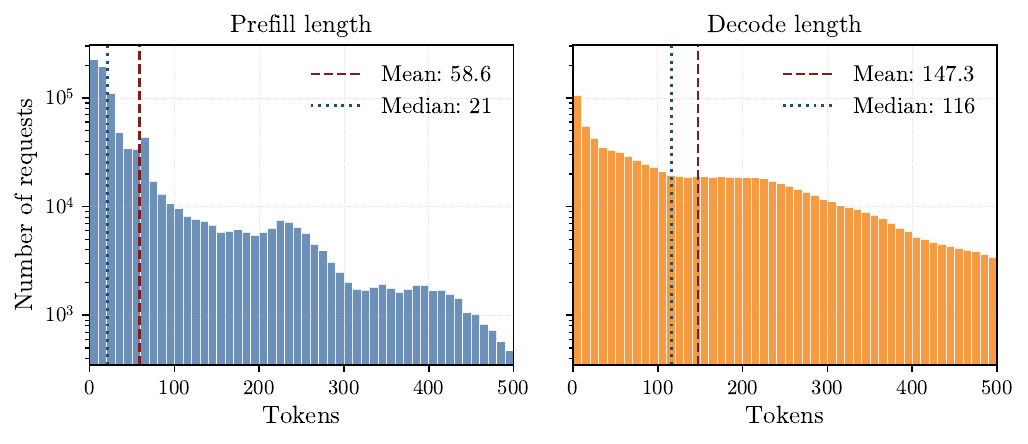}
    \caption{Filtered \texttt{lmsys-chat-1m} prefill and decode length distributions. Dashed lines mark means and medians.}
    \label{fig:lmsys_distribution}
\end{figure}

\paragraph{Dataset.}
The dataset contains approximately $1$ million human--LLM conversations; we treat each user message as the prompt (prefill) and the corresponding assistant response as the generated output (decode). Figure~\ref{fig:lmsys_distribution} shows the prefill-length and decode-length distributions after filtering to requests with both lengths below $500$ tokens, which is the workload population used for sampling. Prefill lengths have mean $58.6$ tokens and median $21$, while decode lengths have mean $147.3$ tokens and median $116$. Short responses remain common: about $31\%$ of prompts decode for at most $50$ tokens, whereas about $14\%$ exceed $300$. For simulation, we sample $5{,}000$ requests from this filtered workload population. Arrivals follow a Poisson process with rate $\lambda$, measured in requests per second. All runs use the same simulator, hardware, and baseline settings as in Section~\ref{sec:exp_synthetic}.

\paragraph{Nested WAIT calibration on the real dataset.}
For the real-data runs, Nested WAIT applies the segment construction described above to the empirical decode-length distribution. The decode horizon \(0\)--\(500\) is divided into \(L\) equal-width segments, and a request is counted against the threshold of the segment containing its current decode stage. It first consumes capacity in the head segment and enters downstream segments only if it continues decoding, so the policy allocates admission capacity across realized continuation stages rather than making scheduling decisions from predicted final response lengths. All real-data arrival-rate runs use the baseline KV-cache cap of approximately \(1.37\times 10^5\) tokens described above. The cap \(\mathrm{tl}\) should not be read as allowing \(\mathrm{tl}\) longest-context requests to reside on the GPU simultaneously: resident KV-cache use depends on the realized prompt lengths and decode stages, and exceeding the memory cap triggers eviction and restart. In the experiments, we set \(\eta=0.05\), choose \(L\) from \(\{1,2,3,4,5,10,20\}\), and use the lowest-latency configuration on this parameter grid for each offered arrival rate. The performance is not monotone in either parameter: a cap that is too small leaves batching and downstream control underused, whereas a cap that is too large can push resident KV-cache usage toward overflow. Similarly, too few segments blur meaningful decode-stage differences, while too many segments fragment the segment-level caps and make prompts wait at segment thresholds too often. The selected setting is therefore the best tradeoff on this parameter grid for the workload and arrival rate.

To relate the real-data arrival rates to the fluid memory condition, Table~\ref{tab:lmsys_mstar} reports \(M^*(\lambda)\) for the same real-data sample. We compute these values by estimating the service-time parameters from our experiments and substituting the empirical prefill/decode length distribution into the fluid memory formula~\eqref{eq:memory_multi2}. The calculation places the fluid memory crossing \(M^*(\lambda)=C\) between \(\lambda=70\) and \(\lambda=80\), at approximately \(74.1\) queries/s. This is consistent with the range where the real-data latency and completion-rate curves begin moving from near-overloaded to overloaded operation; the remaining gap is expected from integer threshold rounding, finite-horizon stochastic variation, and the discrete segment calibration used by Nested WAIT.

\begin{table}[htbp]
\centering
\small
\begin{tabular}{rccc}
\toprule
Arrival rate \(\lambda\) & \(M^*(\lambda)\) (KV tokens) & \(M^*(\lambda)/C\) & Fluid stable region \\
\midrule
50 & \(34{,}377\)  & \(0.25\) & inside \\
60 & \(55{,}758\)  & \(0.41\) & inside \\
70 & \(100{,}327\) & \(0.73\) & inside \\
80 & \(250{,}509\) & \(1.83\) & outside \\
\bottomrule
\end{tabular}
\caption{Fluid memory requirement for the real dataset. The capacity is \(C\approx 1.37\times 10^5\) KV-cache tokens.}
\label{tab:lmsys_mstar}
\end{table}

\begin{figure}[htbp]
    \centering
    \includegraphics[width=0.68\textwidth]{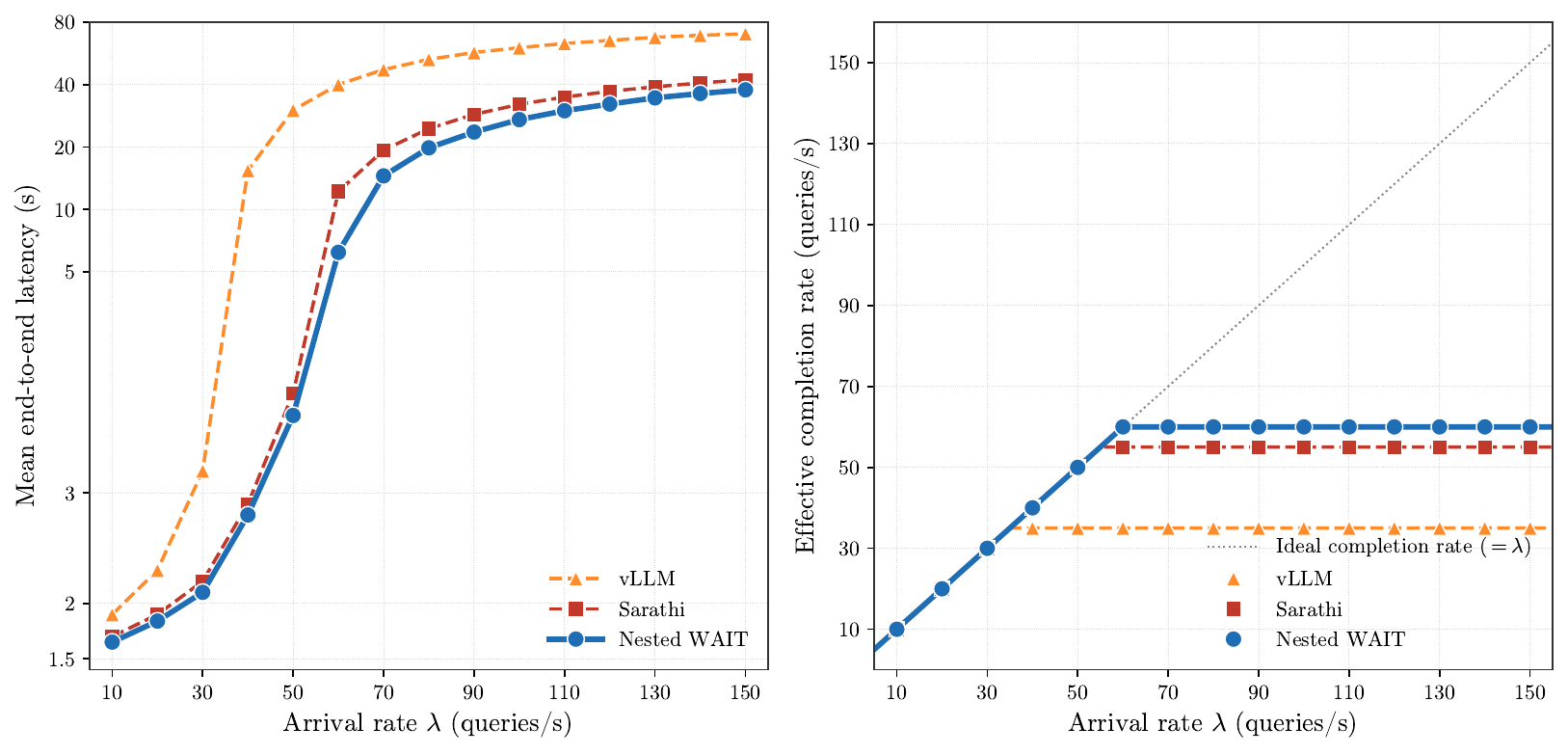}
    \caption{Real-data experiment on \texttt{lmsys-chat-1m} ($5{,}000$ requests): mean end-to-end latency (left) and effective completion rate (right) versus $\lambda$. Nested WAIT has a larger latency gap near overload.}
    \label{fig:lmsys_rate}
\end{figure}

\paragraph{Results.}
Figure~\ref{fig:lmsys_rate} varies the arrival rate $\lambda$ from $10$ to $150$ in steps of $10$. The plotted comparison follows the same qualitative ordering as the synthetic experiments. The baseline schedulers are competitive when the workload is light, but their latencies increase more quickly as the system approaches and then exceeds the range in which completions can keep pace with arrivals. Nested WAIT is most clearly separated from the baselines in these near-overloaded and overloaded regimes. This pattern is consistent with the heterogeneous decode distribution: many requests finish in the early decode stages, while a smaller group remains active for hundreds of generated tokens. Nested WAIT's nested thresholds control how many requests may continue across successive decode-stage ranges, allowing short responses to clear early while limiting the population in later decode stages that consumes KV-cache capacity.

%% file: conclusion.tex
\section{Conclusion and Future Directions}

This paper studies online scheduling for LLM inference, where each request's memory consumption grows during decoding and exceeding capacity triggers evictions that can destabilize the system even when the fluid equilibrium fits in memory. We develop a multi-stage model with endogenous memory growth and a fluid approximation that identifies both the fluid stability region and the memory requirement \(M^*\). Guided by this equilibrium, (Nested) WAIT keeps the stochastic system close to the fluid composition across prefill and decode stages. Its threshold mechanism performs a dimensionality reduction by inducing threshold and boundary queues that capture the memory-coupled state created by endogenous KV-cache growth. Under the linear multi-stage model and the stated memory conditions, WAIT approximates the fluid benchmark asymptotically, with latency and TTFT guarantees; for unknown output lengths, Nested WAIT provides guarantees on throughput, latency, and eviction avoidance under its safety buffer. In Vidur simulations configured for Llama-2-7B on an NVIDIA A100 GPU, they expand the empirically observed stable operating range and reduce latency most clearly near capacity and under overload compared with the tested vLLM and Sarathi configurations. Supplemental real-GPU validation in the appendix provides implementation evidence as well as support for the simulator's accuracy.

Several questions remain open. First, our threshold design uses the prompt-type arrival distribution; adaptive or robust variants that learn this distribution online would extend the algorithm's applicability. Second, practical deployments face time-varying demand. Theorem~\ref{thm:nested_wait_heavy_traffic_time_varying} covers bounded window loads and bounded first-segment waiting times, but sharper demand spikes, long low-arrival intervals, and timeout rules require further analysis, especially for threshold re-tuning and the Nested WAIT safety buffer.

Finally, multi-GPU deployments add inter-device communication costs and parallelization choices. This regime is especially relevant for mixture-of-experts architectures \citep{shazeer2017outrageously, fedus2022switch, jiang2024mixtral}, where token-level routing creates memory and compute imbalance across GPUs. Incorporating communication and routing-aware threshold coordination into the iteration-time model is a natural open direction.

%% file: extension.tex
\section{Extensions}\label{sec:extension}

The WAIT and Nested WAIT algorithms assume constant arrival rates and a fixed number of segments matching the number of prompt types. In practice, however, LLM inference systems face two additional challenges: (i) \textbf{time-varying workloads}, where arrival rates fluctuate over time; (ii) \textbf{numerous prompt types}, where maintaining separate thresholds for hundreds of output lengths becomes infeasible. We address these challenges through two extensions that preserve the core algorithmic principles---eviction prevention and approaching load balance---while adapting the threshold construction to more realistic deployment settings.

First, we extend Nested WAIT to handle time-varying arrival rates $\lambda_j^t$ by replacing constant-rate threshold conditions with accumulated arrival constraints over each iteration window. This gives a conservative finite-horizon guarantee under bounded window loads. The delay bound additionally requires that the first threshold can be collected on the service-normalized time scale, which rules out arbitrarily long low-arrival intervals (Section~\ref{sec:time_varying}). Second, we introduce a \textbf{segment-wise design} that groups many prompt types into a smaller number of decode-stage segments, reducing algorithmic complexity while maintaining the same threshold structure (Section~\ref{sec:segment_design}).

\subsection{Time-Varying Arrival Rates}\label{sec:time_varying}

Time-varying arrival rates capture moderate changes in workload intensity, such as predictable daily variation or slowly changing demand. The threshold mechanism remains valid when the accumulated arrivals during each service window stay below the corresponding completion capacity. Formally, we replace constant rates $\lambda_j$ with deterministic time-dependent intensities $\lambda_j^t$, assumed continuous and uniformly bounded by $\lambda_{\max} < \infty$ to prevent unbounded rates. Type arrivals are independent nonhomogeneous Poisson processes with these intensities. Hidden output-length labels are independent marks conditional on the arrival times and on the scheduler's pre-boundary filtration.

We analyze the finite-horizon guarantee of the Nested WAIT algorithm under time-varying arrival rates, denoted $\lambda_j^t$ for prompt type $j \in [m]$ at physical time $t \in [0, T]$.

For a service-normalized window \(h\), define the expected number of type-\(j\) arrivals in the corresponding physical interval by
\[
    \mathcal A_j^{(\zeta)}(t,h)
    :=
    \zeta\int_t^{\min\{T,t+h/\zeta\}}\lambda_j^u\,du .
\]
The truncation only affects the final partial window and can only reduce the number of arrivals.
Recall from Equation \eqref{eq:wait_thresholds} that
\[
    \Delta^\pi:=\Delta T_{[1,2,\dots,m]}(n_1,\cdots,n_m)
\]
represents the service-normalized iteration cost when processing exactly $n_k$ prompts at each stage in segment $k$, for all $k \in [m]$, as in \eqref{eq:nested_wait_thresholds}, where $n_k$ is the threshold defined in Algorithm \ref{alg:nested_wait}.

For the Nested WAIT policy defined by $\pi = [n_1, \dots, n_m]$, we impose the following uniform slack conditions for all $t\in [0, T]$, $0<h\leq \Delta^\pi$, and \(k=2,\dots,m\), analogous to \eqref{eq:nested_wait_thresholds}:
\begin{equation}
    \label{eq:nested_wait_thresholds_time_varying}
\begin{aligned}
        & \sum_{j=1}^m \mathcal A_j^{(\zeta)}(t,h)
        \leq
        \sum_{j=1}^m \mathcal A_j^{(\zeta)}(t,\Delta^\pi)
        \le \Lambda^\pi < n_1,\qquad
        \Lambda^\pi:=\sup_{u\in[0,T]}\sum_{j=1}^m \mathcal A_j^{(\zeta)}(u,\Delta^\pi), \\
        & p_k^{(\zeta)}(t,h) \le p_k^* < n_k/n_{k-1} < 1,\quad k=2,\dots,m.
\end{aligned}
\end{equation}
where \(p_k^{(\zeta)}(t,h)=(\sum_{j=k}^m\mathcal A_j^{(\zeta)}(t,h))/(\sum_{j=k-1}^m\mathcal A_j^{(\zeta)}(t,h))\) when the denominator is positive, and \(p_k^{(\zeta)}(t,h)=0\) otherwise. The time-varying fluid benchmark is
\[
    \throughput_T^*
    :=
    \frac{1}{T}\int_0^T\sum_{j=1}^m\lambda_j^t l_j'\,dt .
\]
For the downstream safety buffers, set
\[
    p_k^*
    :=
    \sup_{\substack{t\in[0,T],\,0<h\le \Delta^\pi\\
    \sum_{j=k-1}^m\mathcal A_j^{(\zeta)}(t,h)>0}}
    \frac{\sum_{j=k}^m\mathcal A_j^{(\zeta)}(t,h)}
    {\sum_{j=k-1}^m\mathcal A_j^{(\zeta)}(t,h)},\qquad k\ge2,
\]
and set \(p_k^*=0\) if the denominator is zero for all such windows. If a threshold batch contains prompts accumulated over a longer interval, partition that interval into service-normalized subwindows of length at most \(\Delta^\pi\). The continuation ratio over the accumulated set is a denominator-weighted average of the positive-denominator subwindow ratios, and is therefore also bounded by \(p_k^*\).
Without loss of generality, the retained downstream segments satisfy \(p_k^*>0\).
\paragraph{First-segment waiting.}
The window-load conditions above control overload and downstream boundary queues. A delay guarantee also needs a mild lower-side condition: if arrivals nearly stop, a subthreshold first-segment queue can wait for a long physical time before the next batch forms. Let \(\tau_1^{(\zeta)}(t)\) be the physical time needed, starting at time \(t\), for the scaled aggregate external arrival process, continued with the same local intensity model beyond \(t\), to generate \(n_1\) arrivals. For the \(O(1)\) service-normalized delay bound below, assume there is a constant \(H<\infty\), independent of \((\zeta,T)\), such that
\[
    \sup_{t\in[0,T]}\zeta\,\mathbb E[\tau_1^{(\zeta)}(t)]\le H .
\]
This condition is implied, for example, by a uniform lower bound \(\sum_{j=1}^m\lambda_j^t\ge\lambda_{\min}>0\). Without such a condition, the throughput and memory statements remain meaningful, but the service-normalized delay can be dominated by waiting through a long low-arrival interval.
The following theorem extends Theorem~\ref{thm:nested_wait_heavy_traffic} to time-varying arrival rates.

\begin{theorem}
\label{thm:nested_wait_heavy_traffic_time_varying}
Fix a constant \(K<\infty\) independent of \(\zeta\) and \(T\). Starting from the empty system, choose \(\delta\in(0,1)\), possibly depending on \(\zeta\) and \(T\), such that \(\delta\le K(1+\zeta T)^{-1}\). Suppose the thresholds $\pi = [n_1, \dots, n_m]$ satisfy \eqref{eq:nested_wait_thresholds_time_varying}, and the physical memory capacity \(C\) fulfills:
\[
\begin{aligned}
	C &\ge M_{\mathrm{req},\mathrm{tv}}^{(\zeta,\pi)}
	:= M^{\pi} + \sum_{k=2}^m (l + l_{k-1}') \left(n_k + \theta_k^{-1} \ln\!\left( \tfrac{m(1+\zeta T/d_0)}{\delta} \right)\right) \\
	    &= O\!\left( M^\pi + \sum_{k=2}^m \theta_k^{-1}  (l + l_{k-1}') \ln\!\left( \tfrac{m(1+\zeta T/d_0)}{\delta} \right) \right),
\end{aligned}
\]
where the memory usage $M^\pi$ is defined as \eqref{eq:nested_wait_memory} and, for each downstream segment \(k\ge2\), \(\theta_k>0\) is the unique solution of
\[
    e^{-\theta_k n_k}(1-p_k^*+p_k^*e^{\theta_k})^{n_{k-1}}=1,
\]
with lower bound \(\theta_k\ge 8(n_k-n_{k-1}p_k^*)/n_{k-1}\).
All quantities \(p_k^*\), \(\theta_k\), and \(M_{\mathrm{req},\mathrm{tv}}^{(\zeta,\pi)}\) are evaluated for the given scaled finite-horizon system \((\zeta,T)\).
The constants in the \(O(\cdot)\) bounds may depend on the fixed thresholds, output lengths, and the uniform slack margins in \eqref{eq:nested_wait_thresholds_time_varying}.
Then finite-memory Nested WAIT satisfies
$$
    \begin{aligned}
        &\throughput_T^* - \mathbb{E}\left[ \throughput^{(\zeta, \pi)} \right]
   = O\left( (\zeta T)^{-1} \right).
    \end{aligned}
$$
Moreover, the probability that any memory-overflow-induced eviction occurs over the time horizon $[0,T]$ is at most \(\delta\).
If the first-segment waiting condition above also holds, so the finite-horizon expected arrival count is \(\Theta(\zeta T)\) under the bounded-rate model, then the service-normalized delay metrics satisfy
\[
\mathbb{E}\left[ \Latency^{(\zeta, \pi)} \right], \,\mathbb{E}\left[ \TTFT^{(\zeta, \pi)} \right] = O(1).
\]
\end{theorem}

The extension to time-varying rates preserves the dimensionality reduction structure from Theorem~\ref{thm:nested_wait_heavy_traffic}. The key adaptation is that threshold conditions~\eqref{eq:nested_wait_thresholds_time_varying} must hold uniformly over physical time \(t \in[0,T]\), replacing constant-rate arrival counts with the scaled-window arrivals \(\mathcal A_j^{(\zeta)}(t,h)\) over each service-normalized iteration window. These upper-side conditions control overload and memory growth. The separate first-segment waiting condition controls the time needed for the Segment 1 entry queue to reach its threshold. For downstream segments, prompts transitioning from segment \(k-1\) to segment \(k\) are controlled by the corresponding time-varying thinning probability. The Lindley recursion and Doob's maximal inequality apply with worst-case arrival and thinning parameters, bounding queue lengths with the same logarithmic overhead. Appendix~\ref{appendix:: proof of nested optimality, time varying.} provides the detailed proof.

\subsection{Segment Design}\label{sec:segment_design}

When LLM systems handle many distinct output lengths, maintaining individual thresholds for each type becomes impractical: the algorithm must track queue lengths separately for every output length, increasing memory overhead and implementation complexity. The \textbf{segment-wise design} addresses this by grouping similar output lengths into $L \leq m$ segments, where each segment $k$ aggregates types with comparable decode lengths. This reduces algorithmic complexity from $O(m)$ to $O(L)$ while preserving the threshold logic: prompts still classify themselves on the fly, but now at segment boundaries rather than individual type boundaries.

We assume constant arrival rates for clarity, though the approach extends to time-varying rates as discussed in Section~\ref{sec:time_varying}. Consider $m$ prompt types with decode lengths $l_1' < \cdots < l_m'$ and arrival rates $\lambda_1, \lambda_2, \cdots, \lambda_m$. Our objective is to merge these into $L$ ordered segments, where $L \leq m$. Choose cut points \(0=h_0<h_1<\cdots<h_L=m\), so segment \(k\) contains output-length classes \(h_{k-1}+1,\ldots,h_k\).

For each segment $k \in \{1, \dots, L\}$, the total arrival rate is defined as $\lambda_k' = \sum_{h_{k-1}< j\le h_k} \lambda_j$. Additionally, we define $\lambda'(k_1 \to k_2) := \sum_{r=k_1}^{k_2} \lambda_r'$ for $1 \leq k_1 \leq k_2 \leq L$.

Given a policy $\pi=[n_1,\dots,n_L]$ with threshold $n_k$ for the $k$-th segment in Algorithm \ref{alg:nested_wait}, we formulate a linear system analogous to Equation \eqref{eq:nested_wait_thresholds}:
\begin{equation}          \label{eq:nested_wait_thresholds_L}
    \begin{aligned}
    &\Delta T_{[1:L]}(n_{1:L}) < \frac{n_1}{\sum_{r=1}^L \lambda_r'},  \\
    &p_k < \frac{n_k}{n_{k-1}} < 1,\quad \forall k=2,\dots,L,
    \end{aligned}
\end{equation}
where $p_k =\frac{\lambda'(k \to L)}{\lambda'(k-1 \to L)}$ for \(k\ge2\). The strict first inequality gives negative drift for the first segment's entry queue, which is needed for the \(O((\zeta T)^{-1})\) throughput gap and \(O(1)\) service-normalized delay in Theorem~\ref{thm:nested_wait_heavy_traffic, any segment}. Let \(r_k=l_{h_k}'\) be the decode boundary of segment \(k\), with \(r_0=0\). Define \(\delta_1=r_1+1\), \(\bar s_1=r_1/2\), and for \(k\ge2\), \(\delta_k=r_k-r_{k-1}\) and \(\bar s_k=(r_{k-1}+1+r_k)/2\). Then \(\Delta T_{[1,\dots,L]}(n_1,\dots,n_L)=d_0+d_1M^\pi(n_1,\dots,n_L)\), where the base memory used by threshold-sized segment batches is
\[
M^{\pi}(n_1, \dots, n_L) = \sum_{k=1}^L n_k (l+\bar s_k)\delta_k.
\]
For heterogeneous prefill lengths, the segment logic is unchanged because boundaries are determined by decode progress. A pathwise sufficient memory condition uses a segment-wise upper bound \(\bar l_k=\max\{l_j: \text{output-length class }j\text{ can enter segment }k\}\). In the base segment-\(k\) term, replace \(l+\bar s_k\) by \(\bar l_k+\bar s_k\); in the segment-\(k\) boundary-buffer terms, replace \(l+r_{k-1}\) by \(\bar l_k+r_{k-1}\). A class-weighted memory term would require an additional concentration or token-cap argument and is not part of the theorem below.

The segment-wise design achieves two practical goals: (i) \textbf{coarsening} nearby output lengths into shared decode-stage controls; and (ii) \textbf{simplicity} in deployment---the scheduler maintains thresholds for a small number of decode-stage segments rather than for every possible output length. The theoretical guarantee shows that, under the chosen fixed segmentation model, this simplification preserves the same asymptotic throughput rate, with the same logarithmic safety-buffer structure.

\begin{theorem}
\label{thm:nested_wait_heavy_traffic, any segment}
Starting from the empty system, for any fixed number of segments $L$ and fixed partition \(0=h_0<h_1<\cdots<h_L=m\), let constants in the bounds depend on this segmentation. The displayed memory formula is for common prefill length \(l\); with heterogeneous prefill lengths, use the base and boundary-buffer replacements described above. Without loss of generality, retain downstream segments with \(p_k>0\). Fix a constant \(K<\infty\) independent of \(\zeta\) and \(T\), and choose \(\delta\in(0,1)\), possibly depending on \(\zeta\) and \(T\), such that \(\delta\le K(1+\zeta T)^{-1}\). Suppose thresholds $\pi = [n_1, \dots, n_L]$ satisfy Equation \eqref{eq:nested_wait_thresholds_L}. If the physical memory capacity \(C\) satisfies:
\begin{equation}
    \label{eq:nested_wait_memory_L}
    \begin{aligned}
	C &\ge M_{\mathrm{req},L}^{(\zeta,\pi)}
	:= M^\pi
		+ \sum_{k=2}^L (l + r_{k-1}) n_k \\
	&\quad
	+ \sum_{k=2}^L (l + r_{k-1}) \theta_k^{-1}
\ln\left( \frac{m(1+\zeta T/d_0)}{\delta} \right) \\
&= O\biggl( M^\pi
+ \sum_{k=2}^L (l + r_{k-1}) \theta_k^{-1}
\ln\left( \frac{m(1+\zeta T/d_0)}{\delta} \right) \biggr),
    \end{aligned}
\end{equation}
where, for each downstream segment \(k\ge2\), $\theta_k>0$ is the unique solution of \(e^{-\theta_k n_k}(1-p_k+p_ke^{\theta_k})^{n_{k-1}}=1\), with lower bound $8(n_k-n_{k-1}p_k)/n_{k-1}$. The logarithm uses \(m\) as a conservative upper bound on the number of segment boundaries, since \(L\le m\).
Then finite-memory Nested WAIT satisfies:
$$\throughput^* - \mathbb{E}\left[ \throughput^{(\zeta, \pi)} \right] = O\left( (\zeta T)^{-1} \right),$$
$$\mathbb{E}\left[ \Latency^{(\zeta, \pi)} \right] = O(1),$$
$$\mathbb{E}\left[ \TTFT^{(\zeta, \pi)} \right] = O(1).$$
Moreover, the probability that any memory-overflow-induced eviction occurs over the time horizon \([0,T]\) is at most \(\delta\).
\end{theorem}

\paragraph{Proof of Theorem~\ref{thm:nested_wait_heavy_traffic, any segment}.}
The proof applies the Nested WAIT argument in Appendix~\ref{appendix::proof of nested_wait_heavy_traffic} to aggregated segment classes. The only changes are the segment arrival rates and the segment boundaries; the boundary-thinning domination, Lindley bound for downstream boundary queues, and high-probability martingale buffer are the same as in Theorem~\ref{thm:nested_wait_heavy_traffic}. Segment \(k\) contains original output-length classes with aggregate rate \(\lambda_k'=\sum_{h_{k-1}<j\le h_k}\lambda_j\), and the rate of prompts that can reach segment \(k\) is \(\lambda'(k\to L)=\sum_{r=k}^L\lambda_r'\).

The aggregation is coupled to the original class process as follows. Prompts carry their exact output lengths, but Nested WAIT only observes completions when they occur and does not use unrevealed future completion information. Thus, conditional on the scheduler's filtration before boundary \(r_{k-1}\), the selected prompts that have reached that boundary have independent unrevealed continuation indicators with common continuation probability \(p_k=\lambda'(k\to L)/\lambda'(k-1\to L)\). If the selected set is smaller because the segment is partially idle, adding dummy Bernoulli trials only enlarges the comparison batch. Hence the number of prompts that continue to segment \(k\) is dominated by a binomial random variable with parameters \(n_{k-1}\) and \(p_k\), exactly as in the boundary-thinning step of Theorem~\ref{thm:nested_wait_heavy_traffic}. Inside a coarse segment, exact completions before the right boundary \(r_k\) only remove prompts and release memory. The comparison process that keeps every segment-\(k\) prompt resident through \(r_k\) therefore pathwise dominates the actual segment-level queues and memory usage. The base memory term is precisely this conservative overcount. The threshold conditions in \eqref{eq:nested_wait_thresholds_L} give negative drift for the Segment 1 entry queue and for every downstream boundary queue. Therefore the same Lindley-recursion bounds give uniformly bounded expected entry and boundary queues. The terminal-work and time-average queue conversions used in Appendix~\ref{appendix::proof of nested_wait_heavy_traffic} then yield the \(O((\zeta T)^{-1})\) normalized throughput gap and \(O(1)\) service-normalized latency and TTFT bounds.

It remains to check memory. The base term \(M^\pi=\sum_{k=1}^L n_k(l+\bar s_k)\delta_k\) charges the threshold-controlled interior stages of each segment. The resident boundary queue at \(r_{k-1}\), \(k\ge2\), is treated separately, with deterministic component \(n_k(l+r_{k-1})\) and stochastic carryover controlled by the exponential-martingale bound used in Appendix~\ref{appendix_lemma::proof of solution to martingale exists}. Applying that bound over at most \(1+\zeta T/d_0\) processing opportunities and union bounding over downstream segment boundaries gives the logarithmic buffer in \eqref{eq:nested_wait_memory_L}. Since \(L\le m\), using \(m\) in the logarithm is conservative. Let \(\mathcal E_L\) be this high-probability buffer event. Coupling the finite-memory implementation to the threshold-controlled comparison dynamics until the first memory repair, the same resident-boundary memory accounting as in Appendix~\ref{appendix::proof of nested_wait_heavy_traffic} shows that repair is not invoked on \(\mathcal E_L\) whenever \(C\ge M_{\mathrm{req},L}^{(\zeta,\pi)}\). Hence no evictions occur on \(\mathcal E_L\), and the probability of any memory-overflow-induced eviction is at most \(\delta\).

The unconditional performance bounds follow from the same bad-event bridge as in Theorem~\ref{thm:nested_wait_heavy_traffic}. Let \(A_T\) be the total number of external arrivals over \([0,T]\), so \(A_T\sim\mathrm{Poisson}(\mu)\) with \(\mu=\zeta T\sum_{r=1}^L\lambda_r'\). For any event \(F\) with \(\mathbb P(F)\le\delta\), the Poisson split gives \(\mathbb E[A_T{\bf 1}_F]\le 2\mu\delta+C_1\mu e^{-c_1\mu}\). Taking \(F=\mathcal E_L^c\), the finite-horizon censored delay contribution on the bad event is at most \(\zeta T(2\delta+C_1e^{-c_1\mu})=O(1)\), because \(\delta\le K(1+\zeta T)^{-1}\) and \(\sum_{r=1}^L\lambda_r'>0\) is fixed. The bad-event throughput contribution is at most \(l_m'\mathbb E[A_T{\bf 1}_{\mathcal E_L^c}]/(\zeta T)=O((\zeta T)^{-1})\). Combining these bounds with the comparison-dynamics estimates gives the stated finite-memory Nested WAIT performance.

We validate the segment-wise design numerically using a real-data workload derived from the \texttt{lmsys-chat-1m} dataset~\cite{zheng2023lmsys} (Section~\ref{sec:experiments}). Decode lengths range up to 500 tokens in this workload, and the segment arrival rates are proportional to the empirical frequencies of requests whose output lengths fall in the corresponding decode-stage ranges.

We analyze the three components of the memory bound in Equation \eqref{eq:nested_wait_memory_L}: (1) the base batch memory \(M^\pi=\sum_{k=1}^L n_k(l+\bar s_k)\delta_k\), (2) the deterministic boundary-queue term \(\sum_{k=2}^L n_k(l+r_{k-1})\), and (3) the high-probability boundary-queue term \(\sum_{k=2}^L \theta_k^{-1}(l+r_{k-1})\ln\!\left( \frac{m(1+\zeta T/d_0)}{\delta} \right)\).

\begin{figure}[ht]
    \centering
    \includegraphics[width=0.55\textwidth]{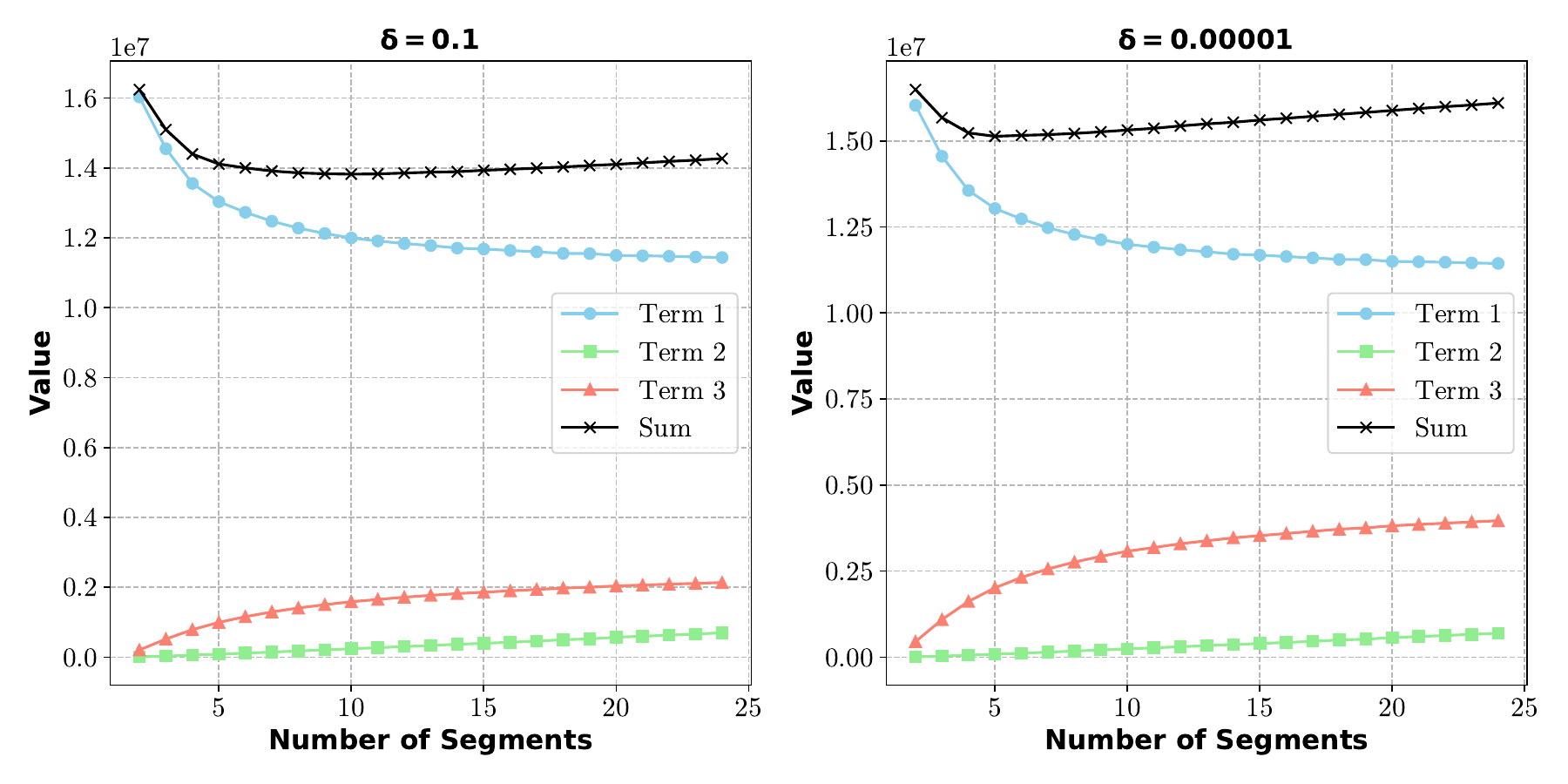}
    \caption{Memory-bound components under low arrival rate (total rate = 50, $T = 200$). Left: $\delta = 0.1$; Right: $\delta = 10^{-5}$.}
    \label{fig:fixed_T_200_rate_50_delta_0.1_1e-5}
\end{figure}

\begin{figure}[ht]
    \centering
    \includegraphics[width=0.55\textwidth]{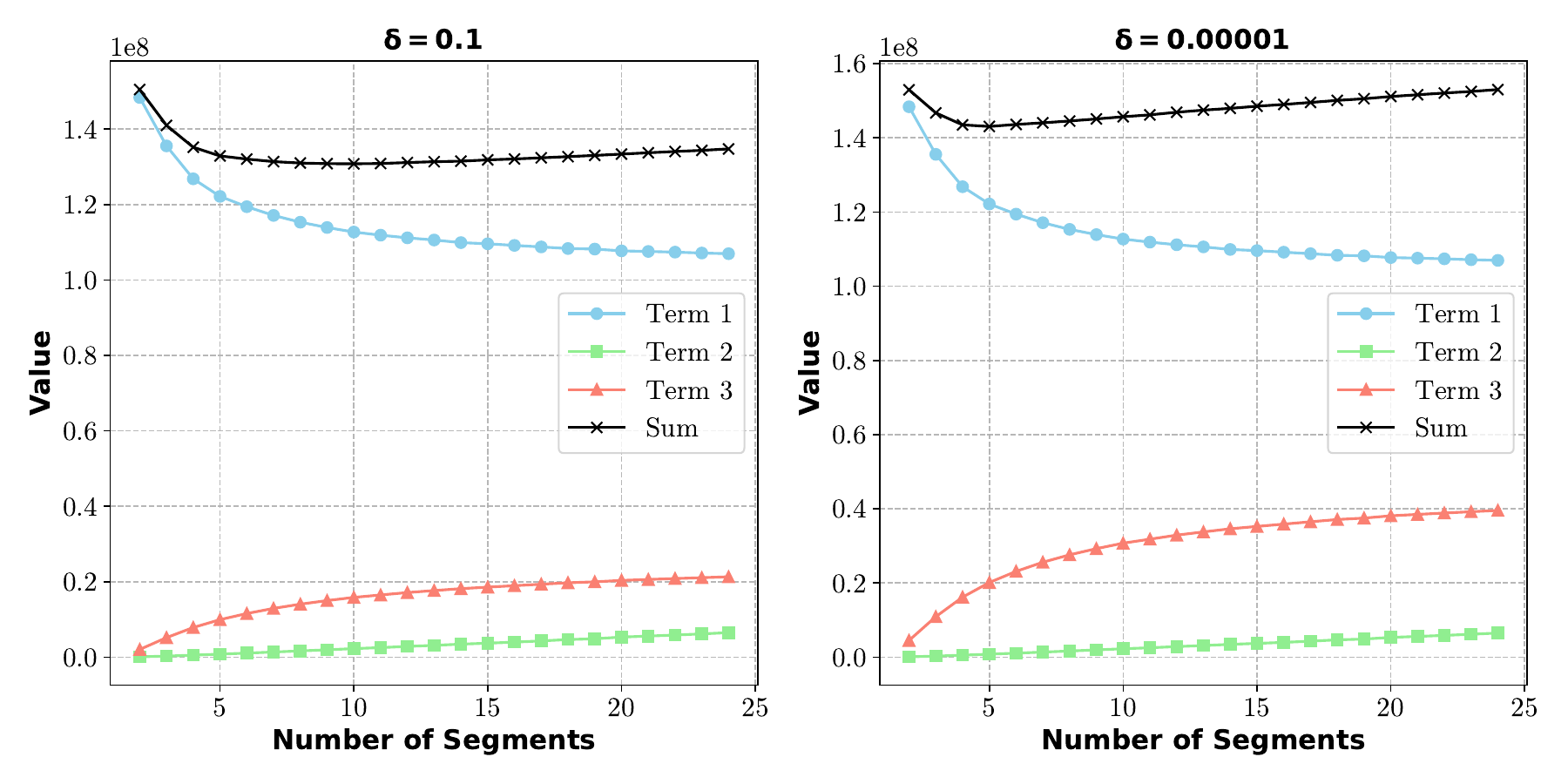}
    \caption{Memory-bound components under high arrival rate (total rate = 500, $T = 200$). Left: $\delta = 0.1$; Right: $\delta = 10^{-5}$.}
    \label{fig:fixed_T_200_rate_500_delta_0.1_1e-5}
\end{figure}

\begin{figure}[ht]
    \centering
    \includegraphics[width=0.55\textwidth]{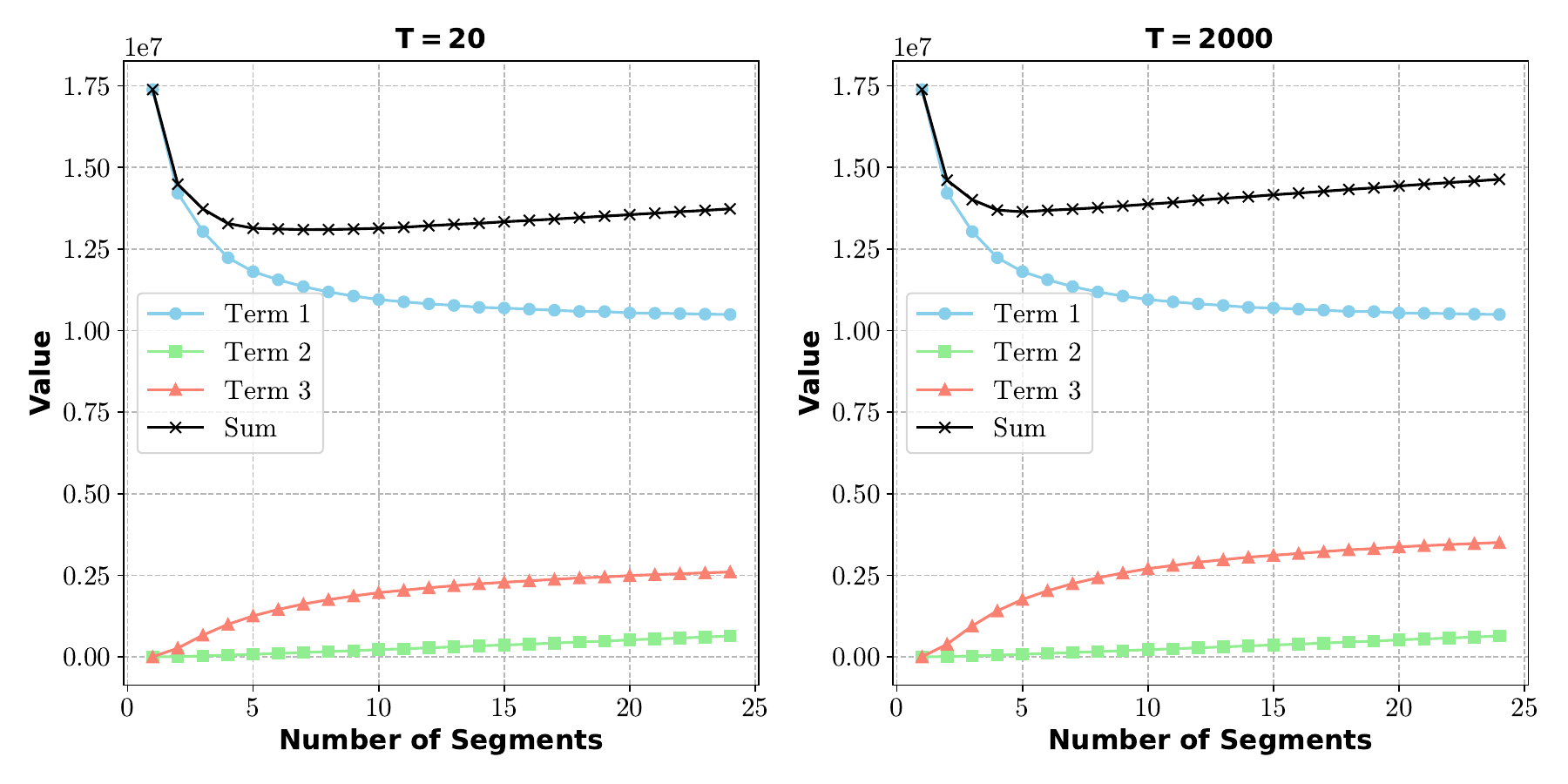}
    \caption{Memory-bound components with varying time horizons (total rate = 50, $\delta = 0.001$). Left: $T = 20$; Right: $T = 2000$.}
    \label{fig:fixed_delta_0.001_rate_50_T_20_2000}
\end{figure}

Figures \ref{fig:fixed_T_200_rate_50_delta_0.1_1e-5}, \ref{fig:fixed_T_200_rate_500_delta_0.1_1e-5}, and \ref{fig:fixed_delta_0.001_rate_50_T_20_2000} plot these three memory components (y-axis) as functions of the number of segments $L$ (x-axis), under varying arrival rates, time horizons $T$, and confidence levels $\delta$. Each panel shows: Term 1 (blue circles) representing the base threshold-batch memory \(M^\pi\), Term 2 (green squares) representing the deterministic queue bound, Term 3 (red triangles) representing the stochastic safety buffer with logarithmic dependence on $T$ and $\delta$, and their sum (black crosses) representing the total required memory \(M_{\mathrm{req},L}^{(\zeta,\pi)}\). Three key observations emerge: (i) Term 1 dominates Terms 2 and 3 across all parameter settings, confirming that the safety buffer remains modest relative to the base batching requirement. (ii) The total required memory (black line) follows a \textbf{U-shaped curve} with respect to $L$: too few segments waste memory by over-provisioning for heterogeneous types (high Term 1 at small $L$), while too many segments increase overhead from maintaining separate queues. (iii) Term 3 exhibits smooth growth as $\delta$ decreases (comparing left vs. right panels in Figures \ref{fig:fixed_T_200_rate_50_delta_0.1_1e-5} and \ref{fig:fixed_T_200_rate_500_delta_0.1_1e-5}) and as $T$ increases (comparing Figure~\ref{fig:fixed_delta_0.001_rate_50_T_20_2000}), consistent with the logarithmic bound in the theoretical analysis.

These extensions demonstrate that the core threshold-based design, eviction prevention through controlled admission, remains robust within the stated modeling conditions. Under the bounded-window, first-segment-waiting, and fixed-segmentation conditions above, the coupling-based analysis extends with logarithmic overhead. Section~\ref{sec:experiments} illustrates this implementation on a real workload, where Nested WAIT uses coarser decode-stage segments rather than maintaining a separate threshold for every possible output length.

%% file: appendix_notation.tex
\section{Notation Summary}
\label{app:notation}

\begin{table}[htbp]
\centering
\caption{Summary of notation}
\label{tab:notation}
\small
\begin{tabular}{cl}
\toprule
\textbf{Symbol} & \textbf{Description} \\
\midrule
$m$ & Number of prompt types \\
$j$ & Index of prompt type, $j \in \{1, \dots, m\}$ \\
$\lambda_j$ & Arrival rate of type-$j$ prompts \\
$l_j$ & Input length of type-$j$ prompts after prefill \\
$l_j'$ & Output length of type-$j$ prompts in the decode phase \\
$s$ & Stage index: $s=0$ for prefill and $s \in \{1, \dots, l_j'\}$ for decode \\
$k$ & Segment index in Nested WAIT \\
$b$ & Batch index \\
$n_{js}^t$ & Number of type-$j$ prompts at stage $s$ at time $t$ \\
$n_j$ & Per-stage threshold for type-$j$ prompts in WAIT \\
$n_k$ & Threshold for segment $k$ in Nested WAIT \\
\midrule
$C$ & GPU memory capacity \\
$\mathcal{G}^t$ & Set of prompts with KV cache on GPU at time $t$ \\
$B^t$ & Batch of prompts processed at iteration $t$, with $B^t \subseteq \mathcal{G}^t$ \\
$\tau$ & Iteration time to process a batch \\
$d_0$ & Fixed overhead time per batch iteration \\
$d_1$ & Time cost per unit of KV cache memory \\
\midrule
$M^*$ & Memory requirement needed to support the fluid equilibrium \\
$n_j^*$ & Equilibrium per-stage inventory of type-$j$ prompts \\
$\throughput^{(T,\pi)}$ & Effective throughput of policy $\pi$ over horizon $[0,T]$ \\
$\Latency^{(T,\pi)}$ & Average latency of policy $\pi$ over horizon $[0,T]$ \\
$\TTFT^{(T,\pi)}$ & Average time to first token of policy $\pi$ over horizon $[0,T]$ \\
$\throughput^*$ & Equilibrium effective throughput, measured in completed decode tokens per unit time \\
$\Pi$ & Class of admissible scheduling policies \\
\bottomrule
\end{tabular}
\end{table}

%% file: Appendix.tex
\section{Proofs of Propositions}

\subsection{Proof of Proposition~\ref{prop:large_rate}}
\label{appendix:proof_large_rate}

\begin{proof}
We use a service-capacity cut. The memory-dependent work required by arriving prompts already saturates the server at the stated boundary; the fixed positive overhead \(d_0\), together with finite memory capacity, makes the service capacity strictly smaller at equality.

Consider a system with only type-\(j\) prompts and finite memory capacity \(C\). Every completed prompt must be selected once for prefill and once for each decode stage \(s=1,\ldots,l_j'\). Across these \(l_j'+1\) service steps, its KV-cache contribution is
\[
    S_j=\sum_{s=0}^{l_j'}(l_j+s)=(l_j'+1)\left(l_j+\frac{l_j'}{2}\right).
\]
Because an iteration with selected memory \(M\) has duration \(d_0+d_1M\ge d_1M\), completing \(C_j(T)\) type-\(j\) prompts by time \(T\) requires at least \(d_1S_j C_j(T)\) units of calendar time, up to the bounded contribution of prompts initially present in the system. Hence
\[
    C_j(T)\le \frac{T}{d_1S_j}+O(1).
\]
This proves a linear backlog gap when \(\lambda_jd_1S_j>1\).

It remains to handle the boundary case \(\lambda_jd_1S_j=1\). Because the GPU memory capacity is finite, there is a finite upper bound \(B_C\) on the number of type-\(j\) prompts that can be served in one iteration; for instance \(B_C\le C/l_j\). Therefore completing \(C_j(T)\) prompts requires at least \(C_j(T)/B_C\) nonempty iterations. Each such iteration incurs the fixed overhead \(d_0\). Combining memory-dependent service and fixed overhead gives
\[
    T \ge d_1S_j C_j(T)+d_0\,\frac{C_j(T)}{B_C}-O(1),
\]
and hence
\[
    C_j(T)\le \frac{T+O(1)}{d_1S_j+d_0/B_C}.
\]
At the boundary, \(\lambda_j=1/(d_1S_j)\), so this upper bound is strictly smaller than the expected arrivals \(\lambda_jT\) by a linear amount. Thus in both the strict and boundary cases, the same capacity cut applied on every prefix \([0,t]\) implies that the expected number of type-\(j\) prompts that have arrived but not completed by time \(t\) is \(\Omega(t)\). Under the finite-horizon delay convention in Section~\ref{sec:model}, unfinished prompts contribute their elapsed time up to the horizon, including time spent in the external prefill queue. Therefore the expected total censored waiting over \([0,T]\), equivalently the area under the unfinished-prompt process, is \(\Omega(T^2)\). Since the expected number of type-\(j\) arrivals over \([0,T]\) is \(\Theta(T)\), the expected average latency satisfies \(\mathbb{E}[\textbf{Latency}^{(T,\pi)}]=\Omega(T)\).
\end{proof}

\subsection{Proof of Proposition~\ref{prop::online policy expected throughput upper bound, multiple type}}
\label{appendix:proof_throughput_upper_bound}

\begin{proof}
The proof compares completed decode-token service with the cumulative decode-token work brought by exogenous arrivals. The fluid calculation identifies the memory requirement and batch composition under which this upper bound is attained in equilibrium.

In the fluid model, the equilibrium parameters are:
$$\Delta T = \frac{d_0}{1 - d_1\sum_{j=1}^m \lambda_j(l_j'+1)(l_j + l_j'/2)}, \quad n_j^* = \Delta T \cdot \lambda_j,$$
$$M^* = \Delta T \cdot \sum_{j=1}^m \lambda_j(l_j'+1)(l_j + l_j'/2), \quad \throughput^* = \sum_{j=1}^m \lambda_j l_j'.$$

Over time horizon $[0,T]$, let $\text{Arrivals}_j^T$ denote the total arrivals of type-$j$ prompts. Under the throughput convention in Section~\ref{sec:model}, the completed output tokens counted over the horizon are bounded by the output-token work brought by these arrivals:
$$\text{Throughput}^{(\pi,T)} \leq \sum_{j=1}^m \text{Arrivals}_j^T \cdot l_j'.$$

Taking expectations and dividing by $T$:
$$\mathbb{E}[\throughput^{(T,\pi)}] \leq \sum_{j=1}^m \mathbb{E}\left[\frac{\text{Arrivals}_j^T}{T}\right] \cdot l_j' = \sum_{j=1}^m \lambda_j l_j' = \throughput^*.$$
\end{proof}

\subsection{Proof of Proposition~\ref{prop::simple random walk, tight bound}}
\label{appendix:proof_boundary_delay_lower_bound}

\begin{proof}
We use a single-type instance at the fluid boundary. It is without loss for this lower bound to take the initial state empty; more generally, any bounded initial resident work only changes the constant \(K\) below. The point of the construction is to remove all type-mix complications: the only remaining source of delay is the stochastic fluctuation of cumulative arrivals around the boundary service capacity.

Consider one prompt type with prefill length \(l\), one-token output \(l'=1\), and arrival rate \(\lambda\). Let
\[
    S=2\left(l+\frac{1}{2}\right)
\]
be the total KV-cache mass accumulated by one prompt over its prefill and decode stages. Choose \(d_0,d_1,\lambda\) so that the fluid equilibrium has iteration length
\[
    \Delta^*=\frac{d_0}{1-d_1\lambda S}
\]
and \(n=\lambda\Delta^*\) is an integer. The fluid memory requirement is then \(M^*=\lambda\Delta^*S=nS\). Set the physical capacity to \(C=M^*\). At this capacity, the fluid equilibrium has room for exactly the threshold-sized batch composition with \(n\) prompts at each stage and no additional buffer.

Now consider the \(\zeta\)-scaled system over a calendar horizon \(T\). Let \(A(t)\) be the cumulative number of external arrivals by time \(t\), so \(A(t)\sim\mathrm{Poisson}(\zeta\lambda t)\). Let \(F(t)\) be the number of prompts that have received their first output token by time \(t\). Since \(l'=1\), receiving the first token is the same as completion.

We use a deterministic prefix capacity cut. For a prefix \([0,t]\), let \(P(t)\) be the number of prefill selections, let \(N(t)\) be the number of nonempty iterations, and let \(R_0\) be the number of initially resident decode-ready prompts. Stage balance gives \(F(t)\le P(t)+R_0\). Thus the selected KV work in this prefix is at least
\[
    lP(t)+(l+1)F(t)
    \ge
    (2l+1)F(t)-lR_0
    =
    SF(t)-O(1).
\]
Since every iteration respects the memory constraint and \(C=nS\), its selected KV work is at most \(C\). Therefore \(N(t)\ge (SF(t)-O(1))/C=(F(t)-O(1))/n\). In the \(\zeta\)-scaled system, each nonempty iteration has fixed-overhead term \(d_0/\zeta\) and selected-work term \(d_1/\zeta\). Hence, for a constant \(K\) depending only on the constructed instance and boundary effects,
\[
    t
    \ge
    \frac{d_0}{\zeta}\frac{F(t)-K}{n}
    +
    \frac{d_1}{\zeta}S(F(t)-K)
    =
    \frac{\Delta^*}{\zeta n}(F(t)-K),
\]
where \(\Delta^*=d_0+d_1nS\). Since \(n=\lambda\Delta^*\), this gives
\[
    F(t)\le \zeta\lambda t+K,\qquad 0\le t\le T.
\]

Let \(U(t)\) be the number of arrivals by time \(t\) that have not yet received a first token. Then
\[
    U(t)\ge \bigl(A(t)-\zeta\lambda t-K\bigr)^+ .
\]
For Poisson random variables, the central limit theorem and uniform integrability of positive parts imply that, for all sufficiently large \(\zeta\lambda t\),
\[
    \mathbb E\!\left[\bigl(A(t)-\zeta\lambda t-K\bigr)^+\right]
    \ge
    c_0\sqrt{\zeta t}
\]
for some constant \(c_0>0\) depending only on the constructed instance. Therefore
\[
    \mathbb E\!\left[\int_0^T U(t)\,dt\right]
    =
    \Omega\!\left(\int_0^T \sqrt{\zeta t}\,dt\right)
    =
    \Omega\!\left(\frac{(\zeta T)^{3/2}}{\zeta}\right).
\]
The integral \(\int_0^T U(t)\,dt\) is the total calendar-time censored TTFT over all arrivals before \(T\). Under the finite-horizon delay convention in Section~\ref{sec:model}, expected total censored delay is normalized by the expected number of arrivals, \(\mathbb E[A(T)]=\zeta\lambda T=\Theta(\zeta T)\). Thus the calendar-time average TTFT is \(\Omega(\zeta^{-1}(\zeta T)^{1/2})\). The proposition reports service-normalized delay by multiplying calendar time by \(\zeta\). Since \(l'=1\), the same lower bound applies to latency. Therefore
\[
    \mathbb{E}[\TTFT^{(\zeta,\pi)}],
    \mathbb{E}[\Latency^{(\zeta,\pi)}]
    \ge
    c\,(\zeta T)^{1/2}
    =
    \Omega((\zeta T)^{1/2})
\]
for some constant \(c>0\) depending only on the constructed instance. This proves the claimed lower bound.
\end{proof}

\subsection{Proof of Proposition~\ref{prop:lower_bound_FCFS}}
\label{appendix:proof_fcfs_lower_bound}

\begin{proof}
We construct a single-type instance in which the fluid equilibrium fits exactly in memory, but FCFS repeatedly selects the wrong finite batch composition. Let \(l=1\), \(l'=1\), and \(\lambda=4\). Choose \(d_1<1/24\) and set \(d_0=1-12d_1\). For this instance, the fluid equation is
\[
    n^*=\lambda\{d_0+d_1(3n^*)\}.
\]
Thus \(n^*=4\), the equilibrium memory is \(M^*=3n^*=12\), and we set \(C=M^*=12\). The fluid batch composition is four prefill prompts and four decode prompts. It has selected memory \(4+2\cdot4=12\), iteration time \(d_0+12d_1=1\), and completes \(\throughput^*=\lambda l'=4\) useful decode tokens per unit time.

First consider a decision epoch with no resident prompts and \(L\ge 12\) prompts waiting externally for prefill. The integer \(L\) is a fixed constant chosen below and does not depend on the time horizon \(T\). This state is feasible because external prefill-waiting prompts carry no resident KV cache. The selected batch composition space is finite: a batch can be represented by \((b_0,b_1)\), the numbers of prefill and decode prompts selected, with feasibility condition
\[
    b_0+2b_1\le 12 .
\]
The unique full-throughput balanced composition is \((4,4)\). However, whenever FCFS reaches a decision epoch with no resident prompts and at least 12 external prompts waiting, it selects the batch \((12,0)\). This batch takes one unit of time and completes no prompt. After the prefill update, the 12 admitted prompts would occupy \(24\) units of decode-stage KV cache, so the LIFO repair rule evicts six of them and leaves six decode-stage prompts resident. At the next decision epoch FCFS must batch those six resident decode prompts. Their selected memory is \(12\), so no external prefill prompt can be admitted in the same iteration. This second batch also takes one unit of time and completes six prompts. The resident state is then empty again.

Thus, as long as the external queue has at least 12 prompts at the beginning of each empty-resident epoch, FCFS is trapped in the two-batch cycle
\[
    (12,0)\quad\longrightarrow\quad(0,6),
\]
which completes only six useful decode tokens in two units of time. Its throughput on this cycle is therefore \(3\), strictly below the fluid benchmark \(4\).

Starting from the empty system, this bad state is reached with a probability that is positive and independent of \(T\). Choose \(\varepsilon\in(0,d_0+d_1)\), and let \(H_L\) be the event that at least \(L+12\) arrivals occur in \([0,\varepsilon]\). Since \(L\) is fixed, \(\Pr(H_L)>0\). On \(H_L\), the first arrival starts a \((1,0)\) batch, and the other arrivals counted in \(H_L\) occur before that batch completes. Additional arrivals before the end of the warm-up can only increase the external queue. At the next decision epoch, FCFS has one decode prompt resident and at least \(L+11\) external prompts, so it selects \((10,1)\). After this batch, the old decode prompt completes and the ten admitted prompts would occupy \(20\) units of decode-stage memory, so the repair rule evicts four and leaves six decode prompts resident. The next batch is therefore \((0,6)\), because these six decode prompts already use all \(12\) units of selected memory. Let \(\sigma\) be the decision epoch after this third batch. Then \(\sigma\le\varepsilon+3\), the resident state is empty, and the external queue has at least \(L\) prompts. The seven completions during this warm-up are a bounded end effect.

It remains to show that, once reached, this cycle has positive probability of persisting over the whole horizon. Let \(Q_r\) be the external queue length at the beginning of the \(r\)-th empty-resident epoch after time \(\sigma\), under the saturated evolution described above. If \(Q_r\ge12\), then over the next two one-unit iterations the system consumes 12 external prompts, returns six evicted prompts to the external queue, and receives \(A_r\sim\mathrm{Poisson}(8)\) new external arrivals. By the strong Markov property and independent increments of the Poisson arrival process, the variables \(A_r\) are iid after \(\sigma\). Hence
\[
    Q_{r+1}=Q_r-6+A_r .
\]
The increments \(A_r-6\) have positive mean \(2\) and finite exponential moments. A random walk with such increments has a strictly positive probability of never crossing any fixed lower level when it is started sufficiently far above that level; moreover this survival probability tends to one as the initial offset grows. Therefore, for a sufficiently large fixed \(L\),
\[
    p_L:=\Pr\left\{L+\sum_{u=0}^{r-1}(A_u-6)\ge12
    \text{ for all }r\ge0\right\}>0
\]
by the standard positive-survival property of a random walk with positive drift. Fix such an \(L\) for the rest of the proof.

Let \(E_L\) be the event that \(H_L\) occurs and that this post-\(\sigma\) random walk never falls below 12. Since on \(H_L\) we have \(Q_0\ge L\), monotonicity of the recursion gives a conditional survival probability at least \(p_L\), and hence
\[
    \Pr(E_L)\ge \Pr(H_L)p_L>0 .
\]
On \(E_L\), after the bounded warm-up time \(\sigma\), FCFS remains in the two-batch cycle forever and completes six one-token prompts every two units of time. Thus \(C(T)/T\to3\) on \(E_L\); the bounded warm-up completions and any incomplete terminal cycle contribute only \(O(1/T)\). Also, by the strong law of large numbers for the Poisson arrivals, \(A(T)/T\to4\) almost surely. Hence, on \(E_L\) up to a null set,
\[
    \lim_{T\to\infty}\frac{A(T)-C(T)}{T}=4-3=1 .
\]
Let \(X_T=(A(T)-C(T))/T\). Since the system starts empty and each completed prompt must have arrived exogenously, \(C(T)\le A(T)\) pathwise, so \(X_T\ge0\). Moreover, all outputs have length one, \(\throughput^*=\mathbb E[A(T)]/T=4\), and \(\mathbb E[\throughput^{(T,\mathrm{FCFS})}]=\mathbb E[C(T)]/T\). Therefore Fatou's lemma gives
\[
\begin{aligned}
    \liminf_{T\to\infty}
    \left\{\throughput^*-\mathbb E[\throughput^{(T,\mathrm{FCFS})}]\right\}
    &=
    \liminf_{T\to\infty}\mathbb E[X_T] \\
    &\ge
    \liminf_{T\to\infty}\mathbb E[X_T\mathbf 1_{E_L}] \\
    &\ge
    \mathbb E\left[\liminf_{T\to\infty}X_T\mathbf 1_{E_L}\right]
    =
    \Pr(E_L)
    >
    0 .
\end{aligned}
\]
This gives
\[
    \throughput^*-\mathbb E[\throughput^{(T,\mathrm{FCFS})}]
    =
    \Omega(1),
\]
as claimed.
\end{proof}

\subsection{Proof of Proposition~\ref{prop:unknown_lower_bound}}
\label{appendix:proof_unknown_lower_bound}

\begin{proof}
We construct a two-output-length instance in which the known-type fluid equilibrium fits exactly in memory, but every non-predictive rule loses a constant amount of useful service in each saturated boundary block. Repeating these blocks gives the long-run gap, as in the proof of Proposition~\ref{prop:lower_bound_FCFS}.

Consider two prompt types with the same observable input features, the same prefill length, and the same first decode token. Type~1 completes after the first decode token, whereas type~2 continues to one additional decode stage. There is no side information correlated with the hidden type. Let a newly admitted prompt be type~2 with probability \(p\in(0,1)\), independently across prompts, and suppose the type is not revealed until the prompt either completes at the first boundary or survives into the second segment. Choose the arrival rates and service parameters so that the known-type fluid equilibrium is integral over one comparison block of length \(\Delta\): the block receives mean \(B\) new prompts, advances \(B\) unrevealed prompts across the first boundary, and allocates downstream capacity for \(s=pB\) long survivors, where \(B\) and \(s\) are integers. We choose the fixed parameters generically so that, among feasible finite selected-composition sequences in this block, this known-type composition is the unique sequence attaining the fluid useful-service count. Set the physical memory to the corresponding known-type requirement \(C=M^*\). At this boundary value the fluid block also contains a deterministic resident composition from the other prefill/decode stages, and this composition together with the \(B\) boundary crossings and \(s\) downstream slots exactly exhausts the memory budget.

We first prove a one-block lower bound. Consider a comparison block that begins with the deterministic resident composition and at least \(B\) unrevealed prompts available at the first-segment entry. Within a block of length \(\Delta\), any feasible policy can execute only finitely many nonempty iterations, because each nonempty iteration takes at least \(d_0>0\), and each selected batch has finitely many integer compositions because \(C<\infty\) and every resident prompt has positive memory. Thus the possible selected-composition sequences in the block form a finite set. Any sequence that differs from the known-type fluid composition by omitting, delaying, or evicting part of the deterministic resident work, or by advancing fewer than \(B\) unrevealed prompts across the first boundary, completes at least one fewer useful decode token than the fluid block. Since the action set is finite, all such off-fluid sequences have a uniform positive useful-service deficit.

It remains only to analyze the unique no-immediate-loss candidate: the policy advances exactly \(B\) unrevealed prompts across the first boundary while reserving downstream room for exactly \(s\) long survivors. Let \(\mathcal F\) be the sigma-field generated by the policy's observations just before this boundary revelation, including past arrivals, past boundary revelations, the current queues, and any internal randomization already realized by the policy. Conditional on \(\mathcal F\), non-predictiveness implies that the number of selected prompts that survive the boundary is
\[
    Y_B\sim\mathrm{Binomial}(B,p),
\]
independently of \(\mathcal F\). Since \(s=pB\) and \(0<p<1\), \(Y_B\) has nonzero variance and
\[
    \rho:=\Pr\{Y_B>s\}>0 .
\]
On this event the realized survivor count exceeds the downstream capacity present at \(C=M^*\). Therefore at least one survivor must be evicted and restarted, or retained as unfinished downstream work that displaces useful service from the current fluid block. Thus the saturated block has a positive lower bound on the expected sum of current useful-service deficit and displaced-work carryover.

It remains to make this block argument repeatable under arbitrary policy-induced states. We use a finite augmented block MDP. The augmentation only helps the controller: at the beginning of each comparison block it may see all arrivals that occur during that block and may use them immediately, but it still does not observe an unrevealed prompt's output length before the first boundary. A lower bound for this augmented controller is therefore a lower bound for the original online policies.

Let \(N_\Delta=\lfloor \Delta/d_0\rfloor+1\). In one block there are at most \(N_\Delta\) nonempty iteration starts, up to one residual iteration may cross a block boundary, and each feasible resident list has bounded length because every resident prompt has positive KV-cache size and total memory is at most \(C\). Choose finite caps \(H\) and \(H_A\) larger than the maximum number of prompt-stage advances, evictions, external arrivals, or deferred first-segment-equivalent units that can affect one block. Work or arrivals beyond these caps can only help fill the saturated first-segment pool or increase future displaced work, so truncating them at the cap cannot make the lower bound harder. At the beginning of block \(r\), define the capped state
\[
    S_r=(R_r,Q_r,E_r,U_r),
\]
where \(R_r\) is the resident list, including each prompt's revealed segment/stage and any eviction priority relevant to the policy, \(Q_r\) is the capped number of unrevealed prompts available at the first-segment entry, \(E_r\) is the capped amount of restarted or deferred work that can displace work in later blocks, and \(U_r\) records the finite residual in-service composition and remaining service category from an iteration crossing the block boundary. This state space is finite. The saturated fluid state \(s^\star\) has the deterministic fluid resident composition, \(Q=B\), \(E=0\), and no residual cross-boundary service.

For each capped state \(x\), a pure block action is a nonanticipative decision tree that, for at most \(N_\Delta\) within-block decision epochs, maps the finite capped within-block observation history to feasible admission, batching, preemption, and eviction decisions. The action may use all capped arrivals in the block from time zero, by the augmentation, but it cannot use unrevealed output-length labels before boundary revelation. Hence each \(\bar{\mathcal A}(x)\) is finite. Dependence on prompt identities, arrival times, older history, or internal randomization does not enlarge this finite model: conditional on the full observable history, any non-predictive selection of unrevealed prompts has conditionally i.i.d. Bernoulli-\(p\) continuation indicators, and only the capped selected composition and revealed boundary outcomes affect the transition kernel and useful-service count. Let \(G_r\) be the useful-service deficit in block \(r\) relative to the known-type fluid block, with tokens from evicted prompts counted as useless as in Section~\ref{sec:model}.

We next rule out zero-average recurrent classes in this finite augmented model. By the generic choice of parameters, from \(s^\star\) the only way to have zero one-block deficit before the boundary revelation is to execute the unique fluid selected-composition sequence: serve all deterministic resident work, advance exactly \(B\) unrevealed prompts across the first boundary, and reserve exactly \(s=pB\) downstream survivor slots. From any other capped state, or under any other pure block action, either the block immediately misses at least one unit of useful service, or it spends service repairing a resident-composition, backlog, residual-service, or displaced-work mismatch. Except for the displaced-work buffer discussed next, any zero-current-deficit repair transition must strictly decrease the finite rank
\[
    R(x)
    :=
    \text{resident-composition shortfall}
    +(B-Q)^+
    +\text{residual-service mismatch}.
\]
Thus any closed zero-average class must reduce to the saturated resident/backlog/residual configuration, with only the displaced-work buffer \(E\in\{0,\ldots,H\}\) possibly varying, and must use the fluid selected-composition sequence whenever it seeks zero current deficit.

On this reduced class, conditional on the full block-start history and on any realized internal randomization before the boundary decision, non-predictiveness gives \(Y_B\sim\mathrm{Binomial}(B,p)\) for the \(B\) boundary-crossing prompts. Let \(Z=Y_B-s\). Then \(\mathbb E[Z]=0\), and because \(0<p<1\) and \(s=pB\in\{1,\ldots,B-1\}\), \(\Pr\{Z>0\}>0\) and \(\Pr\{Z<0\}>0\). If the block has displaced-work level \(E\), the best zero-deficit use of old displaced work can only update the buffer as a finite reflected queue,
\[
    E^+ = \min\{H,\max\{0,E+Z\}\},
\]
with boundary loss at least
\[
    L(E,Z)=(E+Z-H)^+ + (-E-Z)^+ .
\]
The upper-boundary term is excess long-survivor work that cannot fit in memory at \(C=M^*\) even after using the finite buffer; it must be evicted, restarted, or pushed beyond the capped recoverable workload, losing useful service. The lower-boundary term is an unused downstream survivor slot when current survivors plus displaced work are insufficient, leaving part of the fluid block's useful service unfilled. Since the finite reflected chain driven by \(Z\) has both upward and downward moves with positive probability, every closed recurrent class has positive stationary probability of a boundary loss. Hence the average of \(L(E,Z)\), and therefore the average one-block useful-service deficit, is strictly positive on this reduced class. Combining this with the rank argument for all other states shows that every recurrent class of every stationary deterministic selector has strictly positive average one-block deficit. Since the finite augmented MDP has finitely many stationary deterministic selectors and recurrent classes, the minimum average deficit is some \(\epsilon>0\).

By the finite-state average-cost optimality equations, equivalently the dual of the average-cost linear program, there exists a bounded potential \(\Phi\) on capped states such that, for every capped state \(x\) and every pure block action \(a\in\bar{\mathcal A}(x)\),
\[
    \mathbb E\!\left[
        G_r+\Phi(S_{r+1})-\Phi(S_r)
        \mid S_r=x,\ a
    \right]\ge \epsilon .
\]
Now fix any randomized history-dependent non-predictive policy. Conditional on the full block-start history \(\mathcal F_r\), the policy induces a probability distribution over the finite set of pure block actions available at \(S_r\). Averaging the preceding inequality over this distribution gives
\[
    \mathbb E\!\left[
        G_r+\Phi(S_{r+1})-\Phi(S_r)
        \mid \mathcal F_r
    \right]\ge \epsilon
\]
in every block.

Summing the displayed inequality over the \(R_T=T/\Delta+O(1)\) complete comparison blocks and using boundedness of \(\Phi\) gives
\[
    \mathbb E\!\left[\sum_{r=0}^{R_T-1} G_r\right]
    \ge
    \epsilon R_T-O(1).
\]
Hence the accumulated effective-throughput deficit is \(\Omega(T)\). On every sample path, the completed-token rate is bounded by a finite constant \(R_{\max}\) because each nonempty iteration completes at most \(C\) one-token decode steps and takes at least \(d_0\) time. As in Proposition~\ref{prop:lower_bound_FCFS}, bounded end effects therefore give
\[
    \throughput^*-\mathbb E[\throughput^{(T,\pi)}]
    \ge \epsilon_0-O(1/T)
    =
    \Omega(1)
\]
for some \(\epsilon_0>0\) depending only on the constructed instance.
\end{proof}

\input{pf_thm_wait}

\input{pf_thm_nested}

\input{pf_thm_timevarying}

\input{pf_lemmas}

\input{appendix_b_additional}

%% file: pf_thm_wait.tex
\section{Proof of Theorem~\ref{thm:wait_heavy_traffic}}\label{appexidx::proof of wait_heavy_traffic}

This proof establishes the WAIT performance bounds when output lengths are known. The argument proceeds in two parts:
\begin{enumerate}
    \item \textbf{Single-type case}: We construct a coupled dominating process using the Lindley recursion and apply Kingman's bound for queue length expectations.
    \item \textbf{Multiple-type case}: We construct an auxiliary embedded full-threshold process and couple it to event-driven WAIT type by type. This embedded process separates stochastic entry accumulation from the event-driven batch composition while preserving the service opportunities created by the WAIT thresholds.
\end{enumerate}
Key techniques include: Lindley recursion for queue length bounds \citep[Proposition 6.3]{asmussen2003applied}, coupling construction to dominate the original process, Kingman's bound for negative drift processes \citep{kingman1962queues}, and standard random walk results \citep{strait1974maximum}.

Throughout this appendix, we use \(b\) to denote the index of the embedded full-threshold review epochs, which is distinct from the stage index \(s \in \{0,1,\ldots,l_j'\}\) used in the main text for decode stages. We use \(t\) for continuous time and \(B\) for the number of embedded review intervals.

We first consider the single-type case, then extend to multiple types.

\subsubsection{Single-Type Case}

Consider a single prompt type with threshold $n$. For general scheduling algorithms, analyzing this system in continuous time is challenging: KV cache grows dynamically during decode, arrivals are stochastic, and eviction risks complicate the dynamics.

The WAIT policy's \emph{threshold mechanism achieves a dimensionality reduction}. By waiting until exactly $n$ prompts accumulate before initiating a batch, the algorithm enforces a fixed batch size, which ensures constant processing time $\Delta T$ per batch. This decouples the continuous-time memory-constrained dynamics into a tractable discrete-time process: a memory-constrained queueing system becomes a simple random walk. The reduction is enabled by the algorithm design and is what makes the subsequent drift analysis apply.

Specifically, let \(\tilde{\lambda}\) denote the continuous-time arrival rate and let \(\mu=\tilde{\lambda}\Delta T\) be the expected number of arrivals during a threshold-sized batch. The boundary case is \(\mu=n\). For this preliminary calculation, view time through deterministic full-threshold review intervals of length \(\Delta T\), rather than through the exact event-driven batch-completion clock. The embedded review process, indexed by \(b\in\mathbb{N}\), is a zero-drift threshold queue. A review slot is ``stuck'' if fewer than \(n\) prompts are available at that review. This is the single-type version of the embedded comparison process used in the formal multi-type proof below.

\begin{lemma}[Queue Length and Stuck Time]\label{lemma::throughput gap, single}
This lemma connects queue accumulation to stuck review slots in the critical case \(\mu=n\). Let \(X^b\) denote the number of arrivals during review interval \(b\), where \(X^b \sim \mathrm{Poisson}(n)\). Let \(W^b\) denote the residual queue length after review \(b\), with \(W^0=0\). The state transition is:
$$W^{b+1}=W^b+X^b-n\cdot \mathbf{1}\{W^b+X^b\geq n\}.$$
Define \(B_{\emph{stuck}}\) as the number of stuck review slots, where \(W^b+X^b<n\). Then:
$$\mathbb{E}[W^B]= n\cdot\mathbb{E}[B_{\emph{stuck}}].$$
\end{lemma}
\noindent The proof is in Appendix~\ref{appendix_lemma::proof of throughput gap, single}. This lemma connects queue accumulation to missed threshold removals in the embedded review process. To bound $\mathbb{E}[W^B]$, we construct a coupled dominating process that admits tractable analysis via classical queueing techniques.

\begin{lemma}[Coupled Dominating Process]\label{lem:coupled_process, single}
The coupled process starts with a safety buffer (\(2n\) versus 0) and maintains dominance throughout. Define \(\{\tilde{W}^b\}\) by:
$$
\tilde{W}^0 = 2n, \quad \tilde{W}^{b+1} = \max\{ 2n, \tilde{W}^b + X^b - n \}.
$$
Then \(\tilde{W}^b \geq W^b + n\) for all \(b\in \mathbb{N}\).
\end{lemma}
\noindent The proof is in Appendix~\ref{appendix_lemma::proof of coupled_process, single}.

\paragraph{Lindley recursion.} To analyze the coupled process, we apply the Lindley recursion representation \citep[Proposition 6.3]{asmussen2003applied}:
\begin{lemma}[Lindley Representation]\label{lem:lindley_rep}
Let \(S^k=\sum_{r=1}^k (X^r-n)\) be the partial sum process. Then:
$$\tilde{W}^B=2n+\max(S^B, S^B-S^1,\ldots,S^B-S^{B-1},0).$$
\end{lemma}
\noindent This classical queueing representation is made applicable by the WAIT policy's threshold design. Define \(\xi^k = X^k-n\). We use time-reversal symmetry to simplify the maximum expression:
$$\max(S^B, S^B-S^1,\ldots,S^B-S^{B-1},0)=\max_{1\leq k\leq B}\Big(\sum_{r=k}^{B}\xi^r\Big)^+\overset{d}{=}\max_{1\leq k\leq B}\Big(\sum_{r=1}^{k}\eta^r\Big)^+, \quad \eta^r\overset{d}{=}X^1-n.$$
Thus \(\tilde{W}^B \overset{d}{=}2n+\max_{1\leq k\leq B}(\sum_{r=1}^{k}\eta^r)^+\).

Since \(\tilde{W}^B \geq W^B + n\), we have \(\mathbb{E}[W^B]\leq \mathbb{E}[\tilde{W}^B]-n\). Let \(Y^B = \max_{1\leq k\leq B}(\sum_{r=1}^{k}\eta^r)^+\). By \cite[Lemma 2]{strait1974maximum}:
\begin{lemma}[Random Walk Maximum]\label{lem:Y_T}
$$\ex{}{Y^B}=\ex{}{\max_{1\leq k\leq B}(S^k)^+}=\sum_{k=1}^B \frac{1}{k}\ex{}{(S^k)^+}.$$
\end{lemma}
\noindent This standard result \citep{strait1974maximum} decomposes the maximum into a sum over batch indices. For \(\mathbb{E}[(S^k)^+]\), the Cauchy-Schwarz inequality and \(\mathrm{Var}(S^k)=nk\) imply \(\mathbb{E}[(S^k)^+] \leq \sqrt{nk}\). Hence:
$$\mathbb{E}[Y^B] \leq \sqrt{n} \sum_{k=1}^B k^{-1/2}=O(\sqrt{nB}).$$

\begin{lemma}[Throughput Gap Bound]\label{lemma::throughput gap}
$$\mathbb{E}[W^B]-n\leq \mathbb{E}[Y^B]\leq  \sqrt{n} \sum_{k=1}^B \frac{1}{\sqrt{k}}=O(\sqrt{nB}).$$
\end{lemma}
\noindent By Lemma~\ref{lemma::throughput gap, single}, the expected number of stuck review slots is \(\mathbb{E}[B_{\emph{stuck}}]=\mathbb{E}[W^B]/n=O(\sqrt{B})\), so the normalized throughput loss in the embedded review process is \(O(B^{-1/2})\). The same bound applies to every prefix \(b\le B\), and therefore
\[
    \frac{1}{B}\sum_{b=1}^B \mathbb{E}[W^b] \leq \frac{1}{B}\sum_{b=1}^B O(\sqrt{b}) = O(\sqrt{B}).
\]
This time-averaged backlog bound is the quantity used for the latency and TTFT estimates.

\paragraph{Scaling with the horizon.} Let \(\Delta^\pi\) denote the fixed processing time of a threshold-sized batch under policy \(\pi\). In the scaled system, the corresponding processing time is \(\Delta^\pi/\zeta\). Over a calendar horizon \(T\), set
\[
    B_\zeta=\left\lfloor \frac{\zeta T}{\Delta^\pi}\right\rfloor .
\]
Then \(B_\zeta=\zeta T/\Delta^\pi+O(1)\). The terminal backlog bound gives normalized throughput loss \(O(B_\zeta^{-1/2})=O((\zeta T)^{-1/2})\). For delay, the theorem reports service-normalized time, so one scaled threshold-batch interval has length \(\Delta^\pi\); the time-averaged backlog bound therefore gives latency and TTFT contributions \(O(B_\zeta^{1/2})=O((\zeta T)^{1/2})\). The bounded number of service stages is lower order. The event-driven WAIT clock is connected to this review-process calculation by the cumulative service-opportunity coupling below, which includes the single-type case as \(m=1\).

\subsubsection{Multiple-Type Case}

We now extend to \(m\) known output-length classes. The additional issue is that event-driven WAIT need not process the full class composition in every realized batch: if only a subset of class queues has reached its threshold, only that subset is batched. The proof handles this by introducing an auxiliary embedded full-threshold process observed at deterministic review epochs. The actual policy is the event-driven WAIT policy from the main text, which batches all currently eligible types whenever the server becomes idle. The embedded process is not used to compare the total number of realized batches or the total fixed overhead paid by the two systems. Its role is to put each type-\(j\) threshold queue on a common full-threshold service scale, obtain a Lindley-type recursion, and then transfer the resulting bounds back to the actual policy through a sample path coupling of service opportunities.

For a threshold vector \(\pi=[n_1,\ldots,n_m]\), let
\begin{equation}\label{eq:batch_time_def}
\Delta^\pi \equiv \Delta T_{[1,\ldots,m]}
    =  d_0 + d_1 M^\pi,\qquad
M^\pi = \sum_{j=1}^m n_j(l_j^\prime+1)\left( l_j+\frac{l_j^\prime}{2} \right).
\end{equation}
This is the processing time of the full threshold composition, i.e., the batch that contains \(n_j\) prompts from every decode stage of every type \(j\). The load-balance and memory conditions are
\begin{equation}\label{eq:policy_constraints}
\begin{aligned}
&\lambda_j\Delta^\pi \leq n_j \quad \text{for all } j \in [m], \\
&M^{*}\leq M^{\pi}\leq C.
\end{aligned}
\end{equation}
The first line says that, over one full-threshold iteration, the expected type-\(j\) arrivals do not exceed the \(n_j\) type-\(j\) prompts that can be advanced from each active stage. The second line is the corresponding feasibility condition: the threshold composition is at least on the scale of the fluid memory requirement but does not exceed the physical memory \(C\).

\begin{lemma}[GPU-resident memory invariant]\label{lem:wait_resident_memory_invariant}
Starting from the empty system, WAIT maintains
\[
    N_{js}(t)\le n_j,\qquad j\in[m],\; s=1,\ldots,l_j^\prime,
\]
where \(N_{js}(t)\) is the number of resident type-\(j\) prompts at decode stage \(s\). Stage-0 prompts that have not been selected for prefill have no resident KV cache.
\end{lemma}
\noindent The proof is by induction over batch completions. If type \(j\) is not served in a realized batch, its GPU-resident prompts at decode stages are unchanged. If type \(j\) is served, the invariant before the batch implies that each decode stage contains at most \(n_j\) prompts, so WAIT selects all prompts currently present in that stage. After the batch completes, stage \(s+1\) contains only the prompts advanced from stage \(s\), again at most \(n_j\). Thus no decode stage can exceed \(n_j\) resident prompts.

The invariant gives the memory-feasibility statement needed by the policy. If \(a_{j0}(t)\le n_j\) denotes the number of type-\(j\) prompts selected for prefill in the current batch, then at any decision epoch,
\[
    \sum_{j=1}^m\sum_{s=1}^{l_j^\prime}N_{js}(t)(l_j+s)
    +
    \sum_{j=1}^m a_{j0}(t)l_j
    \le
    \sum_{j=1}^m n_j\sum_{s=0}^{l_j^\prime}(l_j+s)
    =
    M^\pi
    \le C .
\]
Thus unselected resident prompts are included in the memory accounting, while unselected stage-0 prompts remain outside the GPU KV-cache state. The same bound holds after the simultaneous completion update: selected stage-\(s\) prompts replace selected stage-\(s+1\) prompts, while final-stage prompts complete and release memory. Thus the invariant is checked consistently at launch epochs, during the selected batch, and after the batch-completion transition.

For any realized event-driven WAIT batch \(r\), let \(a^{(r)}_{js}\le n_j\) be the number of selected type-\(j\) prompts at decode stage \(s\), and let \(J_r\) be the set of selected types. The selected batch composition has size
\[
    M_r
    =
    \sum_{j\in J_r}\sum_{s=0}^{l_j^\prime} a^{(r)}_{js}(l_j+s)
    \le
    \sum_{j=1}^m n_j\sum_{s=0}^{l_j^\prime}(l_j+s)
    =
    M^\pi.
\]
Therefore the physical processing time of the realized batch in the \(\zeta\)-scaled system is
\[
    S_r^{(\zeta)}
    =
    \frac{d_0+d_1M_r}{\zeta}
    \le
    \frac{\Delta^\pi}{\zeta}.
\]
This bound concerns processing time, while Lemma~\ref{lem:wait_resident_memory_invariant} gives memory feasibility. Actual batches may contain fewer types and have shorter duration, but no realized WAIT batch is larger than the full-threshold composition. The fixed overhead \(d_0\) is therefore handled through a calendar-time coupling of cumulative service opportunities, not by matching realized batches one for one. If several same-type threshold batches accumulate during one busy period, later same-type batches may wait behind earlier same-type batches. The coupling below therefore compares cumulative type-wise service counts at deterministic review epochs.

In the \(\zeta\)-scaled system, one full-threshold review interval has physical length \(\Delta^\pi/\zeta\). Define deterministic review epochs
\[
    u_b=b\Delta^\pi/\zeta,\qquad b=0,1,2,\ldots .
\]
The embedded process reviews the entry queues only at these epochs. We use an end-of-interval accounting convention: arrivals accumulate over \((u_b,u_{b+1}]\), and at \(u_{b+1}\) the embedded process removes one type-\(j\) entry batch of size \(n_j\) whenever the type-\(j\) entry queue has reached \(n_j\). The corresponding type-\(j\) service opportunity also advances up to \(n_j\) resident type-\(j\) prompts from each active decode stage. Equivalently, one can view the selected work as occupying the following full review interval. A batch removed at review \(u_{b+1}\) can therefore shift completion accounting by at most \(l_j^\prime+1\) embedded intervals for type \(j\), which contributes only \(O((\zeta T)^{-1})\) to normalized throughput and \(O(1)\) to service-normalized average delay. The full-threshold interval length is feasible by the resident memory invariant above, and it is deliberately chosen as the service scale at which the type-wise threshold queues have Lindley recursions.

Throughout this argument, \(b\) indexes embedded full-threshold review intervals, not realized event-driven WAIT batches.

\begin{lemma}[Type-wise sample path coupling of service opportunities]\label{lem:wait_one_slot_dominance}
Couple actual event-driven WAIT and the auxiliary embedded full-threshold process on the same arrival sample path, both starting empty. Let \(L_\zeta=\Delta^\pi/\zeta\) and \(u_b=bL_\zeta\). For type \(j\), let \(R_j^\pi(t)\) be the cumulative number of type-\(j\) threshold batches of size \(n_j\) started by actual WAIT by time \(t\), and let \(R_j^{\bar\pi}(b)\) be the cumulative number of type-\(j\) batches removed from the embedded entry queue at reviews \(u_1,\ldots,u_b\). Then
\[
    R_j^\pi(u_{b+1})\ge R_j^{\bar\pi}(b),
    \qquad b\ge0,\; j\in[m].
\]
For downstream stages, each corresponding one-stage advancement has the same start-count lag and completes within one additional full-threshold interval.
\end{lemma}
\noindent To prove the entry-batch dominance, fix a type \(j\) and define the actual type-\(j\) entry residual
\[
    Q_j^\pi(t)=A_j(0,t]-n_jR_j^\pi(t).
\]
This identity follows because actual WAIT removes exactly one type-\(j\) entry threshold group whenever it launches a batch containing type \(j\). We prove
\[
    R_j^\pi(u_{b+1})\ge R_j^{\bar\pi}(b)
\]
by induction on \(b\). The claim is trivial for \(b=0\). Assume \(R_j^\pi(u_b)\ge R_j^{\bar\pi}(b-1)\). If the embedded process does not remove a type-\(j\) entry batch at review \(u_b\), then \(R_j^{\bar\pi}(b)=R_j^{\bar\pi}(b-1)\), and the induction hypothesis and monotonicity of \(R_j^\pi\) give the claim. Suppose instead that the embedded process removes one type-\(j\) entry batch at \(u_b\). If \(R_j^\pi(u_b)\ge R_j^{\bar\pi}(b)\), the claim again follows by monotonicity. Otherwise, since all counts are integer and \(R_j^{\bar\pi}(b)=R_j^{\bar\pi}(b-1)+1\), the induction hypothesis implies
\[
    R_j^\pi(u_b)=R_j^{\bar\pi}(b-1).
\]
The embedded removal at \(u_b\) means
\[
    A_j(0,u_b]-n_jR_j^{\bar\pi}(b-1)\ge n_j.
\]
Therefore \(Q_j^\pi(u_b)\ge n_j\), so type \(j\) is eligible in the actual system at time \(u_b\). If the server is idle, WAIT starts a batch including type \(j\) immediately. If the server is busy, the current realized batch completes by \(u_{b+1}\), because every realized batch has duration at most \(L_\zeta\), and the next WAIT launch includes all eligible types, including \(j\), unless type \(j\) has already been served. Hence
\[
    R_j^\pi(u_{b+1})\ge R_j^{\bar\pi}(b).
\]
This proves the cumulative dominance without asserting that each individual same-type threshold batch starts within one full-threshold interval of its own eligibility time. The argument is in calendar time and includes the fixed overhead paid by any subset batch.

The stage-advancement statement follows from the same sample path coupling of service opportunities, with completed one-stage advancements rather than entry starts as the induction variable. A type-\(j\) service opportunity that starts by \(u_{b+1}\) completes no later than \(u_{b+2}\), because its processing time is at most one full-threshold interval. Each such opportunity advances up to \(n_j\) prompts from every active type-\(j\) stage. Propagating the cumulative dominance stage by stage gives only a type-dependent constant review-interval lag across the fixed \(l_j'\) downstream stages. Equivalently, if \(R_{js}^{\pi}(t)\) and \(R_{js}^{\bar\pi}(b)\) count completed type-\(j\) advancements from stage \(s\) to \(s+1\) in the actual and embedded systems, then for each fixed \(j\) and \(s\) there is a constant \(c_{js}\), depending only on \(l_j'\), such that
\[
    R_{js}^{\pi}(u_{b+c_{js}})\ge R_{js}^{\bar\pi}(b).
\]

For the delay conversion, the same coupling gives an entry-queue area comparison. For \(t\in(u_b,u_{b+1}]\), the case \(b=0\) is immediate, and for \(b\ge1\), Lemma~\ref{lem:wait_one_slot_dominance} gives \(R_j^\pi(t)\ge R_j^\pi(u_b)\ge R_j^{\bar\pi}(b-1)\). Since
\[
    A_j(0,u_b]-n_jR_j^{\bar\pi}(b-1)\le \bar W^b_{(j)}+n_j,
\]
we have
\[
    Q_j^\pi(t)
    \le
    \bar W^b_{(j)}+n_j+A_j(u_b,u_{b+1}].
\]
Therefore, for any \(B\),
\[
    \mathbb E\!\left[
    \zeta\int_0^{u_B} Q_j^\pi(t)\,dt
    \right]
    \le
    \Delta^\pi\sum_{b=0}^{B-1}\mathbb E[\bar W^b_{(j)}]
    +O(B).
\]

Lemma~\ref{lem:wait_one_slot_dominance} is the structural step that transfers the embedded-process queue bounds back to event-driven WAIT. It does not require the actual system to use fewer batches than the embedded process. Actual WAIT may realize different subset compositions across time, and same-type threshold batches can accumulate during a busy period, but the cumulative type-wise removals of the embedded process are matched by actual type-wise service opportunities with a constant review-interval lag. Thus it is enough to bound the embedded comparison queues: the constant lag and the bounded in-service pipeline contribute only \(O((\zeta T)^{-1})\) to normalized throughput and \(O(1)\) to the service-normalized delay metrics.

For the embedded process, let \(\bar W^b_{(j)}\) be the residual type-\(j\) entry queue after the \(b\)-th review, and let
\[
    \bar X^b_{(j)}
    =A_j(u_b,u_{b+1}]
    \sim \text{Poisson}\!\left(\zeta\lambda_j\frac{\Delta^\pi}{\zeta}\right)
    =\text{Poisson}(\lambda_j\Delta^\pi).
\]
The arrivals \(\{\bar X^b_{(j)}\}\) are independent across embedded intervals and do not depend on other type queues. With the end-of-interval accounting convention above, the residual entry queue evolves as
\[
    \bar W^{b+1}_{(j)}
    =
    \bar W^b_{(j)}+\bar X^b_{(j)}
    -
    n_j\,\mathbf{1}\{\bar W^b_{(j)}+\bar X^b_{(j)}\ge n_j\}.
\]
This is the point at which the batch-composition issue is reduced to a stochastic comparison: the only coupling across types is the deterministic feasibility condition \(M^\pi\le C\), while the embedded process gives each type its own threshold queue.

\begin{lemma}[Embedded comparison queue bound]\label{lemma::multiple-type coupled dominace}
For each type \(j\), define
\[
\widetilde W_{(j)}^0=2n_j,\qquad
\widetilde W_{(j)}^{b+1}
=\max\{2n_j,\widetilde W_{(j)}^b+\bar X^b_{(j)}-n_j\}.
\]
Then \(\widetilde W^b_{(j)}\ge \bar W^b_{(j)}\) for all \(b\ge0\) and \(j\in[m]\).
\end{lemma}
\noindent The proof is in Appendix~\ref{appendix_lemma::proof of multiple-type coupled dominace}. Let \(\mu_j=\lambda_j\Delta^\pi\). The increment in the dominating process is \(\bar X^b_{(j)}-n_j\), whose mean is \(\mu_j-n_j\le0\) by~\eqref{eq:policy_constraints}. If \(\mu_j=n_j\), the same random-walk maximum argument used in the single-type proof gives
\[
    \mathbb{E}[\bar W^B_{(j)}]
    \le \mathbb{E}[\widetilde W^B_{(j)}]
    =O(B^{1/2}).
\]
The same bound holds for every prefix \(b\le B\), so
\[
    \frac{1}{B}\sum_{b=0}^{B-1}\mathbb{E}[\bar W^b_{(j)}]=O(B^{1/2}).
\]
If \(\mu_j<n_j\), the dominating process has strictly negative drift. Since
\[
\left|\mathbb{E}[\bar X^b_{(j)}-n_j]\right|=n_j-\mu_j,
\]
by Kingman's bound \citep{kingman1962queues}:
\[
    \sup_{B\ge0}\mathbb{E}[\bar W^B_{(j)}]
    \le
    2n_j+
    \frac{\mathbb{E}[(\bar X^b_{(j)}-n_j)^2]}{2(n_j-\mu_j)}
    <\infty .
\]
Combining the zero-drift and negative-drift cases, under the nonpositive drift condition in~\eqref{eq:policy_constraints},
\[
    \sum_{j=1}^m \mathbb{E}[\bar W^B_{(j)}]=O(B^{1/2}),
    \qquad
    \frac{1}{B}\sum_{b=0}^{B-1}\sum_{j=1}^m \mathbb{E}[\bar W^b_{(j)}]=O(B^{1/2}).
\]
Under strict slack \(\mu_j<n_j\) for all \(j\), both quantities are \(O(1)\). If equality holds for at least one type and strict inequality holds for the others, the zero-drift types dominate the aggregate bounds, so the first part of the theorem applies. Types with zero arrival rates can be omitted from \([m]\).

It remains to translate these \(B\)-interval bounds into the theorem's finite-horizon metrics. The terminal queue bound and the prefix-average queue bound play different roles: the former controls missed completions at the horizon, while the latter controls accumulated waiting. The number of completed embedded full-threshold intervals in the physical horizon \([0,T]\) is exactly
\[
    B_\zeta(T)=\max\{b:u_b\le T\}
    =\left\lfloor\frac{\zeta T}{\Delta^\pi}\right\rfloor,
\]
and the residual time is less than one embedded full-threshold interval. This is the number of deterministic review intervals in the embedded process, not the number of actual event-driven WAIT batches. Since \(B_\zeta=\zeta T/\Delta^\pi+O(1)\), \(B_\zeta^{-1/2}=O((\zeta T)^{-1/2})\) and \(B_\zeta^{-1}=O((\zeta T)^{-1})\).

For throughput, the terminal entry backlog controls the completion shortfall. More precisely, for the embedded process,
\[
    \throughput^*-\mathbb{E}[\throughput^{(\zeta,\bar\pi)}]
    \le
    \frac{1}{\zeta T}\sum_{j=1}^m l_j^\prime\,
    \mathbb{E}\!\left[
    \bar W^{B_\zeta}_{(j)}
    +A_j(u_{B_\zeta},T]
    +(l_j^\prime+2)n_j
    \right].
\]
The second term accounts for final partial-interval arrivals, and the last term absorbs the bounded in-service pipeline, startup effects, and the one-interval lag from Lemma~\ref{lem:wait_one_slot_dominance}. Since \(T-u_{B_\zeta}<\Delta^\pi/\zeta\), \(\mathbb{E}[A_j(u_{B_\zeta},T]]\le \lambda_j\Delta^\pi=O(1)\). Thus the terminal-backlog term gives an \(O((\zeta T)^{-1/2})\) gap under nonpositive drift and an \(O((\zeta T)^{-1})\) gap under strict slack.

For delay, we use the prefix-average queue bound instead. Let \(\bar Q^b=\sum_{j=1}^m \bar W^b_{(j)}\). In physical time, each embedded full-threshold interval has length \(\Delta^\pi/\zeta\). The theorem, however, reports service-normalized delay, multiplying elapsed physical time by \(\zeta\). Thus each embedded interval contributes a constant scaled time \(\Delta^\pi\). The entry-area comparison above gives
\[
    \mathbb E\!\left[
    \zeta\int_0^T\sum_{j=1}^m Q_j^\pi(t)\,dt
    \right]
    \le
    \Delta^\pi\sum_{b=0}^{B_\zeta-1}\mathbb E[\bar Q^b]+O(B_\zeta),
\]
where the \(O(B_\zeta)\) term includes final partial intervals and the expected arrivals inside each review interval. The remaining latency contribution comes from prompts that have already entered the GPU-resident pipeline. This part is uniformly bounded in area: by Lemma~\ref{lem:wait_resident_memory_invariant},
\[
    \zeta\int_0^T\sum_{j=1}^m\sum_{s=1}^{l_j'}N_{js}(t)\,dt
    \le
    \zeta T\sum_{j=1}^m n_jl_j'
    =
    O(B_\zeta).
\]
Selected stage-0 prompts currently in prefill service and selected decode prompts currently in service are also bounded by \(\sum_j n_j(l_j'+1)\) at every time, and hence contribute another \(O(B_\zeta)\) service-normalized area. Equivalently, after entry admission, the fixed number \(l_j'\) of type-\(j\) stage advancements and the constant review-interval lag in Lemma~\ref{lem:wait_one_slot_dominance} add only \(O(1)\) to the average service-normalized latency and TTFT. Under the delay convention in Section~\ref{sec:model}, expected finite-horizon delay is the expected total censored delay divided by the expected number of arrivals, which is \(\Theta(B_\zeta)\). Therefore
\[
    \mathbb{E}[\Latency^{(\zeta,\pi)}],
    \mathbb{E}[\TTFT^{(\zeta,\pi)}]
    \le
    O\!\left(
    \frac{1}{B_\zeta}\sum_{b=0}^{B_\zeta-1}
    \mathbb{E}[\bar Q^b]
    \right)+O(1).
\]
The prefix bound gives
\[
    \mathbb{E}[\Latency^{(\zeta,\pi)}],
    \mathbb{E}[\TTFT^{(\zeta,\pi)}]
    =
    O(B_\zeta^{1/2})
    =
    O((\zeta T)^{1/2})
\]
under nonpositive drift. The corresponding calendar-time delay is smaller by a factor \(1/\zeta\). Under strict slack, the same service-normalized expression is \(O(1)\). Lemma~\ref{lem:wait_one_slot_dominance} transfers the throughput bound from \(\bar\pi\) to the actual event-driven WAIT policy, while the delay estimates above use the entry-area and resident-pipeline comparisons directly for actual WAIT. This completes the proof of Theorem~\ref{thm:wait_heavy_traffic}.

%% file: pf_thm_nested.tex
\section{Proof of Theorem~\ref{thm:nested_wait_heavy_traffic}}\label{appendix::proof of nested_wait_heavy_traffic}

The proof analyzes Nested WAIT's threshold dynamics and then shows that the memory condition makes those dynamics feasible with high probability. The main difference from WAIT is that downstream arrivals are generated endogenously: a prompt reaches segment \(k\) only if it survives the boundary \(l_{k-1}'\). The proof therefore follows the queues at these segment boundaries. Segment 1 is a threshold-renewal queue with external Poisson arrivals. For \(k\ge2\), each upstream boundary-crossing opportunity produces a binomially thinned boundary batch. The threshold conditions give negative drift for these embedded queues. Standard reflected-random-walk and exponential-martingale bounds then control their expected size and high-probability excursions; the nonstandard part is the memory accounting that separates fresh boundary arrivals from carryover residual queues.

Let
\[
    \Lambda_k=\sum_{j=k}^m\lambda_j,\qquad
    p_k=\frac{\Lambda_k}{\Lambda_{k-1}},\quad k\ge2,
\]
and write
\[
    \Delta^\pi=\Delta T_{[1,\ldots,m]}(n_1,\ldots,n_m)=d_0+d_1M^\pi .
\]
The theorem assumes \(\Lambda_1\Delta^\pi<n_1\) and \(p_k<n_k/n_{k-1}<1\) for \(k\ge2\). In the \(\zeta\)-scaled system, every realized iteration has physical duration at least \(d_0/\zeta\), so the number of realized processing opportunities over \([0,T]\) is at most
\[
    B_\zeta^{\max}=\left\lceil \frac{\zeta T}{d_0}\right\rceil
    \le 1+\frac{\zeta T}{d_0}.
\]
This is an upper bound on actual event opportunities, not a full-threshold comparison-slot count.

\subsection{Segment-1 Entry Queue}

Prompts waiting to enter segment 1 have not been prefilled, so they are outside the GPU-resident KV-cache state. The only subtlety is that the server may idle while waiting for the segment-1 threshold. We therefore index the segment-1 queue by threshold renewals rather than by realized processing iterations. Let \(R_q\) be the residual number of not-yet-prefilled prompts immediately after the \(q\)-th segment-1 service start, after \(n_1\) prompts have been removed for prefill. During the following service period, the number of external arrivals is stochastically dominated by
\[
    \bar Y^q_{(1)}\sim \mathrm{Poisson}(\Lambda_1\Delta^\pi),
\]
because the physical service period is at most \(\Delta^\pi/\zeta\) and arrivals occur at rate \(\zeta\Lambda_1\). If the residual plus these arrivals is below \(n_1\), the system idles until the next arrivals complete the threshold and then immediately starts the next segment-1 service. Thus
\[
    R_{q+1}\le (R_q+\bar Y^q_{(1)}-n_1)^+ .
\]
This recursion is dominated by the reflected random walk
\[
    \widetilde R^0_{(1)}=2n_1,\qquad
    \widetilde R^{q+1}_{(1)}
    =
    \max\{2n_1,\widetilde R^q_{(1)}+\bar Y^q_{(1)}-n_1\}.
\]
Since \(\mathbb{E}[\bar Y^q_{(1)}]-n_1=\Lambda_1\Delta^\pi-n_1<0\), the standard Lindley-recursion bound for a random walk with negative drift \citep[see, e.g.,][]{kingman1962queues,asmussen2003applied} gives
\[
    \sup_{q\ge0}\mathbb{E}[R_q]
    \le
    2n_1+
    \frac{\mathbb{E}[(\bar Y^q_{(1)}-n_1)^2]}{2(n_1-\Lambda_1\Delta^\pi)}
    <\infty .
\]
The idle time before the next segment-1 service is the time needed to collect fewer than \(n_1\) arrivals, whose expected physical length is \(O(1/\zeta)\); in service-normalized time this contributes \(O(1)\).

\subsection{Boundary Queues for Downstream Segments}

For \(k\ge2\), consider an opportunity at which segment \(k-1\) advances prompts across boundary \(l_{k-1}'\). Let \(A^r_{k-1}\le n_{k-1}\) be the number of prompts that cross this boundary at the \(r\)-th such opportunity. Conditional on the filtration just before output-length revelation at the boundary, the indicators that these prompts continue to segment \(k\) are independent Bernoulli random variables with probability \(p_k=\Lambda_k/\Lambda_{k-1}\). This uses the i.i.d. output-length mix and the fact that service before the boundary is nonanticipating with respect to future output length; before a boundary is reached, prompts with possible remaining output-length classes are exchangeable within the surviving mixture. Therefore
\[
    Y^r_{(k)}\mid A^r_{k-1}\sim \mathrm{Binomial}(A^r_{k-1},p_k),
    \qquad
    Y^r_{(k)}\preceq \bar Y^r_{(k)},
\]
where the dominating variables \(\bar Y^r_{(k)}\sim\mathrm{Binomial}(n_{k-1},p_k)\) can be chosen independent across boundary opportunities.

Let \(V^r_{(k)}\) be the carryover boundary queue at stage \(l_{k-1}'\) after the threshold review at the start of the \(r\)-th opportunity in which segment \(k-1\) is processed. Thus \(V^r_{(k)}\) is a post-review residual, not the pre-review boundary population. This convention preserves the nested, prefix structure of the algorithm. Whenever segment \(k-1\) is active, all earlier segment thresholds have been met. If the boundary queue for segment \(k\) also contains at least \(n_k\) prompts at that review, Nested WAIT includes segment \(k\) in the same prefix batch and removes a threshold batch from the boundary. During the batch, at most \(A^r_{k-1}\le n_{k-1}\) additional prompts may survive segment \(k-1\) and join the boundary. These fresh survivors are part of the just-processed upstream threshold batch, so their KV-cache memory is charged to the threshold-sized batch term \(M^\pi\). Only the carryover residual remains as additional boundary memory. With the binomial domination above, the pre-review queue for the next opportunity is \(V^r_{(k)}+\bar Y^r_{(k)}\), and the post-review residual satisfies
\[
    V^{r+1}_{(k)}
    \le
    V^r_{(k)}+\bar Y^r_{(k)}
    -
    n_k{\bf 1}\{V^r_{(k)}+\bar Y^r_{(k)}\ge n_k\}.
\]
The residual queue is dominated by
\[
    \widetilde W^0_{(k)}=n_k,\qquad
    \widetilde W^{r+1}_{(k)}
    =
    \max\{n_k,\widetilde W^r_{(k)}+\bar Y^r_{(k)}-n_k\}.
\]
This domination follows by induction over boundary-crossing opportunities. If \(V^r_{(k)}+\bar Y^r_{(k)}\ge n_k\), the post-review residual is at most \(V^r_{(k)}+\bar Y^r_{(k)}-n_k\), which is bounded by \(\widetilde W^r_{(k)}+\bar Y^r_{(k)}-n_k\). If \(V^r_{(k)}+\bar Y^r_{(k)}<n_k\), the residual is below \(n_k\), while the dominating recursion is always at least \(n_k\). Hence \(V^r_{(k)}\le\widetilde W^r_{(k)}\) for all \(r\).

The increments \(\bar Y^r_{(k)}-n_k\) have strictly negative drift because
\[
    \mathbb{E}[\bar Y^r_{(k)}-n_k]=n_{k-1}p_k-n_k<0.
\]
Applying the same negative-drift reflected-random-walk bound yields, uniformly in \(r\),
\[
    \mathbb{E}[V^r_{(k)}]
    \le
    n_k+
    \frac{\mathbb{E}[(\bar Y^r_{(k)}-n_k)^2]}{2(n_k-n_{k-1}p_k)}
    <\infty,\qquad k=2,\ldots,m.
\]
Thus all external and downstream boundary queues have bounded expected size, uniformly over \(\zeta\) and \(T\).

We also use the stage-cap invariant induced by the threshold rule. Starting from the empty system, every threshold-controlled interior stage of segment \(k\) contains at most \(n_k\) resident prompts. A batch that includes segment \(k\) advances at most \(n_k\) prompts from each preceding stage into the next stage; a batch that excludes segment \(k\) leaves its interior stages unchanged. The only resident queues not covered by this interior-stage cap are the downstream boundary queues \(V_{(k)}\), \(k\ge2\), which are controlled by the residual-queue bounds above.

\subsection{Throughput and Delay}

We first translate the queue bounds into performance estimates for the threshold-controlled dynamics before memory repair is invoked. Let \(W_{(1)}(T)\) denote the segment-1 entry residual at time \(T\), let \(V_{(k)}(T)\) be the downstream resident boundary queue for \(k\ge2\), and let \(I_k(T)\) be the number of prompts inside the threshold-controlled interior of segment \(k\). The threshold rule gives \(I_k(T)\le n_k\delta_k\), where \(\delta_k\) is the number of interior stages in segment \(k\). There are constants \(\alpha_k,\beta_k\), depending only on the fixed output lengths, such that terminal unfinished decode work is bounded by
\[
    \sum_{k=1}^m \alpha_k W_{(k)}(T)
    +
    \sum_{k=1}^m \beta_k I_k(T)
    +
    A_{\mathrm{tail}}(T),
\]
where \(W_{(k)}=V_{(k)}\) for \(k\ge2\), and \(A_{\mathrm{tail}}(T)\) accounts for arrivals in the final partial opportunity. The expected final-partial-opportunity arrivals are \(O(1)\) because that opportunity has physical length at most \(\Delta^\pi/\zeta\). Consequently,
\[
\begin{aligned}
    \throughput^*-\mathbb{E}[\throughput^{(\zeta,\pi)}]
    &\le
    \frac{1}{\zeta T}
    \mathbb{E}\!\left[
    \sum_{k=1}^m \alpha_k W_{(k)}(T)
    +
    \sum_{k=1}^m \beta_k I_k(T)
    +
    A_{\mathrm{tail}}(T)
    \right]  \\
    &= O((\zeta T)^{-1}).
\end{aligned}
\]

For delay, let \(Q(t)\) be the total number of prompts in the segment-1 entry queue, downstream boundary queues, and threshold-controlled segment interiors. The bounds above, together with the \(O(1/\zeta)\) expected idle period needed to collect fewer than a threshold batch, imply
\[
    \mathbb{E}\!\left[\int_0^T Q(t)\,dt\right]=O(T).
\]
The expected number of arrivals over \([0,T]\) is \(\zeta\Lambda_1T=\Theta(\zeta T)\). Hence the calendar-time censored average latency and TTFT are \(O(1/\zeta)\). The theorem reports service-normalized delays, \(\Latency^{(\zeta,\pi)}=\zeta\Latency_{\rm cal}^{(\zeta,\pi)}\) and \(\TTFT^{(\zeta,\pi)}=\zeta\TTFT_{\rm cal}^{(\zeta,\pi)}\), so
\[
    \mathbb{E}[\Latency^{(\zeta,\pi)}],
    \mathbb{E}[\TTFT^{(\zeta,\pi)}]
    =
    O(1).
\]

\subsection{High-Probability Memory Bound}

The base term
\[
    M^\pi=\sum_{k=1}^m n_k(l+\bar s_k)\delta_k
\]
accounts for threshold-sized segment batches. With the convention used in Theorem~\ref{thm:nested_wait_heavy_traffic}, segment 1 covers stages \(s=0,\ldots,l_1'\), while for \(k\ge2\), the resident boundary queue at \(s=l_{k-1}'\) is tracked separately and segment \(k\)'s interior covers \(s=l_{k-1}'+1,\ldots,l_k'\). The memory accounting separates two boundary terms. Fresh survivors that have just crossed from segment \(k-1\) are a subset of the upstream threshold batch. They have the same KV-cache size as the prompts charged to the \(s=l_{k-1}'\) stage of segment \(k-1\) in \(M^\pi\), so charging them to the base term does not add memory. The only additional memory is the carryover residual left at downstream boundaries across opportunities. Therefore, if \(V^r_{(k)}\le c_k\) for all downstream boundaries, then total resident KV-cache memory is at most
\[
    M^\pi+\sum_{k=2}^m(l+l_{k-1}')c_k .
\]
The deterministic \(n_k\) component of \(c_k\) is on the same scale as the base term: since \(n_k<n_{k-1}\), \((l+l_{k-1}')n_k\) is bounded by the contribution of the terminal stage \(s=l_{k-1}'\) of segment \(k-1\) already included in \(M^\pi\).

Fix \(k\ge2\). Since \(p_k<n_k/n_{k-1}<1\), Appendix~\ref{appendix_lemma::proof of solution to martingale exists} records the standard exponential-martingale construction for binomial increments and gives a unique \(\theta_k>0\) satisfying
\[
    e^{-\theta_k n_k}(1-p_k+p_ke^{\theta_k})^{n_{k-1}}=1.
\]
For \(S^i_{(k)}=\sum_{r=1}^i(\bar Y^r_{(k)}-n_k)\), the process \(e^{\theta_k S^i_{(k)}}\) is a martingale. Doob's maximal inequality \citep[see, e.g.,][]{durrett2019probability} gives
\[
    \mathbb{P}\!\left(\max_{0\le i\le r}S^i_{(k)}\ge x\right)\le e^{-\theta_k x}.
\]
Equivalently, the finite-horizon reflected-random-walk bound in Appendix~\ref{appendix_lemma::proof of solution to martingale exists} gives, for the dominating boundary queue,
\[
    \mathbb{P}\!\left(
    \max_{0\le r\le B}\big(\widetilde W^r_{(k)}-n_k\big)\ge x
    \right)
    \le B e^{-\theta_k x},\qquad B\ge1.
\]
Since \(V^r_{(k)}\le \widetilde W^r_{(k)}\), set
\[
    c_k
    =
    n_k+\theta_k^{-1}
    \ln\!\left(\frac{m(1+\zeta T/d_0)}{\delta}\right).
\]
Applying this finite-horizon bound with \(B=B_\zeta^{\max}\), and then unioning over \(k=2,\ldots,m\), gives
\[
    \mathbb{P}\!\left(
    V^r_{(k)}\le c_k
    \text{ for all } k=2,\ldots,m,\; r\le B_\zeta^{\max}
    \right)\ge1-\delta.
\]
Let \(\mathcal E\) denote the event in the preceding display. Couple the finite-memory Nested WAIT implementation and the threshold-controlled dynamics analyzed above on the same arrivals and continuation marks until the first time memory repair would be invoked. On \(\mathcal E\), the resident memory after every projected transition is at most \(M_{\mathrm{req}}^{(\zeta,\pi)}\le C\), and actual boundary completions can only reduce the projected memory. The same accounting covers both memory-repair sites in Algorithm~\ref{alg:nested_wait} and its full event-driven version, Algorithm~\ref{alg:nested_wait_full}: after completion updates, repair is allowed only from resident boundary queues, and before launch, repair is allowed only from unselected resident boundary prompts in the conservative projected state \(Q^+\).

For the completion-update repair site, fresh survivors occupy slots already charged in \(M^\pi\), and only carryover resident boundary residuals require the \(c_k\) buffers. Under the stage-cap invariant, all non-boundary resident prompts are covered by the base term \(M^\pi\); hence on \(\mathcal E\), the post-completion resident memory is already at most \(M_{\mathrm{req}}^{(\zeta,\pi)}\le C\). For the before-launch repair site, the selected boundary prompts are charged to the first interior stage of their next segment in the projected state \(Q^+\), again within the base term. The only projected memory outside this base accounting comes from unselected resident boundary prompts, and these are exactly the residual queues controlled by the \(c_k\) buffers on \(\mathcal E\). Thus the projected memory before launch is also at most \(M_{\mathrm{req}}^{(\zeta,\pi)}\le C\). Consequently, on \(\mathcal E\), the repair conditions are never met at either repair site. The within-set eviction order is therefore immaterial on \(\mathcal E\), and this argument makes no recovery claim on \(\mathcal E^c\). Hence memory repair is never invoked, the two sample paths coincide on \([0,T]\), and
\[
    \mathbb P\!\left(
    \text{a memory-overflow-induced eviction occurs over }[0,T]
    \right)
    \le
    \mathbb P(\mathcal E^c)
    \le \delta .
\]
We also record the deterministic no-stall property of Nested WAIT's restricted repair rule under the theorem's memory assumption, in particular \(C\ge M^\pi\). The global LIFO pathology in which late-stage resident prompts remain on the GPU while newly accumulated early-stage prompts are repeatedly evicted cannot occur here. Memory repair never evicts the external queue \(Q_{1,0}\) and never evicts segment-interior prompts. It only evicts resident boundary prompts, and before launch it evicts only unselected resident boundary prompts in the projected state. By the stage-cap invariant, if all resident boundary prompts eligible for the relevant repair loop have been removed, the remaining resident memory is covered by the base threshold allocation \(M^\pi\le C\). Thus each repair loop terminates with a feasible state. Moreover, between launches of segment 1, \(Q_{1,0}\) is not reduced by repair and can only grow through exogenous arrivals and restarted boundary prompts. Hence, once the server is idle and \(Q_{1,0}\ge n_1\), the threshold review finds a nonempty prefix and memory repair cannot remove the segment-1 threshold batch; the projected before-launch repair can remove only unselected downstream boundary prompts. If the server is idle and \(Q_{1,0}<n_1\), then it is waiting for exogenous arrivals rather than resident-memory repair, and since \(\Lambda_1>0\), the threshold is reached almost surely in finite time on the continuing arrival process. This is the point where Nested WAIT's resident-boundary repair rule differs from a generic LIFO rule.

It remains to bound the contribution of \(\mathcal E^c\) to the unconditional finite-horizon performance metrics. On \(\mathcal E^c\), the finite-memory implementation may experience evictions and restarts, so it is no longer coupled to the threshold dynamics analyzed above. The deterministic no-stall property rules out a global memory-induced launch blockade at segment 1, but it does not by itself give the sharp threshold-dynamics queue bounds after restarts or for downstream segments. We therefore charge the entire post-failure contribution using pathwise finite-horizon caps. Each request that arrived before \(T\) contributes at most \(T\) calendar time to latency or TTFT, and realized effective throughput remains nonnegative. Evicted prompts are not counted as new exogenous arrivals; returning them to \(Q_{1,0}\) can only affect the finite-horizon path after the failure event, whose contribution is bounded by the estimates below. Let \(A_T\) be the total number of external arrivals over \([0,T]\). Then \(A_T\sim\mathrm{Poisson}(\mu)\), where \(\mu=\zeta\Lambda_1T\) and \(\Lambda_1=\sum_{j=1}^m\lambda_j\). For any event \(F\) with \(\mathbb P(F)\le\delta\),
\[
    \mathbb E[A_T{\bf 1}_F]
    \le
    2\mu\,\mathbb P(F)+\mathbb E[A_T{\bf 1}\{A_T>2\mu\}]
    \le
    2\mu\delta+C\mu e^{-c\mu}
\]
for constants \(C,c>0\), using the standard Poisson Chernoff bound. This bound holds for any event \(F\), with no independence requirement between \(F\) and \(A_T\). Taking \(F=\mathcal E^c\), the finite-horizon convention gives a pathwise service-normalized latency or TTFT contribution of at most \(\zeta T\) per external arrival on \(\mathcal E^c\). This uses the convention in Section~\ref{sec:model} that expected finite-horizon delay is normalized by the expected number of arrivals, not by the realized arrival count. After normalization by \(\mathbb E[A_T]=\mu\), the failure-event contribution to either delay metric is at most
\[
    \zeta T\frac{\mathbb E[A_T{\bf 1}_{\mathcal E^c}]}{\mu}
    \le
    \zeta T\left(2\delta+Ce^{-c\mu}\right)=O(1),
\]
because \(\delta\le K(1+\zeta T)^{-1}\) and \(\Lambda_1\) is fixed.

For throughput, let \(X_T\) be the effective throughput of the threshold-controlled dynamics and let \(Y_T\) be the effective throughput of the finite-memory Nested WAIT implementation. Since \(X_T=Y_T\) on \(\mathcal E\) and \(Y_T\ge0\),
\[
    \throughput^*-\mathbb E[Y_T]
    \le
    \throughput^*-\mathbb E[X_T]
    +
    \mathbb E[X_T{\bf 1}_{\mathcal E^c}].
\]
The threshold-process effective throughput over \([0,T]\) is at most \(l_m' A_T/(\zeta T)\), so the failure event adds at most
\[
    \frac{l_m'}{\zeta T}\mathbb E[A_T{\bf 1}_{\mathcal E^c}]
    =
    O(\delta)+O(e^{-c\mu})
    =
    O((\zeta T)^{-1})
\]
to the normalized throughput gap. Combining these bad-event bounds with the threshold-dynamics estimates above gives the stated unconditional throughput, latency, and TTFT bounds for the finite-memory Nested WAIT implementation.

Finally, Appendix~\ref{appendix_lemma::proof of solution to martingale exists} also gives
\[
    \theta_k\ge \frac{8(n_k-n_{k-1}p_k)}{n_{k-1}},
\]
which is the stated drift-to-buffer scaling. This completes the proof.

%% file: pf_thm_timevarying.tex
\section{Proof of Theorem~\ref{thm:nested_wait_heavy_traffic_time_varying}}\label{appendix:: proof of nested optimality, time varying.}

The proof is a conservative version of the Nested WAIT argument. Instead of matching each instantaneous arrival pattern exactly, we dominate each service-window arrival count and each downstream thinning probability by worst-case quantities over the finite horizon. This gives the same bounded-queue and logarithmic-buffer structure as the constant-rate proof. The first-segment waiting condition is used only when converting bounded queues into service-normalized delay bounds; the throughput and memory conclusions use only the upper-side window-load conditions and the finite-horizon safety buffer. The structure proceeds as follows:
\begin{enumerate}
    \item \textbf{First segment analysis}: Poisson arrivals with time-varying rates are dominated by a time-invariant process using the worst-case scaled-window arrival count \(\Lambda^{\pi}\) from Section~\ref{sec:time_varying}. We couple this queue to a Lindley process and apply Kingman's inequality.
    \item \textbf{Subsequent segments analysis}: Poisson-binomial boundary arrivals with time-varying thinning probabilities are pathwise dominated by a binomial process with worst-case continuation probability \(p_k^*\) defined in Section~\ref{sec:time_varying}. The same Lindley coupling then controls each downstream boundary queue.
    \item \textbf{Throughput, delay, and memory}: Bounded expected terminal work yields the normalized throughput gap. The service-normalized delay bound additionally uses the first-segment waiting condition to control idle waiting before the first segment. High-probability memory bounds follow from the same martingale construction with worst-case parameters.
\end{enumerate}

We use $b$ to denote the \emph{batch index} and $k \in \{1, \ldots, m\}$ to denote the \emph{segment index}. The total number of batches is denoted by $B$.

\subsection{First Segment Analysis}

For the first segment ($k=1$), we use the same threshold-renewal convention as in Appendix~\ref{appendix::proof of nested_wait_heavy_traffic}. Idle periods before a segment-1 service start contain fewer than \(n_1\) prompts, so they contribute only a bounded amount of terminal unfinished work and do not affect the memory bound. The first-segment waiting condition in Section~\ref{sec:time_varying} is needed later only to bound these idle periods in service-normalized time. At renewal epochs, the first-segment queue is dominated by the comparison recursion
$$ W_{(1)}^{b+1} = W_{(1)}^{b} + Y_{(1)}^b  - n_1 \cdot \mathbf{1}\{ W_{(1)}^{b} + Y_{(1)}^b \geq n_1 \},  \quad Y_{(1)}^b \sim \text{Poisson}\Big( \sum_{j=1}^m \mathcal A_j^{(\zeta)}(t(b),\Delta T_b) \Big),  $$
where $t(b)$ denotes the physical start time of batch $b$ and $\Delta T_b$ denotes the service-normalized processing time of that batch, so its physical duration is \(\Delta T_b/\zeta\). If the actual queue is below threshold, the server waits until enough arrivals occur and then removes \(n_1\); the displayed recursion is therefore an upper comparison for queue length, not an exact event-time identity.

\paragraph{Worst-case domination.} Time-varying arrival rates prevent exact load balance because the arrival composition changes over the horizon. We therefore use a conservative domination argument. Since $\Delta T_b\leq \Delta T_{[1,\dots,m]}$ (the service-normalized time for a full batch containing all $m$ types), the arrivals during any realized service period are dominated by arrivals over a full service-normalized window. This gives the process $\{\bar{W}_{(1)}^{b}\}$:
$$ \bar{W}_{(1)}^{b+1} = \bar{W}_{(1)}^{b} + \bar{Y}_{(1)}^b  - n_1 \cdot \mathbf{1}\{ \bar{W}_{(1)}^{b} + \bar{Y}_{(1)}^b \geq n_1 \},  \quad \bar{Y}_{(1)}^b \sim \text{Poisson}\Big(  \sum_{j=1}^m \mathcal A_j^{(\zeta)}(t(b),\Delta T_{[1,\dots,m]}) \Big).  $$
By~(\ref{eq:nested_wait_thresholds_time_varying}), the aggregate arrival count during any full service-normalized window is bounded by the worst-case quantity \(\Lambda^\pi<n_1\). This enables a time-invariant dominating process that provides tractable bounds:
$$ \tilde{W}_{(1)}^{b+1} = \tilde{W}_{(1)}^{b} + \tilde{Y}_{(1)}^b  - n_1 \cdot \mathbf{1}\{ \tilde{W}_{(1)}^{b} + \tilde{Y}_{(1)}^b \geq n_1 \},  \quad \tilde{Y}_{(1)}^b \sim \text{Poisson}(\Lambda^{\pi}).  $$
This time-invariant process enables the same queueing bound used in the constant-rate proof. Define $X_{(1)}^b = \tilde{Y}_{(1)}^b - n_1$. The process $\{ \tilde{W}_{(1)}^{b}\}$ is dominated by a Lindley process:
$$ \overline{W}_{(1)}^{0} = 2n_1,\quad  \overline{W}_{(1)}^{b+1} = \max \{  2n_1, \overline{W}_{(1)}^{b} + X_{(1)}^b \},\quad \forall b \in \mathbb{N}. $$
Thus we have the dominance chain:
$$ W_{(1)}^{b}\leq \bar{W}_{(1)}^{b} \leq \tilde{W}_{(1)}^{b} \leq \overline{W}_{(1)}^{b}, \quad \forall b \in \mathbb{N}.$$
Since $n_1> \Lambda^{\pi}$, the process $\{\overline{W}_{(1)}^{b} \}$ has negative drift. By Kingman's inequality \citep{kingman1962queues}, with
$$\text{Var}(X_{(1)}^b ) = \Lambda^{\pi},\quad \left|  \ex{}{X_{(1)}^b} \right| = n_1 - \Lambda^{\pi},$$
we obtain
$$  \ex{}{W_{(1)}^b}\leq  2n_1 + \frac{\mathbb{E}[(\tilde{Y}_{(1)}^b-n_1)^2]}{2\left(n_1 -\Lambda^{\pi}\right)}. $$
\subsection{Subsequent Segments Analysis ($k \geq 2$)}

For segment $k$ ($2\leq k\leq m$), arrivals come from prompts that completed segment $k-1$. The batch index $b$ for segment $k$ counts only batches containing segment $k-1$. These indices differ across segments but are bounded by the total batch count $B$. As in Appendix~\ref{appendix::proof of nested_wait_heavy_traffic}, \(W^b_{(k)}\) is the post-review carryover residual at the downstream boundary; fresh survivors from segment \(k-1\) are charged to the upstream threshold batch, and only residual carryover receives the extra safety buffer.

\paragraph{Time-varying thinning probabilities.} Consider the $b$-th arrival to segment $k$, which comes from the $b$-th batch containing segment $k-1$. The thinning probabilities depend on the prompt arrival times in this batch and vary over time. For a prompt that has reached boundary \(k-1\) at physical time \(s\), its conditional continuation probability is the instantaneous ratio \(\sum_{j=k}^m\lambda_j^s/\sum_{j=k-1}^m\lambda_j^s\) when the denominator is positive. Because the intensities are continuous and \(p_k^*\) is the supremum over arbitrarily short positive windows in Section~\ref{sec:time_varying}, this instantaneous ratio is at most \(p_k^*\). If the prompts in a boundary batch accumulated over a longer interval, the aggregate continuation ratio is a denominator-weighted average of positive-denominator subwindow ratios of length at most \(\Delta^\pi\), and is therefore also bounded by \(p_k^*\). Thus, conditional on the filtration before boundary revelation, the survivor indicators in a boundary batch are independent Bernoulli random variables with possibly different probabilities \(q_i\le p_k^*\). Their sum \(Y_{(k)}^b\) is Poisson-binomial and is stochastically dominated by \(\mathrm{Binomial}(n_{k-1},p_k^*)\). Equivalently, using shared uniforms \(U_i^b\), write the actual survivor indicators as \({\bf 1}\{U_i^b\le q_i\}\) and the dominating indicators as \({\bf 1}\{U_i^b\le p_k^*\}\); then \(Y_{(k)}^b\le \bar Y_{(k)}^b\) pathwise, where \(\bar Y_{(k)}^b\sim \mathrm{Binomial}(n_{k-1},p_k^*)\). By~(\ref{eq:nested_wait_thresholds_time_varying}) and the definition of \(p_k^*\),
$$p_k^* < \frac{n_k}{n_{k-1}}.$$
The boundary-arrival count therefore satisfies
$$Y_{(k)}^b \le \bar Y_{(k)}^b,\quad \bar Y_{(k)}^b\sim \text{Binomial}(n_{k-1},p_k^*),\quad \forall b\geq 0,$$
and the state transition rule is
$$W_{(k)}^{b+1} = W_{(k)}^{b} + Y_{(k)}^{b} - n_k \cdot\mathbf{1}\{ W_{(k)}^{b} + Y_{(k)}^{b}\geq n_k \}, $$
where $b$ indexes batches containing segment $k-1$, since new arrivals to segment $k$ occur only when segment $k-1$ is processed.

Define $X_{(k)}^b = Y_{(k)}^b - n_k$. Although \(Y_{(k)}^b\) is Poisson-binomial and time-varying, the coupling construction follows the same induction as Appendix~\ref{appendix_lemma::proof of nested coupled process}:
\begin{lemma}[Coupling for Time-Varying Case]\label{lemma: coupling, time varying}
Define a coupled process $\{ \tilde{W}_{(k)}^b\}$ by:
\begin{equation*}
\begin{aligned}
\tilde{W}_{(k)}^0 &= n_k, \\
\tilde{W}_{(k)}^{b+1} &= \max\{ n_k, \, \tilde{W}_{(k)}^b+X_{(k)}^b\}, \\
X_{(k)}^b &= Y_{(k)}^b - n_k.
\end{aligned}
\end{equation*}
Then $\tilde{W}^b_{(k)}\geq W^b_{(k)}$ for all $b\in \mathbb{N}$.
\end{lemma}
To apply Kingman's inequality \citep{kingman1962queues}, we construct a time-invariant dominating process using the pathwise dominating counts \(\bar Y_{(k)}^b\):
\begin{equation*}
\begin{aligned}
\bar{W}_{(k)}^0 &= n_k, \\
\bar{W}_{(k)}^{b+1} &= \max\{ n_k,\, \bar{W}_{(k)}^{b}+\bar{X}_{(k)}^b \}, \\
\bar{X}_{(k)}^b &= \bar{Y}_{(k)}^b  -n_k, \\
\bar{Y}_{(k)}^b &\sim \text{Binomial}(n_{k-1}, p_k^*), \quad \forall b\in \mathbb{N}.
\end{aligned}
\end{equation*}
Because the Lindley update is monotone in both the current state and the increment, the shared-uniform coupling gives the dominance chain:
$$\bar{W}^b_{(k)}\geq \tilde{W}^b_{(k)}\geq W^b_{(k)},\quad \forall b\in \mathbb{N}.$$
The process $\bar{W}^b_{(k)}$ has the Lindley representation:
\begin{equation*}
\begin{aligned}
\bar{W}_{(k)}^0 &= n_k, \\
\bar{W}_{(k)}^{b+1} &= n_k + \max_{0\leq i\leq b}\{ \bar{S}_{(k)}^i \}, \\
\text{where } \bar{S}_{(k)}^0 &= 0,\; \bar{S}_{(k)}^i = \sum_{r=1}^i \bar{X}_{(k)}^{r},\; 1\leq i\leq b.
\end{aligned}
\end{equation*}
Since
$$ \ex{}{\bar{X}_{(k)}^b} = n_{k-1}\cdot p_{k}^* - n_k <0,\quad \forall b\in \mathbb{N}, $$
the coupled process $\{\bar{W}_{(k)}^b \}$ has negative drift. With
$$\text{Var}(\bar{X}_{(k)}^b) = n_{k-1}p_k^*(1-p_k^*),\quad \left|\ex{}{\bar{X}_{(k)}^b}\right|=n_k - p_k^* n_{k-1},$$
Kingman's inequality \citep{kingman1962queues} yields:
$$\ex{}{\max_{0\leq i\leq b}\{ \bar{S}_{(k)}^i \}  }\leq  \frac{  n_{k-1}p_k^*(1-p_k^*) }{ 2(n_k - n_{k-1}p_k^*)}. $$
Thus, the expected queue length is bounded:
$$\ex{}{W_{(k)}^b}\leq \ex{}{\tilde{W}_{(k)}^b}\leq \ex{}{\bar{W}_{(k)}^b} \leq n_k + \frac{\mathbb{E}[(\bar{Y}_{(k)}^b-n_k)^2]}{2(n_k - n_{k-1}p_k^*)}.$$
\subsection{Throughput, Delay, and High-Probability Bounds}

The bounded queue estimates above imply that the expected terminal unfinished work is \(O(1)\). This includes the first-segment entry queue, the downstream boundary queues, the bounded segment interiors, and the final partial service opportunity. Dividing this terminal work by the scaled horizon \(\zeta T\) gives
\[
    \throughput_T^*-\mathbb{E}[\throughput^{(\zeta,\pi)}]
    =
    O((\zeta T)^{-1}).
\]
This argument does not require a lower bound on the arrival rate. If arrivals fall below the first threshold near the end of the horizon, fewer than \(n_1\) prompts can remain in the external entry queue, which is still a bounded terminal effect.

For service-normalized delay, the queue-integral argument from Appendix~\ref{appendix::proof of nested_wait_heavy_traffic} applies to the service periods and downstream boundary queues because their expected lengths are uniformly bounded. The only additional issue in the time-varying setting is waiting for the Segment 1 entry queue to reach its threshold during low-arrival intervals. The first-segment waiting condition gives \(\sup_t\zeta\,\mathbb E[\tau_1^{(\zeta)}(t)]<\infty\), so these idle waiting periods add only \(O(1)\) in service-normalized time. Hence
\[
    \mathbb{E}[\Latency^{(\zeta,\pi)}],
    \mathbb{E}[\TTFT^{(\zeta,\pi)}]
    =
    O(1)
\]
whenever the first-segment waiting condition holds.

For high-probability memory bounds, the analysis follows the same martingale construction applied to the time-invariant coupled process $\{ \bar{W}_{(k)}^b \}$ defined above. Each realized iteration has duration at least \(d_0/\zeta\), so the number of realized opportunities over \([0,T]\) is at most \(B_\zeta^{\max}=\lceil \zeta T/d_0\rceil\). By the same argument as in Appendix~\ref{appendix::proof of nested_wait_heavy_traffic}, with $\bar{X}_{(k)}^{r} = \bar{Y}_{(k)}^r - n_k$, $\bar{Y}_{(k)}^r \sim \text{Binomial}(n_{k-1}, p_k^*)$, and $p_k^*<n_k/n_{k-1}<1$, the memory bound is:
$$M_{\mathrm{req},\mathrm{tv}}^{(\zeta,\pi)} = M^{\pi} + \sum_{k=2}^m n_k \cdot (l + l_{k-1}') + \sum_{k=2}^m \theta_k^{-1}\ln\left( \frac{m(1+\zeta T/d_0)}{\delta} \right) \cdot (l + l_{k-1}'),$$
where $\theta_k$ ($2\leq k\leq m$) is the unique positive solution to:
\begin{equation}
e^{-\theta_k n_k} (1 - p_k^* + p_k^* e^{\theta_k})^{n_{k-1}} = 1.\label{eq::eq of theta, time varying}
\end{equation}
Since \(p_k^*\) is used to dominate every realized thinning probability, the same exponential-martingale argument as Appendix~\ref{appendix_lemma::proof of solution to martingale exists} applies with this worst-case binomial increment. The lower bound for \(\theta_k\) follows from the same calculation.

Let \(\mathcal E_{\mathrm{tv}}\) be the event that every downstream boundary residual is bounded by the corresponding deterministic-plus-logarithmic buffer in \(M_{\mathrm{req},\mathrm{tv}}^{(\zeta,\pi)}\) over all opportunities up to \(B_\zeta^{\max}\). The preceding martingale bound and a union bound over downstream segments give \(\mathbb P(\mathcal E_{\mathrm{tv}})\ge1-\delta\). Couple the finite-memory Nested WAIT implementation to the time-varying threshold-controlled comparison dynamics on the same arrivals and continuation marks until the first memory repair. On \(\mathcal E_{\mathrm{tv}}\), the same resident-boundary accounting as in Appendix~\ref{appendix::proof of nested_wait_heavy_traffic} applies: threshold-controlled interiors are charged to \(M^\pi\), fresh survivors are charged to the just-processed upstream batch, and only carryover resident boundary residuals require the logarithmic buffers. Hence the projected resident memory is at most \(M_{\mathrm{req},\mathrm{tv}}^{(\zeta,\pi)}\le C\) at both repair sites. Therefore memory repair is not invoked on \(\mathcal E_{\mathrm{tv}}\), the finite-memory implementation coincides with the comparison dynamics on \([0,T]\), and no memory-overflow-induced evictions occur on that event.

It remains to transfer the comparison-dynamics estimates to the finite-memory implementation without conditioning on \(\mathcal E_{\mathrm{tv}}\). Let \(A_T\) be the total number of external arrivals over \([0,T]\). Then \(A_T\) is Poisson with mean
\[
    \mu_T=\zeta\int_0^T\sum_{j=1}^m\lambda_j^u\,du .
\]
The uniform upper bound on arrival rates gives \(\mu_T=O(\zeta T)\). For any event \(F\) with \(\mathbb P(F)\le\delta\), the same Poisson split as in Theorem~\ref{thm:nested_wait_heavy_traffic} gives
\[
    \mathbb E[A_T{\bf 1}_F]\le 2\mu_T\delta+C_1\mu_T e^{-c_1\mu_T}
\]
for constants \(C_1,c_1>0\), with no independence requirement between \(F\) and \(A_T\). Taking \(F=\mathcal E_{\mathrm{tv}}^c\), the bad-event throughput contribution is at most \(l_m'\mathbb E[A_T{\bf 1}_{\mathcal E_{\mathrm{tv}}^c}]/(\zeta T)=O((\zeta T)^{-1})\), because \(\delta\le K(1+\zeta T)^{-1}\) and \(\sup_{\mu\ge0}\mu e^{-c_1\mu}<\infty\).

For delay and TTFT, the theorem invokes the first-segment waiting condition. Under this lower-side condition in the bounded-rate model, \(\mu_T=\Theta(\zeta T)\). Hence the finite-horizon censored bad-event contribution, normalized by \(\mathbb E[A_T]=\mu_T\), is at most
\[
    \zeta T\frac{\mathbb E[A_T{\bf 1}_{\mathcal E_{\mathrm{tv}}^c}]}{\mu_T}
    \le
    \zeta T\left(2\delta+C_1e^{-c_1\mu_T}\right)
    =
    O(1).
\]
Combining these bad-event bounds with the comparison-dynamics estimates above gives the stated unconditional finite-memory performance bounds and the eviction probability bound.

%% file: pf_lemmas.tex
\section{Proof of Lemmas}\label{sec:pf_lemmas}

Throughout this section, we use $b$ for the embedded review index and $B$ for the total number of review slots.

\subsection{Proof of Lemma~\ref{lemma::throughput gap, single}}\label{appendix_lemma::proof of throughput gap, single}
Write \(n\) for the critical single-type threshold, so \(X^b\sim\mathrm{Poisson}(n)\). Taking expectations on both sides of the state transition equation gives
$$\ex{}{W^{b+1}}=\ex{}{W^b}+n - n \mathbb{P}(W^b+X^b\geq n).$$
Summing from \(b=0\) to \(B-1\), and noting that \(W^0=0\), we obtain
$$\ex{}{W^B}=n B - n \ex{}{B-B_{\text{stuck}}}=n\cdot \ex{}{B_{\text{stuck}}}.$$

\subsection{Proof of Lemma~\ref{lem:coupled_process, single}}\label{appendix_lemma::proof of coupled_process, single}
We prove by induction on the review index \(b\). The coupled process \(\{\tilde{W}^b\}\) starts with a safety buffer of \(2n\), compared to the embedded review process starting at 0, ensuring it remains above \(W^b+n\) throughout. For the inductive step, we consider two cases based on whether the review process removes a threshold batch or experiences a stuck review slot.

\paragraph{Base case.} For \(b=0\), \(\tilde{W}^0=2n\geq n=W^0+n\).

\paragraph{Inductive step.} Assume \(\tilde{W}^b \geq W^b+n\) for some \(b \geq 0\). We show \(\tilde{W}^{b+1} \geq W^{b+1}+n\) by considering two cases:

\textbf{Case 1 (Threshold batch removed):} If \(W^b+X^b\geq n\), the queue has accumulated enough prompts to remove a threshold batch, so \(W^{b+1}=W^b+X^b-n\). Then
$$\tilde{W}^{b+1}=\max\{ 2n, \tilde{W}^b + X^b - n \}\geq \tilde{W}^b + X^b - n\geq( W^b+n )+ X^b - n=W^{b+1}+n.$$

\textbf{Case 2 (Stuck review slot):} If \(W^b+X^b<n\), insufficient prompts are available and the review process cannot remove a threshold batch. The queue simply accumulates arrivals: \(W^{b+1}=W^b+X^b\). The safety buffer ensures dominance:
$$\tilde{W}^{b+1}=\max\{ 2n, \tilde{W}^b + X^b - n \}\geq 2n>W^{b+1}+n.$$
By induction, \(\tilde{W}^b \geq W^b+n\) for all \(b \in \mathbb{N}\).

\subsection{Proof of Lemma~\ref{lemma::multiple-type coupled dominace}}\label{appendix_lemma::proof of multiple-type coupled dominace}
The lemma compares the type-\(j\) entry queue in the periodic comparison system with a reflected random walk that always subtracts \(n_j\) after adding one slot of arrivals, but never falls below the safety level \(2n_j\). Fix type \(j\) and write \(\bar W^b=\bar W^b_{(j)}\), \(\widetilde W^b=\widetilde W^b_{(j)}\), \(n=n_j\), and \(\bar X^b=\bar X^b_{(j)}\). We prove by induction that \(\widetilde W^b\ge \bar W^b\).

The claim holds at \(b=0\), since \(\widetilde W^0=2n\) and \(\bar W^0=0\). Suppose \(\widetilde W^b\ge \bar W^b\). If \(\bar W^b+\bar X^b\ge n\), then the comparison queue serves \(n\) prompts and
\[
    \bar W^{b+1}=\bar W^b+\bar X^b-n.
\]
Using the induction hypothesis,
\[
    \widetilde W^{b+1}
    =\max\{2n,\widetilde W^b+\bar X^b-n\}
    \ge \widetilde W^b+\bar X^b-n
    \ge \bar W^{b+1}.
\]
If \(\bar W^b+\bar X^b<n\), then no type-\(j\) service occurs in the comparison queue and \(\bar W^{b+1}=\bar W^b+\bar X^b<n\). In this case \(\widetilde W^{b+1}\ge2n>\bar W^{b+1}\). Thus the domination holds for all \(b\), and the argument applies independently to every type \(j\).

\subsection{Nested WAIT Coupling Bound}\label{appendix_lemma::proof of nested coupled process}
This lemma is used when a downstream boundary queue receives binomially thinned arrivals from the previous segment. We index the queue after the threshold review at the start of an opportunity in which the previous segment is processed. If the downstream boundary already has a threshold batch, Nested WAIT removes that batch in the same prefix batch; newly surviving prompts generated by the previous segment are treated as fresh boundary input for the next review.

Let \(W^b_{(k)}\) be the residual resident boundary queue for segment \(k\ge2\) after this review, and let \(Y^b_{(k)}\) be the number of newly surviving prompts at the next boundary-crossing opportunity. The queue satisfies
\[
    W^{b+1}_{(k)}
    \le
    W^b_{(k)}+Y^b_{(k)}
    -
    n_k{\bf 1}\{W^b_{(k)}+Y^b_{(k)}\ge n_k\}.
\]
Define \(X^b_{(k)}=Y^b_{(k)}-n_k\) and
\[
    \tilde{W}_{(k)}^0=n_k,\qquad
    \tilde{W}_{(k)}^{b+1}
    =
    \max\{n_k,\tilde{W}_{(k)}^b+X^b_{(k)}\}.
\]
We prove by induction that \(\tilde{W}^b_{(k)}\ge W^b_{(k)}\). At \(b=0\), the domination holds because \(\tilde W^0_{(k)}=n_k\) and \(W^0_{(k)}=0\). Suppose it holds at \(b\). If \(W^b_{(k)}\ge n_k\), then
\[
    W^{b+1}_{(k)}
    \le W^b_{(k)}+Y^b_{(k)}-n_k
    \le \tilde{W}^b_{(k)}+X^b_{(k)}
    \le \tilde{W}^{b+1}_{(k)}.
\]
If \(W^b_{(k)}<n_k\) but \(W^b_{(k)}+Y^b_{(k)}\ge n_k\), a threshold batch is removed at the next boundary review, and
\[
    W^{b+1}_{(k)}
    \le W^b_{(k)}+Y^b_{(k)}-n_k
    \le \tilde{W}^b_{(k)}+X^b_{(k)}
    \le \tilde{W}^{b+1}_{(k)}.
\]
If \(W^b_{(k)}+Y^b_{(k)}<n_k\), then
\[
    W^{b+1}_{(k)}=W^b_{(k)}+Y^b_{(k)}<n_k\le\tilde{W}^{b+1}_{(k)}.
\]
Thus the domination holds for all \(b\in\mathbb{N}\).

\subsection{Martingale Root Bound for Nested WAIT}\label{appendix_lemma::proof of solution to martingale exists}
This auxiliary calculation records the exponential tilting parameter used in the high-probability memory bound. Let \(0<p_k<n_k/n_{k-1}<1\), let \(\bar Y\sim\mathrm{Binomial}(n_{k-1},p_k)\), and set \(X=\bar Y-n_k\). The moment-generating equation
\[
    \mathbb{E}[e^{\theta X}]
    =
    e^{-\theta n_k}(1-p_k+p_ke^{\theta})^{n_{k-1}}
    =
    1
\]
has a unique positive solution \(\theta_k\). This is the standard Cramer--Lundberg root for a negative-drift random walk with bounded positive jumps \citep[see, e.g.,][]{asmussen2003applied}. With \(S^i_{(k)}=\sum_{r=1}^i X^r_{(k)}\), the process \(e^{\theta_k S^i_{(k)}}\) is therefore a mean-one martingale, and Doob's maximal inequality \citep[see, e.g.,][]{durrett2019probability} yields
\[
    \mathbb{P}\left(\max_{0\le i\le r}S^i_{(k)}\ge x\right)
    \le e^{-\theta_k x}.
\]
We will use the corresponding finite-horizon bound for the reflected process
\[
    Z^0_{(k)}=0,\qquad
    Z^{b+1}_{(k)}=\max\{0,Z^b_{(k)}+X^b_{(k)}\}.
\]
If \(\max_{0\le b\le B}Z^b_{(k)}\ge x\), then some excursion of the random walk has gained at least \(x\): there are \(0\le a<b\le B\) such that \(\sum_{r=a}^{b-1}X^r_{(k)}\ge x\). Applying the preceding maximal inequality from each possible excursion start and unioning over at most \(B\) starts gives
\[
    \mathbb{P}\left(\max_{0\le b\le B}Z^b_{(k)}\ge x\right)
    \le B e^{-\theta_k x},\qquad B\ge1.
\]
For completeness, we also record the elementary drift-to-root bound used in the theorem. Writing \(D=n_k-n_{k-1}p_k>0\) and \(g(\theta)=\log\mathbb{E}[e^{\theta X}]\), we have \(g(0)=0\), \(g'(0)=-D\), and \(g''(\theta)\le n_{k-1}/4\). Applying Taylor's theorem at the positive root \(g(\theta_k)=0\) gives
\begin{align}
    \theta_k \geq \frac{8 (n_k - n_{k-1} p_k)}{n_{k-1}}.\label{eq::theta lower bound}
\end{align}

%% file: appendix_b_additional.tex
\section{Additional Experiments}
\label{app:additional_experiments}

This appendix reports supplemental robustness and implementation-fidelity analyses for Section~\ref{sec:experiments}. Section~\ref{app:nested_wait_pseudocode} records the full event-driven Nested WAIT pseudocode used for implementation. Section~\ref{app:no_wait} isolates the role of the accumulation requirement by comparing WAIT with a threshold-only variant. Section~\ref{app:gpu_validation} validates the simulator against direct GPU measurements and reports supplemental physical-GPU comparisons on single-type and real-data workloads. Section~\ref{app:multi_seed} examines repeated-run variability near the observed transition from near-overloaded to overloaded operation, and Section~\ref{app:pd_disagg} evaluates the policy in a prefill-decode-disaggregated deployment.

\subsection{Nested WAIT implementation pseudocode}
\label{app:nested_wait_pseudocode}

Algorithm~\ref{alg:nested_wait_full} gives the event-driven implementation corresponding to the main-text description in Algorithm~\ref{alg:nested_wait}. It makes explicit the queue transitions at segment boundaries and the resident-boundary memory-repair rule.

\input{nested_wait_full_algorithm}

\subsection{Threshold without waiting}
\label{app:no_wait}

We isolate the role of waiting on the single-type \texttt{p512d20} workload by comparing full WAIT with a threshold-only variant, denoted \emph{WAIT-no-wait}. For each arrival rate, the two policies use the same parameter setting and hence the same threshold upper bound. The threshold-only variant enforces this upper bound but does not wait for enough requests to accumulate before service. The comparison therefore holds the admission-capacity scale fixed and changes only whether service is delayed to form a full threshold batch.

\begin{figure}[ht]
    \centering
    \includegraphics[width=0.78\textwidth]{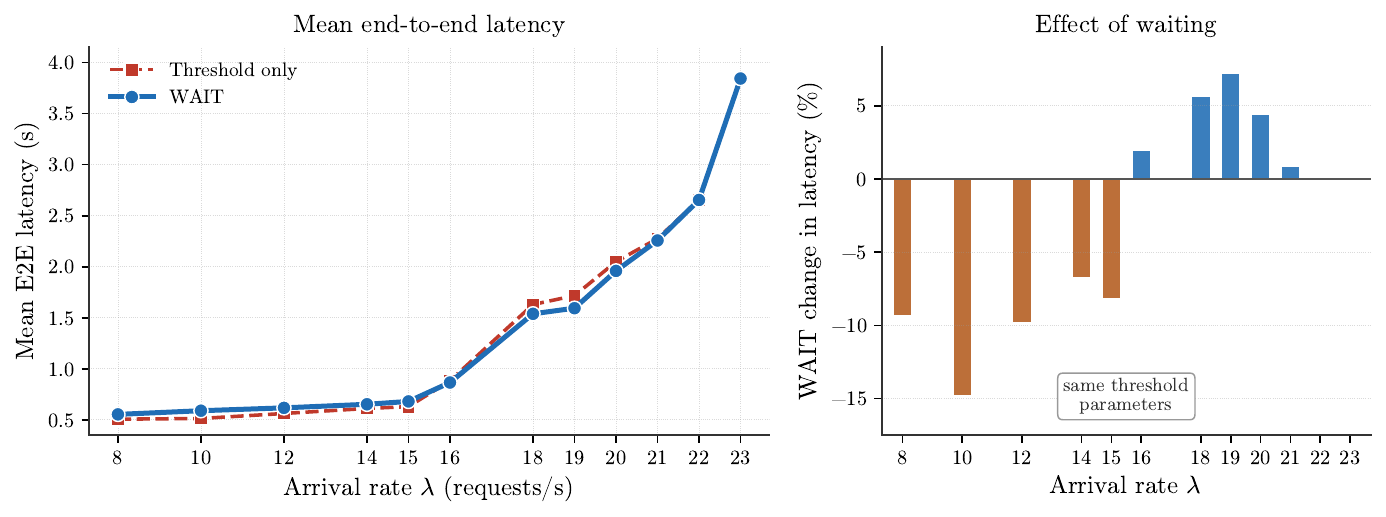}
    \caption{Effect of waiting on the single-type \texttt{p512d20} Vidur workload. At each arrival rate, full WAIT and the threshold-only variant use the same parameter setting; the comparison only changes whether service waits for enough requests to accumulate before scheduling.}
    \label{fig:wait_gate_ablation}
\end{figure}

Figure~\ref{fig:wait_gate_ablation} reports the comparison across arrival rates. At low arrival rates, waiting introduces a modest latency cost: requests may wait to accumulate even when the server has enough capacity to process them immediately. This cost shrinks as the arrival rate increases. Near the overload boundary, waiting becomes useful because it regulates the timing at which work enters service, reducing latency relative to the threshold-only variant. Once the system is overloaded, the upper bound becomes the dominant control, and the two variants again behave similarly.

\subsection{Real-GPU validation}
\label{app:gpu_validation}

We complement the Vidur experiments in Section~\ref{sec:experiments} with measurements on a physical NVIDIA A100~80\,GB serving Llama-2-7B. These runs serve two purposes. First, direct GPU measurements calibrate the iteration-time model used by Vidur. Second, we implement WAIT's admission logic using SGLang~0.5.7, an open-source LLM serving framework~\citep{sglang-github}, and test the resulting scheduler in end-to-end GPU runs. We report three comparisons: iteration-time measurements against Vidur's prediction, a single-type GPU workload corresponding to Section~\ref{sec:exp_synthetic}, and a real-data GPU workload corresponding to Section~\ref{sec:exp_real}.

\begin{figure}[ht]
    \centering
    \includegraphics[width=0.6\textwidth]{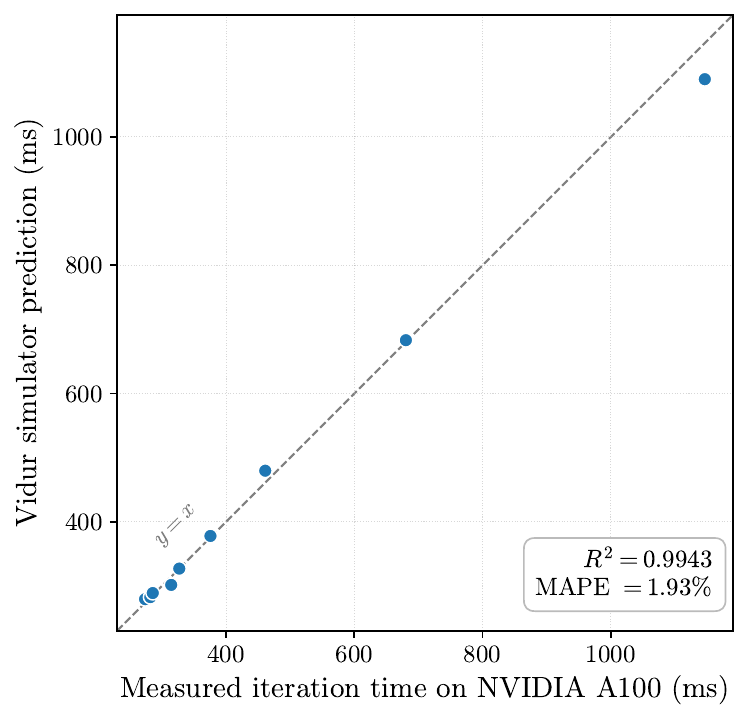}
    \caption{Simulator calibration against NVIDIA A100~80\,GB GPU measurements. Each marker is one batch-size configuration: its horizontal coordinate is the measured GPU batch inference time, and its vertical coordinate is the Vidur prediction. The comparison uses prefill length $256$ and decode length $20$, with powers-of-two batch sizes through the profiled range $B\le128$ and one extrapolation check to $B=256$. The dashed line is the $y=x$ reference. The fitted linear iteration-time model in Equation~\eqref{eq:time_consump} achieves $R^2=0.9943$ and mean absolute percentage error $1.93\%$ on the plotted grid.}
    \label{fig:sim_fidelity}
\end{figure}

\paragraph{Simulator vs GPU.}
To calibrate the simulator in the operating region used in Section~\ref{sec:experiments}, we compare Vidur's iteration-time predictions with direct measurements on the same A100~80\,GB GPU running Llama-2-7B under the SGLang baseline scheduler. Vidur's raw Llama-2-7B attention profiling covers batch sizes up to $128$. Within this profiled range, the validation uses batch sizes $B \in \{1,2,4,\ldots,128\}$ with prefill length $256$ and decode length $20$; we also test extrapolation to $B=256$. In Figure~\ref{fig:sim_fidelity}, each point is one batch-size configuration: the horizontal coordinate is the measured GPU batch inference time, and the vertical coordinate is Vidur's prediction. Points near the $y=x$ line therefore indicate accurate simulator predictions. On the plotted grid, the fitted linear iteration-time model has mean absolute percentage error $1.93\%$ and $R^2=0.9943$; extrapolation to $B=256$ has absolute percentage error $4.93\%$. The fitted intercept and slope are $d_0 \approx 276$\,ms and $d_1 \approx 0.0115$\,ms per KV-cache token, respectively, consistent with the time model in Equation~\eqref{eq:time_consump}. This agreement supports the Vidur-based latency and operating-range comparisons in Section~\ref{sec:experiments} for the workloads studied here.

\begin{figure}[ht]
    \centering
    \includegraphics[width=0.65\textwidth]{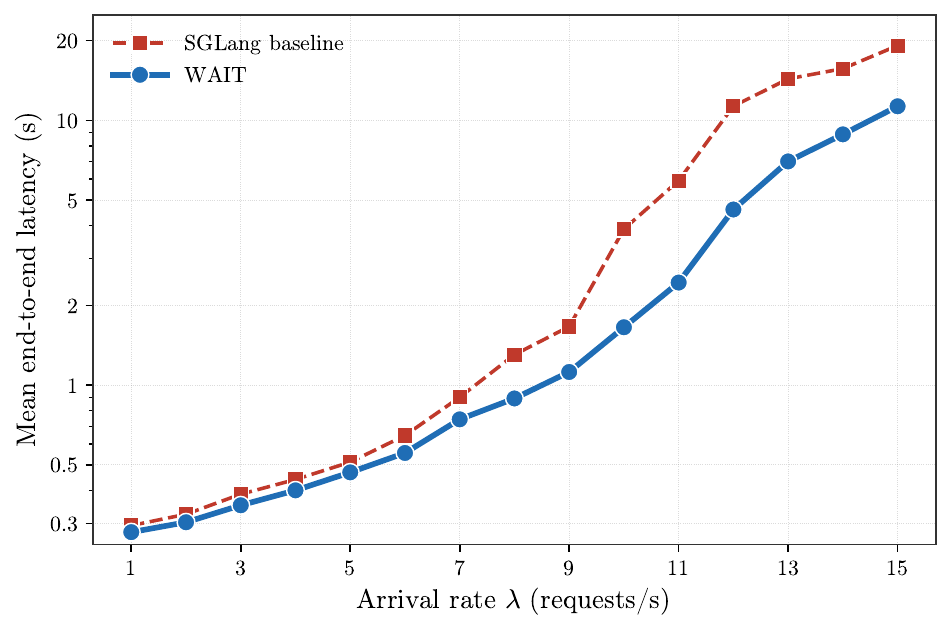}
    \caption{Single-type GPU experiment on SGLang~0.5.7 (NVIDIA A100~80\,GB, Llama-2-7B, random prompts with input length $512$ and output length $20$): mean end-to-end latency versus arrival rate. WAIT is compared with the feasible SGLang baseline scheduler configuration used in this experiment and yields a mean latency reduction of $29.7\%$ across the arrival-rate grid.}
    \label{fig:sglang}
\end{figure}

\paragraph{Single-type workload on GPU.}
Using the same single-type workload as Section~\ref{sec:exp_synthetic}, we run end-to-end SGLang experiments with Poisson arrivals at $\lambda \in \{1,2,\ldots,15\}$ requests/s and $500$ prompts per rate. The SGLang baseline scheduler uses \texttt{chunked\_prefill\_size}$=256$ and \texttt{prefill\_max\_requests}$=8$; its default configuration triggers memory-overflow failures at several mid-to-high rates and is therefore not a feasible baseline on this workload. WAIT uses the same parameter values at all rates. Its system-wide batch-size cap is $\mathrm{tl}=21$, and its per-stage WAIT threshold is $n_1=1$. The chunk size is $384$, so a prefill of length $512$ is processed as $K=\lceil512/384\rceil=2$ prefill blocks before the request enters the $20$ decode-token stages. Thus, the single-type pipeline has $K+20=22$ stages: WAIT schedules at most one request from each active stage, while $\mathrm{tl}=21$ caps how many requests can be served concurrently across these stages. The SGLang patch implements this policy through the batch-size cap and the per-request prefill chunk cap; a transition block of size $10$ switches the scheduler back to SGLang's native rule in lightly loaded states. Figure~\ref{fig:sglang} reports mean end-to-end latency as $\lambda$ varies. Under this GPU configuration, WAIT is consistently lower-latency than the feasible SGLang baseline scheduler, with an average reduction of $29.7\%$ across the arrival-rate grid.

\begin{figure}[ht]
    \centering
    \includegraphics[width=0.58\textwidth]{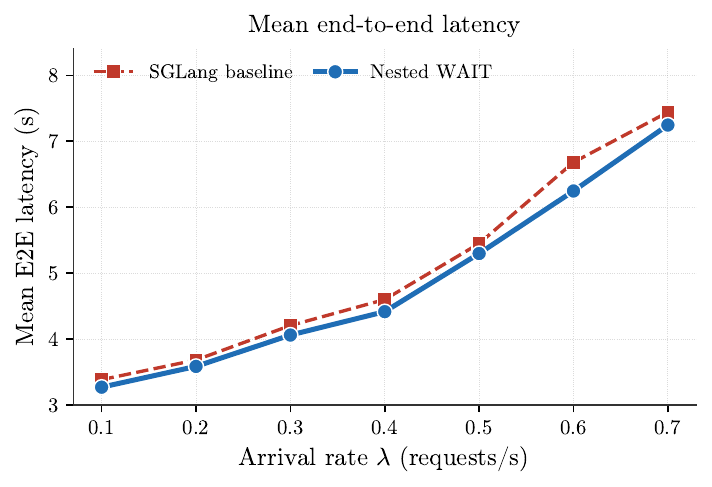}
    \caption{Real-data GPU experiment on SGLang~0.5.7 with the lmsys-chat-1m workload distribution: mean end-to-end latency versus arrival rate for the SGLang baseline scheduler and Nested WAIT. The Nested WAIT runs use the rate-specific parameter settings in Table~\ref{tab:sglang_realtrace}.}
    \label{fig:sglang_realtrace}
\end{figure}

\paragraph{Real dataset on GPU.}
Using the same lmsys-chat-1m workload distribution as Section~\ref{sec:exp_real}, we run paired SGLang experiments with Poisson arrivals at $\lambda \in \{0.1, 0.2, \ldots, 0.7\}$ requests/s and $200$ prompts per rate. The server is restarted before each rate so that the SGLang baseline and Nested WAIT are compared from the same clean GPU state. Nested WAIT uses chunk size $384$ and a rate-specific system-wide batch-size cap $\mathrm{tl}$. For the real-data runs, $\mathrm{tl}$ is distributed over the decode horizon through the listed segment partitions, using the same continuation-based allocation described in Section~\ref{sec:exp_real}. Table~\ref{tab:sglang_realtrace} reports the batch-size caps and decode-stage partitions used in these GPU runs. Figure~\ref{fig:sglang_realtrace} reports mean end-to-end latency in absolute units. Across the tested rates, Nested WAIT has lower latency than the SGLang baseline, with the largest reduction at $\lambda = 0.6$ and an average reduction of $3.6\%$.

\begin{table}[ht]
\centering
\small
\begin{tabularx}{0.82\textwidth}{>{\centering\arraybackslash}p{0.18\textwidth}>{\centering\arraybackslash}p{0.14\textwidth}>{\centering\arraybackslash}X}
\toprule
$\lambda$ (req/s) & $\mathrm{tl}$ & Decode-stage segments \\
\midrule
$0.1$ & $40$  & $[0,175), [175,350), [350,500]$ \\
$0.2$ & $50$  & $[0,210), [210,420), [420,500]$ \\
$0.3$ & $60$  & $[0,200), [200,400), [400,500]$ \\
$0.4$ & $70$  & $[0,175), [175,350), [350,500]$ \\
$0.5$ & $70$  & $[0,200), [200,400), [400,500]$ \\
$0.6$ & $70$  & $[0,175), [175,350), [350,500]$ \\
$0.7$ & $95$  & $[0,250), [250,500]$ \\
\bottomrule
\end{tabularx}
\caption{Nested WAIT parameter settings for the real-data GPU experiment on SGLang~0.5.7 with the lmsys-chat-1m workload distribution, NVIDIA A100~80\,GB, Llama-2-7B, and $200$ prompts per rate. The column $\mathrm{tl}$ gives the system-wide batch-size cap; the decode-stage ranges specify the segment partition over the $500$-token decode horizon. Once these parameters are fixed, the per-stage thresholds are computed by the continuation-based allocation rule in Section~\ref{sec:exp_real}.}
\label{tab:sglang_realtrace}
\end{table}

\subsection{Repeated-run variability}
\label{app:multi_seed}

The main arrival-rate figures in Section~\ref{sec:experiments} report averages over $10$ independent simulation replications at each plotted configuration. To examine sampling variability near the observed transition from near-overloaded to overloaded operation, we additionally repeat the near-transition configurations $20$ times and report the corresponding deviations.

The repeated-run diagnostic preserves the same latency ordering as the main figures across independent arrival draws.

\begin{table}[ht]
\centering
\small
\begin{tabular}{llccc}
\toprule
Workload & $\lambda$ & vLLM (s) & Sarathi (s) & WAIT (s) \\
\midrule
Single-type & $22$ & $18.23\,\pm\,2.74$ & $1.85\,\pm\,0.33$ & $\mathbf{1.03\,\pm\,0.04}$ \\
Single-type & $23$ & $26.78\,\pm\,3.35$ & $5.08\,\pm\,1.72$ & $\mathbf{1.29\,\pm\,0.10}$ \\
Two-type & $22$ & $56.70\,\pm\,2.22$ & $4.06\,\pm\,1.66$ & $\mathbf{2.36\,\pm\,0.57}$ \\
Two-type & $23$ & $80.25\,\pm\,2.37$ & $8.41\,\pm\,1.74$ & $\mathbf{5.66\,\pm\,1.88}$ \\
\bottomrule
\end{tabular}
\caption{Repeated-run reproducibility check near the observed transition from near-overloaded to overloaded operation. Each entry reports mean end-to-end latency together with the sample standard deviation across repeated simulation runs.}
\label{tab:multi_seed}
\end{table}

\subsection{Prefill-decode-disaggregated deployment}
\label{app:pd_disagg}

Production LLM serving increasingly uses a \emph{prefill-decode disaggregation} (PD) architecture, in which prefill and decode are executed by separate services and the KV cache is transferred between them~\citep{patel2024splitwise,zhong2024distserve}. This architecture aligns more closely with the memory-dependent timing model in Section~\ref{sec:model}. Because the decode server no longer mixes prefill and decode work, each iteration processes one token for each active post-prefill job, and the iteration time is driven primarily by active KV-cache usage. The linear iteration-time model in~\eqref{eq:time_consump} is therefore especially appropriate in this setting. The corresponding control decision is how many post-prefill jobs to keep in concurrent decoding; the WAIT threshold implements this decision as an upper bound on the decode-side service population.

\begin{figure}[ht]
    \centering
    \includegraphics[width=0.65\textwidth]{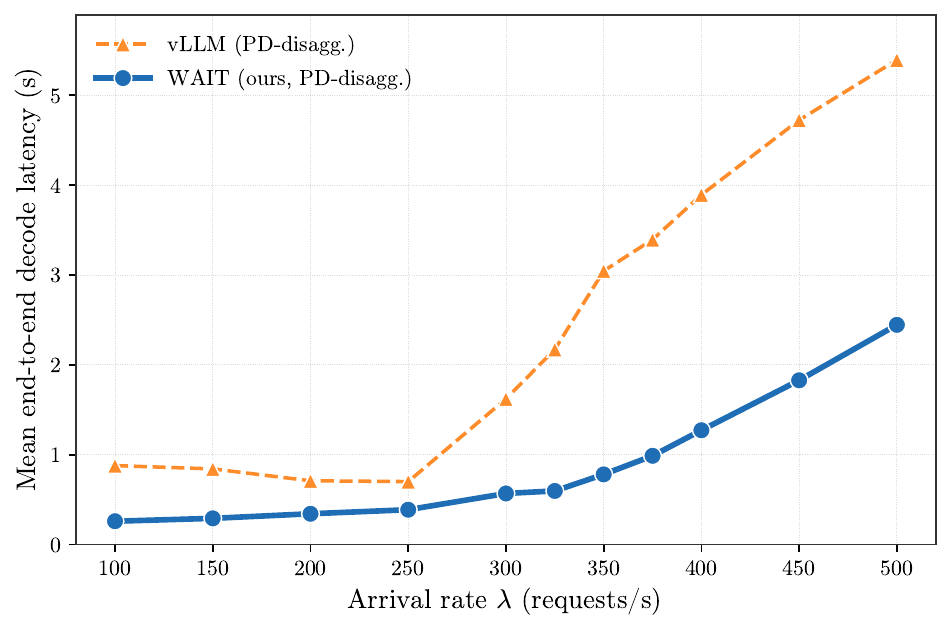}
    \caption{Prefill-decode-disaggregated deployment (\texttt{p630d20}, decode-only): mean end-to-end decode latency versus arrival rate. WAIT attains between $45\%$ and $55\%$ latency reduction over the vLLM PD baseline across the tested arrival-rate range.}
    \label{fig:pd_disagg}
\end{figure}

We implement this setting in Vidur by using a decode-only request generator: each request enters the simulated decode server after prefill has completed, with post-prefill context length $630$ tokens and decode length $20$ (\texttt{p630d20}). Arrivals follow a Poisson process with $\lambda \in \{100, 150, \ldots, 500\}$ requests/s. In this decode-only setting, Sarathi's chunked-prefill rule no longer changes the scheduling decision, so vLLM's PD scheduler (\texttt{vLLM (PD)}) represents the baseline. Figure~\ref{fig:pd_disagg} reports mean end-to-end decode latency: WAIT improves over vLLM at every rate, with the relative gap ranging from $45\%$ at $\lambda = 100$ to $55\%$ at $\lambda = 500$. Thus, in this tested decode-only setting, the threshold rule improves delay by controlling the decode-side batch composition.

%% file: nested_wait_full_algorithm.tex
\begin{breakablealgorithm}
\caption{Full Event-Driven Implementation of Nested WAIT}
\label{alg:nested_wait_full}
{\footnotesize\linespread{0.98}\selectfont
\begin{algorithmic}[1]
\Require Memory capacity \( C \), arrival rates \( \lambda_j \) for \( j \in [m] \), thresholds \( n_k \) for segments \( k \in [m] \)
\Require Output lengths \( l_1' < l_2' < \cdots < l_m' \) for output-length classes \( j \in [m] \)
    \Statex \Comment{$Q_{1,0}$ is external; for $k\ge2$, $Q_{k,l_{k-1}'}$ waits outside segment $k$ but remains GPU-resident}
    \Statex \Comment{Prompts in stages $l_{k-1}'+1,\ldots,l_k'$ have already been admitted into segment $k$}
    \Statex \Comment{Memory repair uses resident boundary queues; before launch, only unselected boundary prompts are eligible}
    \State Initialize all segment-entry, boundary, and interior queues to zero
    \State Initialize event queue; set current time \( t \gets 0 \)
    \State Set server status to idle
    \While{True}
        \State Wait for the next event (arrival or batch completion)
        \If{event is an arrival}
            \State \( Q_{1,0} \gets Q_{1,0} + 1 \)  \Comment{New prompts enter segment 1 at stage 0}
        \ElsIf{event is a batch completion for prefix \(1,\dots,k\)}
            \State Retrieve the selected counts \(a_{k's}\) stored with the completed prefix batch
            \State Update all affected queues simultaneously using pre-completion queue values
            \For{each processed segment \( k'\le k \)}
                \For{each selected nonterminal stage \(s<l_{k'}'\)}
                    \State \(Q_{k',s}\gets Q_{k',s}-a_{k's}\), \quad \(Q_{k',s+1}\gets Q_{k',s+1}+a_{k's}\)
                \EndFor
                \If{\(k'<m\)}
                    \State Let \(c_{k'}\) be the number of selected prompts at \(l_{k'}'\) that complete there
                \Else
                    \State Set \(c_m\gets a_{m,l_m'}\)
                \EndIf
                \State Set the survivor count \(r_{k'}\gets a_{k',l_{k'}'}-c_{k'}\)
                \State \(Q_{k',l_{k'}'}\gets Q_{k',l_{k'}'}-a_{k',l_{k'}'}\)
                \If{\(k'<m\)}
                    \State \(Q_{k'+1,l_{k'}'}\gets Q_{k'+1,l_{k'}'}+r_{k'}\) \Comment{Survivors join the next resident boundary queue}
                \Else
                    \State Clear KV caches of completed prompts
                \EndIf
            \EndFor
            \While{\(M_{\mathrm{res}}(Q)>C\) and a resident boundary queue is nonempty}
                \State Choose \(h\ge2\) whose resident boundary queue contains the most recently admitted prompt
                \State \(Q_{h,l_{h-1}'}\gets Q_{h,l_{h-1}'}-1\), \quad \(Q_{1,0}\gets Q_{1,0}+1\)
                \State Discard the evicted prompt's KV cache and accumulated progress
            \EndWhile
            \State Set server status to idle
        \EndIf
        \State Find largest \( k \) such that \( Q_{k', l_{k'-1}'} \geq n_{k'} \) for all \( k' \leq k \) \Comment{All segments up to $k$ meet threshold}
        \If{server is idle and such \( k \) exists}
            \State Set \(S_1=\{0,\ldots,l_1'\}\)
            \State For \(k'=2,\ldots,k\), set \(S_{k'}=\{l_{k'-1}',\ldots,l_{k'}'\}\)
            \State For every \(k'\le k\) and \(s\in S_{k'}\), set \(a_{k's}\gets\min\{n_{k'},Q_{k',s}\}\)
            \State Form the prefix batch from all selected prompts and store \(a_{k's}\)
            \State Let \(Q^+\) be the conservative projected post-iteration queue state
            \While{\(M_{\mathrm{res}}(Q^+)>C\) and an unselected resident boundary prompt exists}
                \State Choose \(h\ge2\) whose unselected resident boundary prompts include the most recently admitted prompt
                \State \(Q_{h,l_{h-1}'}\gets Q_{h,l_{h-1}'}-1\), \quad \(Q_{1,0}\gets Q_{1,0}+1\)
                \State \(Q^+_{h,l_{h-1}'}\gets Q^+_{h,l_{h-1}'}-1\)
                \State Discard the evicted prompt's KV cache and accumulated progress
            \EndWhile
            \State Process batch; already resident prompts outside the batch wait with their KV caches retained
            \State Set server status to busy until the completion event for this prefix batch
        \EndIf
    \EndWhile
\end{algorithmic}
}
\end{breakablealgorithm}